\documentclass[10pt,twocolumn,letterpaper]{article}

\usepackage[pagenumbers]{wacv} 

\usepackage[toc,page,header]{appendix}
\usepackage{minitoc}

\usepackage{graphicx}
\usepackage{booktabs}
\usepackage{dsfont}
\usepackage{multirow}
\usepackage{multicol}
\usepackage{pifont}
\usepackage{amsmath}
\usepackage{epigraph}
\usepackage{float}
\usepackage[dvipsnames]{xcolor}

\usepackage[T1]{fontenc}
\usepackage[breakable, theorems, skins]{tcolorbox}
\tcbset{enhanced}
\DeclareRobustCommand{\mybox}[2][gray!20]{
\begin{tcolorbox}[
        breakable,
        left=0pt,
        right=0pt,
        top=0pt,
        bottom=0pt,
        colback=#1,
        colframe=#1,
        width=\columnwidth,
        enlarge left by=0mm,
        boxsep=5pt,
        arc=0pt,outer arc=0pt,
        ]
        {\fontfamily{lmtt}\selectfont\footnotesize #2}
\end{tcolorbox}
}

\definecolor{wacvblue}{rgb}{0.21,0.49,0.74}
\usepackage[pagebackref,breaklinks,colorlinks,allcolors=wacvblue]{hyperref}
\hypersetup{
    colorlinks=true,
    urlcolor=magenta
}

\def\wacvPaperID{614} 
\def\confName{WACV}
\def\confYear{2026}

\title{Spacewalk-18: A Benchmark for Multimodal and Long-form \\Procedural Video Understanding in Novel Domains}

\author{Zitian Tang\textsuperscript{*} \quad
Rohan Myer Krishnan\textsuperscript{*} \quad
Zhiqiu Yu \quad
Chen Sun\\
Brown University}

\begin{document}

\doparttoc 
\faketableofcontents 

\maketitle
\begin{abstract}
Learning from (procedural) videos has increasingly served as a pathway for embodied agents to acquire skills from human demonstrations. To do this, video understanding models must be able to obtain structured understandings, such as the temporal segmentation of a demonstration into sequences of actions and skills, and to generalize the understandings to novel environments, tasks, and problem domains. In pursuit of this goal, we introduce Spacewalk-18, a benchmark containing two tasks: (1) step recognition and (2) video question answering, over a dataset of temporally segmented and labeled tasks in International Space Station spacewalk recordings. In tandem, the two tasks quantify a model's ability to: (1) generalize to novel domains; (2) utilize long temporal context and multimodal (\eg visual and speech) information. Our extensive experimental analysis highlights the challenges of Spacewalk-18, but also suggests best practices for domain generalization and long-form understanding. Notably, we discover a promising adaptation via summarization technique that leads to significant performance improvement without model fine-tuning.
The Spacewalk-18 benchmark is released at \url{https://brown-palm.github.io/Spacewalk-18}.
\end{abstract}
\vspace{-2em}
    
\section{Introduction}
\label{sec:intro}

\begin{figure}[t]
    \centering
    \includegraphics[width=\linewidth]{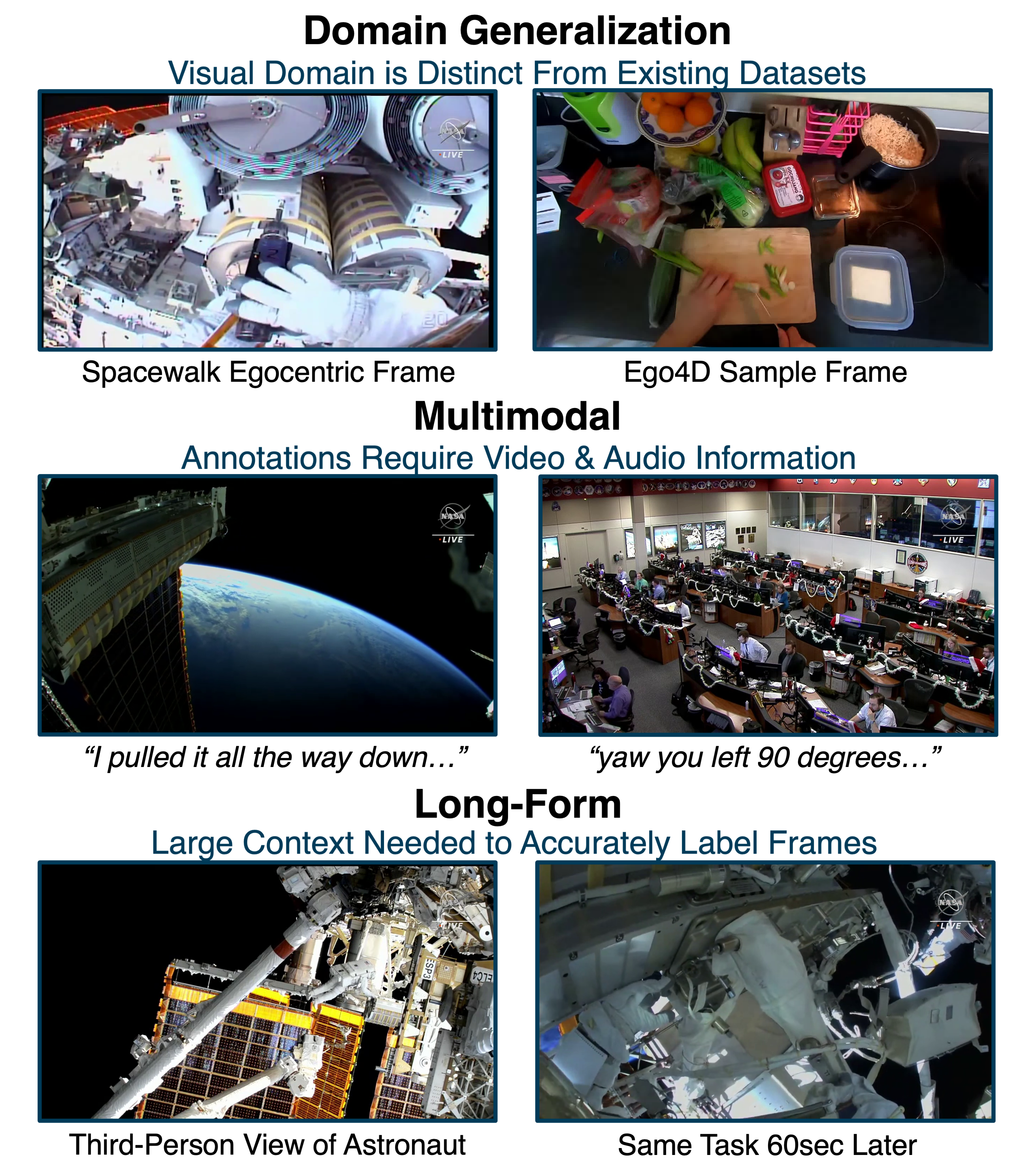}
    \vspace{-15pt}
    \caption{\textbf{Key properties of Spacewalk-18}: (1) Domain generalization: a side-by-side comparison of a sample frame from Spacewalk-18 and Ego4D~\cite{Ego4D} illustrates our benchmark's novel domain. (2) Multimodal: the visual content of the left frame does not align with its audio/speech. Instead, the speech corresponding to the right frame describes the left frame. (3) Long-form: the left frame shows the  astronaut working on the solar array and the right frame shows him releasing bolts. These contextualize each other to identify that he is releasing the solar array in both frames. }
    \label{fig:teaser}
    \vspace{-15pt}
\end{figure}

\setlength{\epigraphwidth}{\columnwidth}
\epigraph{This is Ground Control to Major Tom\\
You've really made the grade\\
And the papers want to know whose shirts you wear\\
Now it's time to \textit{leave the capsule}, if you dare}{\textit{Space Oddity}
\vspace{-10pt}}

Procedural videos, such as how-to or cooking, are produced to spread knowledge for human learners. With the recent advances in video understanding and robotic learning~\citep{bharadhwaj2023zero,bahl2023affordances,sontakke2023roboclip}, these videos have  become increasingly valuable learning resources for robots too. However, in order for machines to efficiently learn complex tasks and work alongside humans, they must be able to distill complex procedures into series of steps with only a few examples, often in a previously unseen environment. 
This challenge inspires the development of procedural video understanding systems that are able to generalize to previously unseen scenarios, and require minimal human supervision.

We introduce a new benchmark, \textbf{Spacewalk-18}, to advance multimodal, long-form, and procedural video understanding in a \textit{novel} domain. 
Whereas most of the existing procedural video benchmarks are sourced from daily household scenarios~\citep{ActivityNet, Breakfast, Salads, COIN, EpicKitchens, Ego4D}, the Spacewalk-18 dataset comprises video clips from 18 recorded extravehicular activities (spacewalks) outside the International Space Station. Videos in this domain are naturally limited by the number of recorded spacewalks. As illustrated in \cref{fig:teaser}, our dataset is first and foremost a benchmark to test pre-trained video-language models' (\eg~\cite{alayrac2022flamingo,wang2022internvideo}) capability to \textit{leave the capsule} and generalize to novel domains. Spacewalk-18 is also inherently multimodal and requires effective incorporation of long-form temporal context.

In Spacewalk-18,
astronauts go on spacewalks for a variety of reasons, including to perform experiments, test equipment, or carry out maintenance/repairs. 
The recordings follow a fairly rigid agenda, often illustrated through a short animated sequence giving an overview of the steps planned for the spacewalk. Since our goal is to evaluate a (pre-trained) model's generalization capability, we follow the design choice of recent benchmarks with similar goals (\emph{e.g.},~\citep{PerceptionTest,EgoSchema}) and provide a moderate-sized training data for adaptation rather than pre-training. In order to collect detailed and dense temporal annotations for the training, validation, and test videos,
we introduce a new protocol for temporal segmentation and action labeling to efficiently annotate the recordings and categorize content from the mission into the corresponding animated steps. 

\begin{figure*}[t]
    \centering
    \includegraphics[width=.89\linewidth]{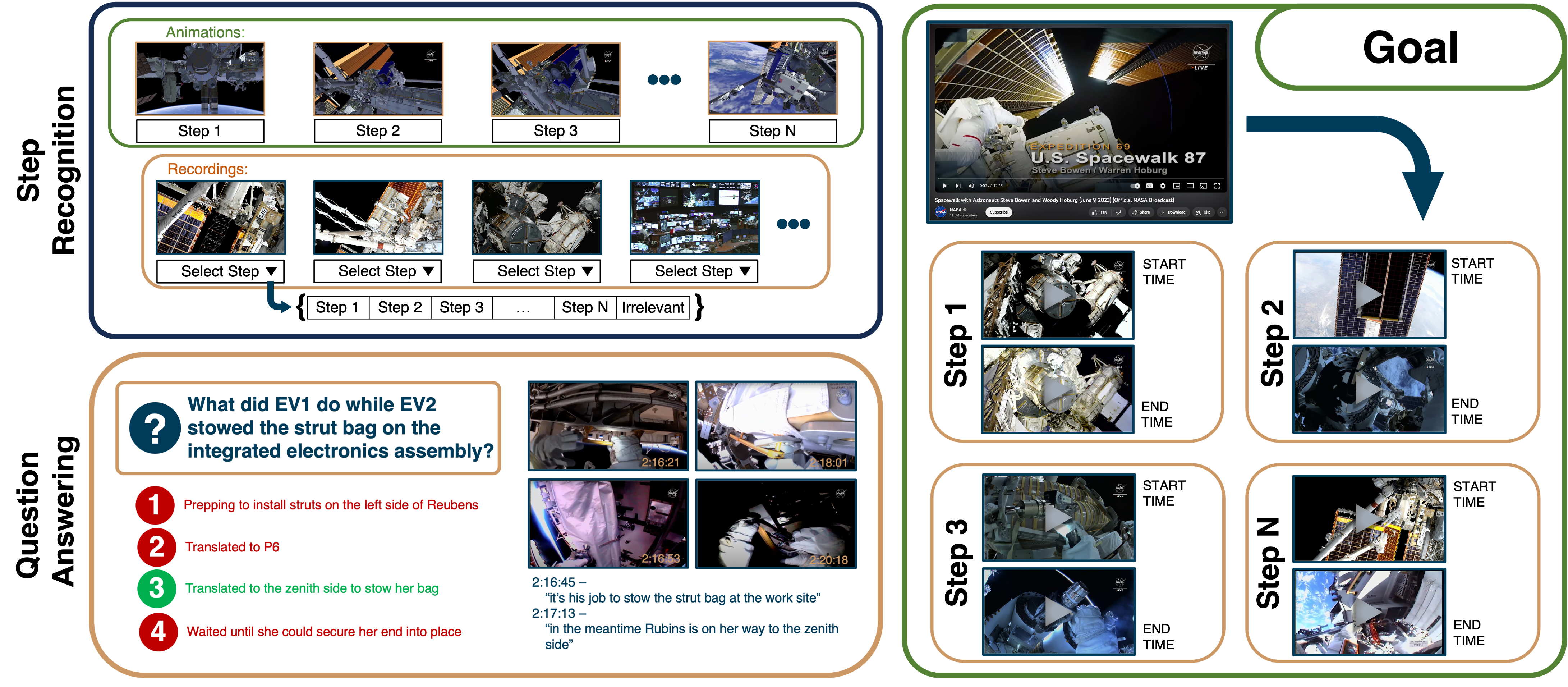}
    \vspace{-5pt}
    \caption{A spacewalk recording can be 7 or 8 hours long. The \textit{step recognition} task aims to assign each video clip in the recording a step label, which is illustrated by a short animation and a text description. The \textit{question answering} task targets video reasoning with long-term multimodal context. Both serve as intermediate benchmarks towards the ``Goal'', which aims to represent a long procedural video as a sequence of steps and their corresponding video demonstrations for understanding and reasoning.}
    \label{fig:tasks}
    \vspace{-1em}
\end{figure*}

As illustrated in Figure~\ref{fig:tasks}, the collected annotations provide a structured representation of each spacewalk recording as a sequence of steps, each of which is in the form of a text description (\emph{e.g.}, ``install thermal blanket on degraded antenna''), an animated illustration, and the temporal boundaries within the recording where the step is performed. We define two proxy tasks for evaluation: 
\begin{enumerate}
    \item The \textit{step recognition} task evaluates the model's ability to generalize to the spacewalk domain and incorporate video and text content into predictions.
    \item The \textit{question answering} task benchmarks the model's ability to perform spatiotemporal reasoning.
\end{enumerate}

We evaluate state-of-the-art video-language models,
including contrastive video-language models (VLMs)~\citep{kevin2022egovlp, xu-etal-2021-videoclip, wang2022internvideo, wang2024internvideo2} and video large language models (VLLMs)~\citep{wang2023vamos, llavanextvideo, videollama2, shen2024longvu, bai2025qwen25vltechnicalreport, zhu2025internvl3exploringadvancedtraining}.
Through our experiments, we identify several best practices for incorporating multimodal information and long-form temporal context. Our human evaluation shows that an average English speaker achieves 67\% step recognition performance after viewing 3.5-minute ``training'' example video, far exceeding any state-of-the-art VLLMs or proprietary APIs (\emph{e.g.}, GPT-5), suggesting a large room for improvement. Surprisingly, we discover that the most effective adaptation technique is to provide a frozen VLLM a summarized video as its context, and the relative gain is significantly larger compared to directly fine-tuning a model on the training split of Spacewalk-18.

Our contributions are three-fold: 
First, we propose a new procedural video benchmark on learning structured video representations and reasoning. 
Second, we collect the Spacewalk-18 dataset with a new protocol for efficient annotation. 
It contains 96 hours of densely annotated videos and spans over 455 animated steps. 
Finally, we conduct extensive experimental analyses and discover the adaptation via summarization technique for effective domain generalization in Spacewalk-18.

\section{Related Work}
\label{sec:related}

\noindent\textbf{Procedural Video Understanding} has important applications in video summarization~\citep{narasimhan2022tl}, human machine interaction~\citep{nagarajan2020ego}, procedural planning~\citep{chang2020procedure}, and robotic learning~\citep{wu2023daydreamer,dai2023learning}. Common tasks defined for procedural video understanding include temporal action segmentation~\citep{ding2023temporal,dwibedi2020counting} and detection~\citep{richard2016temporal}, step localization~\citep{zhukov2019cross, COIN} and prediction~\citep{abu2018will,Ego4D,girdhar2021anticipative}. They either use dense segment-level annotations or weak video-level labels~\citep{richard2017weakly,miech2019howto100m,souvcek2022multi}. Existing procedural video benchmarks are mainly sourced from two domains, how-to videos from online platforms~\citep{COIN,zhukov2019cross,miech2019howto100m,souvcek2022look}, and ego-centric videos collected from recruited actors~\citep{EpicKitchens,Ego4D,sener2022assembly101} in kitchens and other household scenarios. More recently, the Perception Test~\citep{PerceptionTest} benchmark, aims to evaluate the zero-shot generalization capabilities of video-language foundation models. Our Spacewalk-18 benchmark is complementary to all of the existing procedural video understanding benchmarks: The spacewalk recordings are scripted and narrated in detail, yet cannot be solved by the visual or language modality alone. They exhibit strong structures and dependencies, and unfold over long temporal horizons. More importantly, unlike existing videos that are captured in kitchens and on earth, the space station is a novel domain, which simulates the real-world scenario of model deployment in unseen environments.

\noindent\textbf{Video-language Foundation Models.} Inspired by the success with large language models~\citep{devlin2018bert,brown2020language}, video-language foundation models have been proposed by training with large amounts of image and video data with masked token prediction~\citep{fu2021violet} or video-task matching~\citep{paprika}. The videos are often accompanied by text descriptions such as speech narration~\citep{sun2019videobert,fu2021violet,zellers2021merlot,zellers2022merlot}.
When video and language are encoded separately, a contrastive learning objective can be employed~\citep{kevin2022egovlp,xu-etal-2021-videoclip} in a similar fashion as its image-based counterpart~\citep{Radford2021LearningTV}. The objectives can be combined~\citep{yu2022coca} and the encoders for different modalities can be shared~\citep{wang2022allinone}. In addition to joint multimodal pre-training, researchers have demonstrated the effectiveness of adapting visual descriptions into a large language model, such as with gated cross-attention~\citep{alayrac2022flamingo}, instructional tuning~\citep{liu2023visual, llavanextvideo, videollama2}, linearly projecting the visual embeddings into the language space~\citep{li2023blip,moon2023anymal}, or augmenting large language models with video frame captions~\citep{wang2023vamos,zhang2023simple}. Despite their amazing progress, we have shown that the state-of-the-art video-language models cannot generalize to Spacewalk-18 benchmark, whether zero-shot or with fine-tuning.

\noindent\textbf{Long-form Video Understanding} is an important open question for video representation learning. Earlier approaches~\citep{tapaswi2016movieqa} aim to extract high-level information such as character relationships from movies, where the solution is dominated by language-based approaches. Wu \etal~\citep{LVU} proposed an object-centric self-supervised framework, along with a benchmark for long-form video understanding. To encode long-form temporal context, memory-based approaches~\citep{wu2019long} have been proposed, along with new architectures that better scale with longer sequences~\citep{li2022mvitv2,islam2022long}. Our work is inspired by the recent dataset, EgoSchema~\citep{EgoSchema}, for question answering from long-form videos. Spacewalk-18 is complementary to EgoSchema, which is based on the same household videos as in Ego4D~\citep{Ego4D}. We also compute the temporal certificate~\citep{EgoSchema} to measure the context needed for humans to solve a video understanding task, and observe that our task requires 40\% more temporal context than EgoSchema.

\section{The Spacewalk-18 Dataset}
\label{sec:datacollection}

As shown in Figure~\ref{fig:tasks}, our goal is to segment spacewalk recordings into series of steps.
A typical spacewalk video contains an animated preview of the steps to be performed, followed by the recording of astronauts performing the steps over multiple hours. We aim to annotate the spacewalk into a sequence of these steps. Each step corresponds to multiple video clips from recording, denoted by start and end times. Each step is accompanied with a text description annotated by a human expert, and illustrated by a video clip from the animated preview. We obtained 18 spacewalk recordings from YouTube. The total duration is 96 hours. For each video, we chunk the audio files into 10 minute clips and feed them to Deepgram\footnote{\url{https://deepgram.com}} to extract speech transcripts.

\subsection{Labeling Process}

Building our dataset requires temporally segmenting and labeling very long videos. Due to the presence of video clips irrelevant to the spacewalk (\emph{e.g.}, scene of the mission control center
in Figure~\ref{fig:teaser}), the temporal boundaries of a single step can be fragmented, making annotating the temporal boundaries time-consuming and measuring the inter-annotator agreement challenging. We propose to over-segment the recordings into short clips, each of which corresponds to at most a single step. We then ask the annotators to label each clip from a pre-defined list of steps for a spacewalk, while having access to an unbounded-length context from its neighboring video clips. The human annotators are thus free from having to draw detailed temporal boundaries, or to take multiple passes over the long video for step definition, temporal segmentation, and step labeling. We observe that this design drastically reduced human worker hours.

\noindent\textbf{Define the Label Space.} Our label space is derived from the animated preview for each spacewalk. We manually segment the animations into steps ourselves, and label each segment with a step caption.
On average, this results in around 25 steps per spacewalk video.

\noindent\textbf{Create Video Clips.} We segment each video into short clips using PySceneDetect's\footnote{\url{https://github.com/Breakthrough/PySceneDetect}}
shot boundary detection algorithm.
We find that due to the long nature of the steps and tendency of the camera to switch angles often in spacewalk recordings, a given shot-segmented clip rarely spans over one step: we randomly inspect the annotated clips and observe that only 6\% of them contain more than one step. Overall, only 4\% of the frames are mislabeled due to shot boundaries. While imperfect, we believe the benefit of scaling up annotation significantly outweights the minimal impacts on evaluation metrics.

\noindent\textbf{Annotation Interface.} To collect annotations, we build the Spacewalk Video Annotation Tool (example interface is shown in \Cref{app:labeling_process}).
The interface displays the animated steps and clips for human annotators to label. They watch the annotation clips and select a label for each clip from the set of step labels, ``Irrelevant'', or ``Unsure''. This interface allows them to view the entire spacewalk recording as context. The ``Irrelevant'' label categorizes any clip that does not contain footage of one of the steps for the given spacewalk. This includes shots of the mission control center, noisy shots (\emph{e.g.}, blue screen), and shots of ``get-ahead'' steps that were not originally planned for the spacewalk. We have at least three human workers annotate each clip and we choose the most commonly selected label as the true label. Our annotation tool is publicly released.

\noindent\textbf{Merge Adjacent Clips.} Since we intentionally over-segment the long spacewalk recordings before collecting annotations, we include a final step to use the collected annotations to obtain true temporal boundaries by merging all adjacent clips with identical labels.

\noindent We provide \textbf{supplementary videos} to illustrate the Spacewalk videos and our collected annotations.

\subsection{Dataset Statistics, Diversity, and Difficulty}
The 18 spacewalk videos are segmented into in total 31,812 clips for human annotation.
After annotation, adjacent clips with the same step labels are merged as post-processing, resulting in 3,753 merged clips.
These clips have an average length of 92 seconds and span over 96 hours in total.
There are 455 animated step labels across the dataset (excluding ``Irrelevant''). Each step takes on average 5 merged clips and has an average of 9 minutes of video.
To analyze the diversity of the dataset, we filter the nouns and verbs from the step captions to obtain a list of objects and actions.
Our dataset has 51 diverse objects (\eg \textit{battery}) and 47 atomic actions (\eg \textit{install} and \textit{ingress}) (See \Cref{app:statistics}).

Spacewalk-18 is \textit{open-vocabulary}, as the list of step labels varies across different videos.
Example labels in the training set include ``EV1 \& EV2 remove the first battery'' and ``EV2 ingresses foot restraint''. In the test set, there are similar labels, such as ``Chris \& Bob remove new battery from slot A'' and ``Chris enters foot restraint''.

The visual content in Spacewalk-18 is out-of-domain for most existing models. The shots of astronauts working in space are visually distinct from any worldly actions. There is also a mix of first- and third-person point of view camera angles. Additionally, the two astronauts in each spacewalk often work on different steps in the same time, leading to steps appearing in multiple non-continuous segments.
To reduce the impact of task difficulty on annotation accuracy,
we offer training to annotators from online platform, and those who achieve at least 80\% accuracy on a held-out set after training are recruited to annotate the dataset.

\subsection{Task Definition on Spacewalk-18}
\label{sec:task_definition}
    The 18 recordings in Spacewalk-18 contain varying numbers of steps. To ensure each split contains a balanced number of steps, we manually split the dataset into training, validation, and test sets in a ratio of 10:2:6. We introduce two multimodal long-form video understanding tasks on Spacewalk-18, step recognition and question answering:

    \noindent\textbf{Task 1 - Step Recognition.}
        Our dataset divides the multi-hour spacewalk recording into several steps.
        Each step is described by an animated video clip, the transcript of its narration, and an annotated caption.
        Our step recognition task aims to recognize these steps from the spacewalk recording.
        Given a timestamp $t$ and the list of $K$ steps in the corresponding spacewalk mission, a model is tasked with determining the step occurring at timestamp $t$ by predicting a label from $\{0, 1, 2, \cdots, K\}$ (label $0$ stands for ``Irrelevant''). We notice that temporal contexts are crucial for recognizing the steps, from which we can see the astronauts' actions and know the completed and remaining steps. However, current video-language models are incapable of digesting hours-long spacewalk recordings. So we set a context window length $w$ and offer the video clip $\left[t-w/2, t+w/2\right]$ to the model. The clip includes both visual content and textual speech transcripts. We test models with varying $w$ to investigate their ability to understand temporal contexts.
        
        We construct 2000 samples from each training, validation, and test video, resulting in an average of 1 sample for every 10 seconds of video.
        As the visual content of spacewalk videos does not change rapidly, these samples are sufficient to represent an entire video.
        To balance the categories, $2000/(K+1)$ timestamps are uniformly sampled from each step's corresponding video clips.
        These timestamps are used as the middle timestamps $t$ in the task definition above across different context lengths.
        
        To evaluate model performance, we calculate Accuracy and mean Average Precision (mAP) to measure how accurately each sample is recognized. By merging temporally adjacent samples with the same predictions into intervals, we derive a temporal segmentation of each spacewalk mission. We also adopt the standard intersection-over-union (IoU) metric with the ground truth step boundary annotations to measure the segmentation correctness. We found that sampling more training or validation examples (thus denser in time) has little impact on the IoU metric, which are defined on continuous temporal boundaries. Detailed metric discussions can be found in \Cref{app:metrics}.

    \noindent\textbf{Task 2 - Question Answering.}
    For ease of evaluating video-language models on temporal understanding and reasoning with Spacewalk videos, we further introduce the question answering task. We first split each Spacewalk video into hour-long consecutive video segments, and then manually collect questions that are either high-level (\eg \textit{what is the goal of this mission}) or detailed and requires temporal localization and reasoning (\eg \textit{What type of equipment does the astronaut retrieve first, and how is it utilized during the mission}). Question answering is formulated as multiple-choice, where four candidates are provided. The questions and all possible choices are annotated by experienced annotators. To facilitate the question collection, we leverage the step annotations to automatically generate certain types of questions, such as ``\textit{What did EV1 do while EV2 did} \texttt{[task]}''. Overall, we collected 376 questions for testing a pre-trained VLM's zero-shot performance.

\subsection{Characteristics of Spacewalk-18}
\label{sec:tempcert}

Our benchmark requires semantic understanding and temporal reasoning abilities in a truly \textit{unique} domain. A comparable benchmark is the recently released Perception Test~\citep{PerceptionTest}, which is also designed to measure similar skills and generalization capabilities. Spacewalk-18 nicely complements benchmarks like Perception Test as they focus on different domains (daily life vs. space) and has comparable total video durations (74 vs. 96 hours).

\begin{figure}
    \centering
    \includegraphics[width=0.9\linewidth]{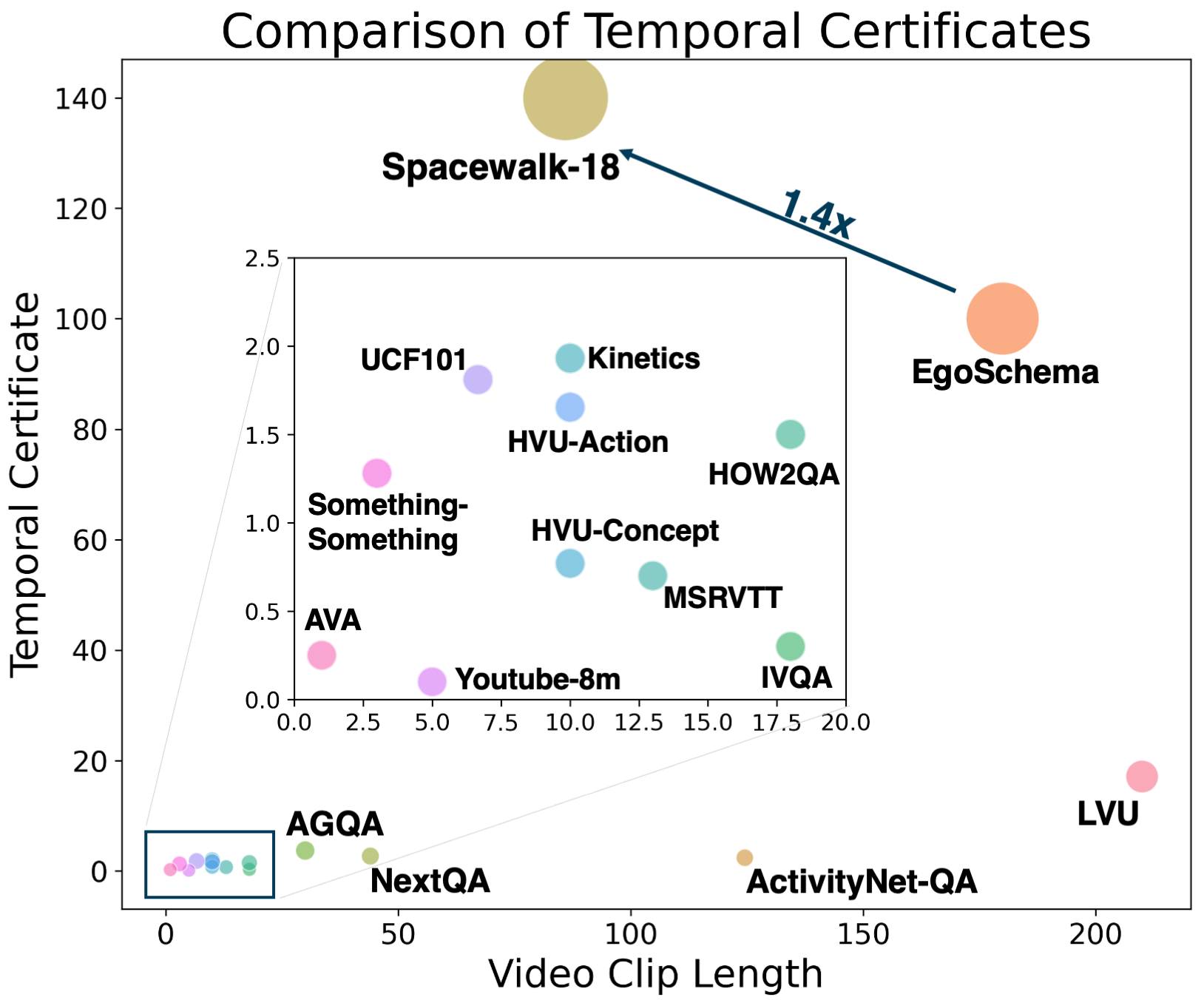}
    \vspace{-3pt}
    \caption{Temporal certificate (``long-form-ness'') lengths across commonly adopted datasets with action recognition and question answering annotations. Spacewalk-18 is 1.4x the length of the nearest comparable (EgoSchema). Figure adapted from \cite{EgoSchema}.}
    \label{fig:tempcert}
    \vspace{-15pt}
\end{figure}

\noindent\textbf{Long-form-ness:} Spacewalk-18 not only contains long videos, but also requires high amounts of context (long-form video understanding) in order to annotate. We use the temporal certificate \citep{EgoSchema} metric to quantify the long-form-ness of our video dataset. The temporal certificate measures the amount of video context required for a given dataset/task. Human verifiers are provided with an annotated video clip and are asked to select the amount of the video they require to be confident that the provided label is correct. EgoSchema~\citep{EgoSchema} defines benchmarks with certificate length around 1 second as short video tasks, 10 seconds as long-form, and 100 seconds as very long-form. 

To collect this metric, we split each clip into 5-second chunks and task human workers with selecting the minimum subset of these clips that they need to be confident in the given label.
In Figure~\ref{fig:tempcert}, we benchmark our dataset with eight annotators over 2.5 hours of spacewalk video. Our dataset has an average clip length of 89 seconds and a temporal certificate length of 140 seconds. This is 1.4x the length of the nearest dataset and places it in the category of ``very long-form video datasets'' \citep{EgoSchema}. 

\noindent\textbf{Multi-modality:} We also observe that the temporal understanding and reasoning tasks in Spacewalk-18 are inherently multimodal. Table~\ref{tab:modality_task1} shows the human performance when asked to solve the step recognition task, where they can freely explore the temporal context from the entire spacewalk video when needed. We can see that humans perform the best when both video and audio (speech transcipts) are available, with an accuracy of 67.0\%. This multimodal performance is higher than when humans only have access to video (52.2\%) or audio (39.1\%).

\begin{table}[t]
    \centering
        \scalebox{0.7}{
        \hspace{-15pt}
        \begin{tabular}{cccc}
            \toprule
            Benchmarks & Duration (h) & \# Annotations & Domain \\
            \midrule
            \textcolor{gray}{Step recog. (task 1)} & &\textcolor{gray}{Temporal Segments}  \\
            \midrule
            Ikea-FA~\cite{toyer2017human} & 4 & 2 k & Assembling\\
            Breakfast~\cite{Breakfast} & 77 & 8 k & Cooking \\
            YouCook2~\cite{zhou2018towards} & 176 & 14 k & Cooking \\
            EPIC-KITCHENS~\cite{damen2022rescaling} & 100 & 89 k & Cooking\\
            EgoPER~\cite{lee2024error} & 28 & 8 k & Cooking \\
            Assembly101~\cite{sener2022assembly101} & 513 & 105 k & Assembling\\
            EPIC-Tent~\cite{jang2019epictent} & 5 & 1 k & Assembling\\
            Spacewalk-18 (ours) & 96 & 4 k & Spacewalk\\
            \midrule
            \textcolor{gray}{Question ans. (task 2)} & & \textcolor{gray}{Questions}\\
            \midrule
            EgoSchema~\cite{EgoSchema} & 250 & 5 k  & Open-domain \\
            Video-MME-L~\cite{fu2024video} & 206 & 900 & Open-domain\\
            P. Test 1hr-walk~\cite{heyward2024perception} & 10 & 70 & Tours \\
            Spacewalk-18 (ours) & 96 & 376  & Spacewalk \\
            \bottomrule
        \end{tabular}
        }
        \vspace{-8pt}        
        \caption{Statistics comparison with single-domain step recognition benchmarks and recent long-form video QA benchmarks.}
        \vspace{-15pt}
    \label{tab:benchmark_comparison}
    \end{table}

\noindent\textbf{Comparison with Relevant Datasets:}  Table~\ref{tab:benchmark_comparison} compares Spacewalk-18 with other single-domain step recognition datasets. Our dataset has comparable total duration as relevant datasets, while focusing on a unique domain than cooking or assembling furniture. We also compare with recent long-form video QA benchmarks, which shows that our question size is reasonable as a single-domain benchmark.
\section{Recognition and Reasoning Models}
\label{sec:method}
    
    For step recognition, we benchmark both contrastive video-language models (contrastive VLMs) and video large language models (VLLMs). Contrastive VLMs can provide aligned video and text embeddings while VLLMs can process video and language simultaneously and perform video question answering.
    Both of them are evaluated in zero-shot scenario to demonstrate how pre-trained models generalize to Spacewalk-18.
    We also evaluate contrastive VLMs in fine-tuning scenario, which can effectively adapt pre-trained models to novel domains \citep{rosenfeld2022domain}. Both last-layer and all-layer fine-tuning are considered here.

    In Spacewalk-18, the animation of the $i$-th step is described by an animation video $V_i^a$, a transcript of its narration $T_i^a$, and a step caption $C_i^a$.
    A spacewalk recording clip centered at timestamp $t$ with length $w$ contains a video clip $V_{t,w}$ and a transcript $T_{t,w}$.
    When evaluating contrastive VLMs, we format recognition as retrieval, similar to how contrastive VLMs can be used for zero-shot classification.
    We employ separate video encoder $F_v(\cdot)$ and text encoder $F_t(\cdot)$ to derive clip features, and match the spacewalk recording clips with step animations.
    To evaluate VLLMs, we feed the videos and transcripts into the models and treat our tasks as multi-choice video question answering.
    Unless otherwise mentioned, all the models use Sparse Frame Sampling to process videos, where we uniformly sample $k$ frames from the entire clip $V_{t,w}$ regardless of the video length and feed them into the video encoder. $k$ is the number of frames used during model pre-training, which may differ between different models. For the question answering task, a model is given the entire hour-long input video, and it needs to search for the relevant evidence from the long-form input, making the task more challenging.

    \subsection{Step Recognition and QA by VLLMs}
        
        \noindent\textbf{Zero-shot.}
        We format the step recognition task as multi-choice video question answering as the following. 
        Given a spacewalk clip with $K$ steps in the mission, we provide the video $V_{t,w}$ and transcript $T_{t,w}$ to the model and ask ``\emph{Which step does the frame in the middle of this video belong to?}'' We also provide the caption $C_i^a$ and transcript $T_i^a$ of each step and require the model to choose a step index between $0$ and $K$. See our prompt in \Cref{app:vllm_prompts}. We follow the same strategy for the question answering task, with the only difference that the choice candidates are provided by each question. We focus on the zero-shot setup and report performance not only on state-of-the-art VLLMs, but also proprietary APIs, such as GPT-4o.

    \subsection{Step Recognition by Contrastive VLMs}
    \label{sec:method_step_recog}

        \noindent\textbf{Zero-shot.}
        To derive a feature for a step animation, we first extract the video, transcript, and caption features respectively, and then concatenate them. Formally, the feature of the $i$-th step is
            $f_i^a = [F_v(V_i^a),\; F_t(T_i^a),\; F_t(C_i^a)]$.
        To construct a feature for the ``Irrelevant'' category, we write a textual description of it and extract its text feature. The description is
        \emph{DES=“The mission control center, noisy shots (e.g. blue screen), or tasks not planned for the spacewalk.”}
        Hence the “Irrelevant” category feature is
            $f_0^a = [F_t(\text{DES}),\; F_t(\text{DES}),\; F_t(\text{DES})]$.
        For a recording clip, we concatenate the video and transcript features:
            $f^s = [F_v(V_{t,w}),\; F_t(T_{t,w}),\; F_t(T_{t,w})]$.
        Here, the transcript feature is repeated so that the feature dimensionality is the same as those of the animation features.
        
        To form a prediction, we compute the similarities $f^s\cdot f_i^a$ between a recording clip and an animation step, and pick the step with the highest similarity.

        \noindent\textbf{Last-layer Fine-tuning.}
        We fine-tune a linear layer upon the pre-trained models using the training set.
        Specifically, we freeze the models and train a linear layer $G_\theta(\cdot)$ mapping the spacewalk clip features to the animation features.
        We minimize the cross entropy loss during training, which is
        \begin{equation}
            \mathcal{L}=-\log\frac{\exp\left(G_\theta(f^s)\cdot f_y^a\right)}{\sum_{1\leq i\leq K}\exp\left(G_\theta(f^s)\cdot f_i^a\right)}.
        \end{equation}
        After training, instead of constructing or learning a feature for the ``Irrelevant'' category, we find a threshold $\tau$ and recognize all clips whose similarities to task steps are all below $\tau$ as ``Irrelevant''.
        This is more reasonable than an ``Irrelevant'' feature because this category contains a subspace formed by various concepts rather than a single concept.
        With the threshold $\tau$, we make predictions by
        \begin{equation}
            \hat y =\begin{cases}
                \arg\max_{1\leq i\leq K}G_\theta(f^s)\cdot f_i^a,&\max G_\theta(f^s)\cdot f_i^a\geq\tau,\\
                0, &\text{otherwise}.
                \end{cases}
        \end{equation}
        We find the $\tau$ with the best F1 score on the the validation set and use it in the test phase. In practice, this method performs better than constructing an ``Irrelevant'' feature through text descriptions.

        \noindent\textbf{All-layer Fine-tuning.} We fine-tune the entire backbone of a pre-trained model to encode spacewalk video clips. We freeze the animation features and use the same cross entropy loss to fine-tune the model. The threshold $\tau$ for the ``Irrelevant'' category is also set using the validation set.

        \noindent\textbf{Incorporating Longer Temporal Context.} Our Sparse Frame Sampling approach naturally incorporates long temporal context with the same computational cost, but may lose important details due to sampling. We therefore explore several alternatives to incorporate temporal context: \textbf{Dense Frame Sampling} at 1 FPS, which are directly encoded by the video encoder to obtain a single video embedding. This approach is bounded by GPU memory since the number of frames scales linearly with the context duration. \textbf{Long-term Feature Bank (LFB)}~\cite{wu2019long}, which first divides the long context into seconds-long segments and leverages a frozen video encoder to extract one video embedding for each segment with a frame sampling rate of 1 FPS. 
        
        We explore the following LFB variants following~\cite{wu2019long}: Average pooling over the query and all of the context features to form a single embedding (\textbf{LFB Avg}); Average pooling the history context and future context separately, and concatenate them together with the query embedding (\textbf{LFB Cat}); Learning to aggregate temporal context via non-local blocks~\cite{Wang_2018_CVPR} (\textbf{LFB NL}) or a two-layer Transformer encoder (\textbf{LFB TF}). More details can be found in \Cref{app:lfb}.
\section{Experiments}
\label{sec:experiments}

    Our experiments show that current video-language models perform significantly worse than humans on Spacewalk-18. We also demonstrate the importance of both vision and language on our tasks. Furthermore, the capability of LFB in incorporating temporal contexts is verified.

    \subsection{Evaluated Models}
        We evaluate contrastive VLMs and VLLMs on Spacewalk-18. For contrastive VLMs, we test \textbf{EgoVLP}~\citep{kevin2022egovlp}, \textbf{VideoCLIP}~\citep{xu-etal-2021-videoclip}, \textbf{InternVideo}~\citep{wang2022internvideo}, and \textbf{InternVideo2}~\citep{wang2024internvideo2}. For VLLMs, we test open-source \textbf{LLaVA-Next-Video}~\citep{llavanextvideo}, \textbf{VideoLLaMA2}~\citep{videollama2}, \textbf{LongVU}~\citep{shen2024longvu}, \textbf{Qwen2.5-VL}~\citep{bai2025qwen25vltechnicalreport}, and \textbf{InternVL3}~\citep{zhu2025internvl3exploringadvancedtraining}, and proprietary \textbf{GPT-4o} and \mbox{\textbf{GPT-5}}. Moreover, recent works \citep{wang2023vamos, zhang2023simple} solve VideoQA tasks effectively by feeding generated video frame captions into LLMs. Hence, we develop a \textbf{caption-enhanced LLM} with \texttt{LLaVA-1.5-13B}~\citep{liu2023improvedllava} as the video frame captioner and GPT-4o as the LLM reasoner. See the selected checkpoints and number of sampled frames in \Cref{app:evaluated_models}.

    \subsection{Human Performance Evaluation}
        On the step recognition task, we evaluate human performance with all our annotators.
        Each of them is assigned part of the video clips and has access to an unlimited video context.
        We measure the accuracy of each human performer and take their average as the human performance.
        Overall, they achieve an accuracy of 67\%.
        This can be viewed as the ``upperbound'' performance on step recognition.

    \begin{table}[t]
    \centering
        \scalebox{0.6}{
        \setlength{\tabcolsep}{3pt}
        \begin{tabular}{c|ccc|ccc|ccc|c}
            \toprule
            \multirow{3}{*}{Method} & \multicolumn{9}{c|}{\textbf{Step Recognition}} & \multirow{2}{*}{\textbf{QA}} \\
            \cmidrule(r){2-10}
            & \multicolumn{3}{c}{$w$ = 1 min} & \multicolumn{3}{c}{$w$ = 3 min} & \multicolumn{3}{c|}{$w$ = 5 min} \\
            \cmidrule(r){2-4}
            \cmidrule(r){5-7}
            \cmidrule(r){8-11}
            & Acc. & mAP & IoU & Acc. & mAP & IoU & Acc. & mAP & IoU & Acc. \\
            \midrule
            Random & 4.22 & - & 1.51 & 4.22 & - & 1.51 & 4.22 & - & 1.51 & 25.00 \\
            Human$^*$ & - & - & - & - & - & - & 67.0 & - & - & -\\
            \midrule
            \textcolor{gray}{Zero-shot} \\
            \midrule
            EgoVLP & 7.18 & 8.28 & 1.90 & 7.71 & 9.08 & 1.67 & 7.59 & 9.83 & 1.59 & - \\
            VideoCLIP & 8.00 & 10.24 & 2.12 & 6.58 & \textbf{11.11} & 2.65 & 7.85 & 10.68 & 2.38 & - \\
            InternVideo & 9.35 & \textbf{10.78} & 2.96 & 9.48 & 11.01 & 2.73 & 9.07 & \textbf{11.56} & 2.97 & - \\
            InternVideo2 & 6.33 & 9.38 & 1.61 & 5.64 & 9.05 & 1.53 & 7.08 & 9.72 & 1.63 & - \\
            LLaVA-Next-Video-34B & 10.35 & - & 3.77 & 13.71 & - & 5.07 & 13.82 & - & 4.92 & 29.52 \\
            VideoLLaMA2-7B & 9.34 & - & 2.88 & 14.37 & - & 5.45 & 17.32 & - & 6.28 & 31.65 \\
            LongVU-7B & 8.58 & - & 3.04 & 10.49 & - & 3.81 & 12.01 & - & 4.41 & 36.44 \\
            Qwen2.5-VL-7B & 13.33 & - & 4.54 & 19.79 & - & 7.62 & 22.99 & - & 8.52 & 33.78 \\
            InternVL3-8B & 16.07 & - & 6.19 & 21.72 & - & 8.83 & 25.75 & - & 10.79 & 31.65 \\
            Caption-enhanced LLM & \textbf{18.56} & - & \textbf{9.33} & \textbf{26.32} & - & \textbf{12.89} & 28.49 & - & 13.16 & 30.37 \\
            GPT-4o & 15.47 & - & 6.36 & 21.65 & - & 9.41 & 26.40 & - & 11.55 & 32.45 \\
            GPT-5 & - & - & - & - & - & - & \textbf{36.16} & - & \textbf{18.42} & \textbf{46.54} \\
            \midrule
            \textcolor{gray}{Last-layer Fine-tuning}\\
            \midrule
            EgoVLP & 6.34 & 8.92 & 2.17 & 9.68 & 10.66 & 3.03 & \textbf{10.21} & 10.86 & 3.18 & - \\
            VideoCLIP & 8.40 & 9.80 & 3.21 & 9.98 & 11.14 & 3.71 & 8.87 & 10.39 & 3.46 & - \\
            InternVideo & \textbf{10.12} & \textbf{12.17} & \textbf{4.02} & \textbf{11.13} & \textbf{12.68} & \textbf{4.04} & 10.08 & \textbf{12.53} & \textbf{4.04} & - \\
            InternVideo2 & 9.41 & 10.36 & 3.53 & 8.96 & 10.00 & 2.98 & 8.39 & 9.97 & 2.97 & - \\
            \midrule
            \textcolor{gray}{All-layer Fine-tuning}\\
            \midrule
            InternVideo & 13.21 & 11.77 & 4.60 & 13.34 & 12.93 & 4.54 & 12.93 & 13.09 & 4.63 & - \\
            \bottomrule
          \end{tabular}
        }
    \vspace{-8pt}
    \caption{Model performances on Spacewalk-18. GPT-5 performs the best among all the models on both step recognition and question answering tasks. Both last-layer and all-layer fine-tuning improve the performances of contrastive VLMs. $*$: Humans have access to unlimited context.}
    \label{tab:main_results_task1}
    \vspace{-10pt}
    \end{table}
    
    \subsection{Main Results}
        Table~\ref{tab:main_results_task1} shows the model performances on Spacewalk-18. Due to the high computational cost of all-layer fine-tuning, it is only employed on InternVideo.
        
        On the step recognition task, we report the results under a few context lengths and conduct a thorough exploration of it in Section~\ref{sec:context}.
        The best accuracy on this task is 25.75\% (InternVL3-8B) for open-source models and 36.15\% (GPT-5) for proprietary models. Their performances are far less than the human performance of $67\%$. This verifies that our task is reasonable for humans but challenging to models. 
        VLLMs outperform contrastive VLMs, and significantly when the context window is long.
        Among the contrastive VLMs, InternVideo performs the best. Both last-layer and all-layer fine-tuning can boost constrastive VLMs, and all-layer fine-tuning yields higher performance than last-layer fine-tuning. However, the improvements are marginal, indicating that adapting models to extremely rare domains remains challenging.

        Besides VLMs, we evaluate a video-only model, VideoMAE~\citep{tong2022videomae}, on the step recognition task.
        Since the task is open-vocabulary, we train an MLP to project the VideoMAE embeddings to the SentenceBert~\citep{reimers-2019-sentence-bert} text embeddings of the step captions. It achieves 10.34\% accuracy, 12.91\% mAP, and 2.97\% IoU with 5-minute context window length.

        On the question answering task,
        GPT-5 shows strong performance (46.54\%). The most performant open-source model is LongVU (36.44\%), which has token compression mechanism designed for long-form video understanding.
        In contrast, all other models outperform the random baseline marginally.
        For both tasks, we provide \textbf{qualitative results} including success and failure cases in \Cref{app:qual}.

    \subsection{Effect of Modality}
    \label{sec:modality}
    
        In Table~\ref{tab:modality_task1}, we ablate the input modality on the step recognition task using zero-shot video-language models. And we also measure the human performances in multimodal and unimodal cases.
        In most cases, human and open-source video-language models perform the best when both modalities are provided, demonstrating the inherently multimodal nature of our task.
        However, the proprietary VLLMs achieve the highest accuracy when only text is given, probably due to their strong language prior.
        Besides, we notice that humans can better use the visual inputs, while our models benefit more from the texts. We hypothesize that humans excel in extracting abstract concepts from videos and match them with other modalities, while it is difficult for video-language models.
        Moreover, while the overall performance of InternVideo is worse than VLLMs, it outperforms most of the models in the video-only setting, demonstrating its better visual generalization to rare visual domains.

    \begin{table}[t]
    \centering
        \scalebox{0.72}{
        \begin{tabular}{c|ccc|ccc}
            \toprule
            \multirow{3}{*}{Method} & \multicolumn{6}{c}{Accuracy} \\
            \cmidrule(r){2-7}
            & \multicolumn{3}{c}{$w$ = 1 min} & \multicolumn{3}{c}{$w$ = 5 min} \\
            \cmidrule(r){2-4}
            \cmidrule(r){5-7}
            & V & T & V+T & V & T & V+T \\
            \midrule
            Human$^*$ & - & - & - & 52.2 & 39.1 & 67.0 \\
            InternVideo & \textbf{7.84} & 9.15 & 9.35 & 6.26 & 9.68 & 9.07 \\
            LLaVA-Next-Video & 5.22 & 9.38 & 10.35 & 4.93 & 14.65 & 13.82 \\
            VideoLLaMA2 & 4.39 & 8.03 & 9.34 & 4.56 & 13.03 & 17.33 \\
            Qwen2.5-VL & 6.59 & 13.71 & 13.33 & 6.98 & 22.78 & 22.99 \\
            Caption-enhanced LLM & 5.48 & \textbf{18.92} & \textbf{18.56} & 5.13 & 31.03 & 28.49 \\
            GPT-4o & 6.86 & \textbf{18.92} & 15.48 & 6.68 & 31.03 & 26.40 \\
            GPT-5 & - & - & - & \textbf{13.00} & \textbf{39.18} & \textbf{36.16} \\
            \bottomrule
        \end{tabular}
        }
        \vspace{-8pt}
        \caption{Ablation about input modality on step recognition task. The models are evaluated in zero-shot scenario. V: video; T: text (captions and transcripts). *: Human has unlimited context and accesses transcripts in the form of audio.\vspace{-10pt}}
    \label{tab:modality_task1}
    \end{table}
        
            \begin{figure}[t]
                \centering
                \includegraphics[width=\linewidth]{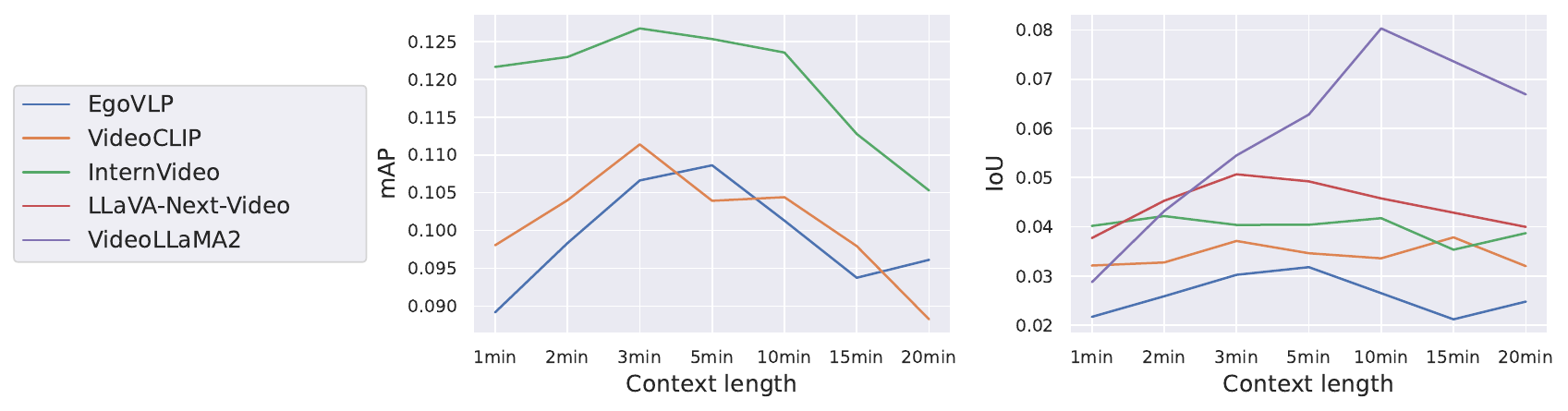}
                \vspace{-20pt}
                \caption{Ablation on context length. We test the models under various context lengths. Contrastive VLMs are last-layer fine-tuned while MLLMs are zero-shot.
                When the temporal context is extremely long, the models can no longer benefit from it.
                }
                \label{fig:context_length}
                \vspace{-10pt}
            \end{figure}

            \begin{figure}[t]
                \centering
                \includegraphics[width=\linewidth]{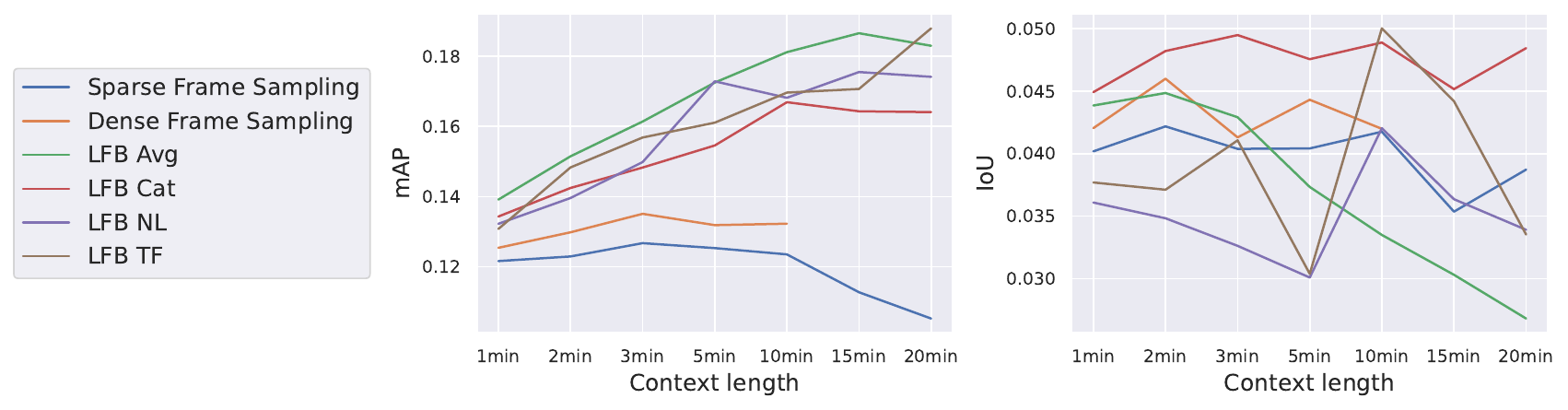}
                \vspace{-20pt}
                \caption{Performances of different temporal context incorporation methods built upon frozen InternVideo features. While LFB methods yield increasing mAP when the temporal context extends, one-time feed-forward models with either sparse or dense frame sampling cannot benefit from the context.
                }
                \label{fig:context_length_LFB}
                \vspace{-15pt}
            \end{figure}

    \subsection{Leveraging Temporal Context}
    \label{sec:context}

        As shown in Figure~\ref{fig:context_length}, unlike humans that have temporal certificate of 2.3 minutes, the open-source video-language models cannot benefit from very long context on both tasks.
        As context window length increases, their performances initially improve, but start to decline after reaching their peaks.
        
        To address this issue, we explore the temporal context incorporation approaches for contrastive VLMs introduced in Section~\ref{sec:method_step_recog} on the step recognition task. As Table~\ref{tab:modality_task1} shows that contrastive VLMs have the most robust visual capability to rare domains, it is more effective to explore video context incorporation using them than VLLMs. The performance curves with respect to context lengths are shown in Figure~\ref{fig:context_length_LFB}. The two one-time feed-forward methods -- Sparse/Dense Frame Sampling -- have similar performances, and they are not improved given expanded temporal context windows as well. This indicates that naively sampling more frames cannot help pre-trained video-language models to understand long-form videos.
        In contrast, the mAP curves of all the LFB methods show significant upward trends, verifying their ability to incorporate temporal contexts in long videos. Among them, LFB Cat, which concatenates past, present, and future video features, outperforms other methods on the accuracy and IoU metrics, demonstrating its effectiveness.

    \subsection{Adaptation via Summarization}

    We finally investigate how to effectively adapt a pre-trained model to solve Spacewalk-18. Our key inspiration is that for long-form, novel-domain videos, a model may adapt to and understand the videos by watching their summarizations.
    
    In the step recognition task, the full list of steps is available to the VLLMs.
    This list in fact serves as an oracle summarizing the overall spacewalk mission, which might contribute to their better performance than contrastive VLMs.
    To ablate its influence, we give the caption-enhanced LLM only one step each time and ask it to rate each step from 1 to 10.
    In \Cref{tab:adaptation}, the step oracle indeed improves its performance from $23.50\%$ to $28.49\%$, which is an even higher gain than fine-tuning the InternVideo.
    
    For the QA task, we simulate the summarization by either re-purposing the 5-minute animation video of each spacewalk, or by assuming a known list of steps that occur within the question video. These steps are represented by step captions and middle frames.
    In Table~\ref{tab:adaptation},
    the animation and step oracle provide significant improvements over GPT-4o. While both animation and step oracle should be considered as privileged information, they highlight the direction to adapt pre-trained models via \textit{condensed} knowledge.

    \begin{table}[t]
   
    \centering
    \scalebox{0.75}{
        \begin{tabular}{c|c|c|c}
            \toprule
            Task & Method & Summary & Accuracy (\%) \\
            \midrule
            & InternVideo & - & 9.48 \\
            Step& InternVideo (fine-tuned) & - & 13.34\\
            Recog.& Caption-enhanced LLM & - & 23.50 \\
            & Caption-enhanced LLM & Step Oracle & 28.49\\
            \midrule
            & GPT-4o & - & 32.45 \\
            QA & GPT-4o & Animation & 55.59 \\
            & GPT-4o & Step Oracle & 81.12 \\
            \bottomrule
        \end{tabular}
    }
    \vspace{-8pt}
    \caption{Adaptation via summarization significantly improves Spacewalk-18 performance on both tasks.}
     \label{tab:adaptation}
     \vspace{-15pt}
    \end{table}
\section{Conclusion}
\label{sec:conclusion}

We introduce the Spacewalk-18 benchmark to evaluate video-language models' capability to generalize to unseen domains, and their comprehension of multimodal information and long-term temporal context.
We demonstrate that while average human annotators achieve competitive performance on our benchmark, existing video-language models still struggle with domain generalization and long-form video understanding. Meanwhile, we discover that a promising direction for model adaptation is to provide video summary as its context, even without model fine-tuning.

\vspace{0.5em}
\noindent\textbf{Acknowledgements:}
This work is supported by NASA, Samsung Advanced Institute of Technology, and a Richard B. Salomon award for Chen Sun. We thank Karttikeya Mangalam and Raiymbek Akshulakov for their kind help with EgoSchema temporal certificate evaluation, and Professor Stefanie Tellex for her inspiration on looking into Spacewalk videos. Our research was conducted using computational resources at the Center for Computation and Visualization at Brown University.
{
    \small
    \bibliographystyle{ieeenat_fullname}
    \bibliography{main}
}
\clearpage
\setcounter{page}{1}
\setcounter{table}{0}
\renewcommand{\thetable}{A\arabic{table}}
\setcounter{figure}{0}
\renewcommand{\thefigure}{A\arabic{figure}}
\maketitlesupplementary
\appendix

\addcontentsline{toc}{section}{Appendix} 
\part{Appendix} 
\parttoc 

\vspace{5em}
\section{Dataset Construction}
\label{app:data_construct}

\noindent\textbf{Sourcing Live Streams.} First, a list of spacewalks at the International Space Station is sourced from \href{https://www.nasa.gov/international-space-station/space-station-spacewalks/}{NASA’s website}. For each spacewalk from 2019 to 2023, we identify whether there is a recording available on YouTube with an animation sequence near the beginning. We find a total of 18 spacewalk recordings that meet this criteria. Table \ref{tab:videolinks} has a list of the NASA summary for each expedition and the duration of each YouTube video.

\begin{table}[t]
\centering
\scalebox{0.9}{
\begin{tabular}{lcc}
\toprule
Expedition Date & Video Duration & Annotated Duration \\
\midrule
\multicolumn{3}{c}{\textcolor{brown}{Training set}} \\
\href{https://go.nasa.gov/30M7D0o}{October 6, 2019} & 8:42:06 & 5:25:23\\
\href{https://go.nasa.gov/2XCbsp7}{November 22, 2019} & 8:58:45 & 6:10:02 \\
\href{https://go.nasa.gov/2QX5Kwj}{January 15, 2020} & 9:03:04 & 6:26:18 \\
\href{http://go.nasa.gov/36okAAb}{January 25, 2020} & 7:59:43 & 5:22:10 \\
\href{https://go.nasa.gov/36kmo01}{January 27, 2021} & 8:25:35 & 5:55:27 \\
\href{https://blogs.nasa.gov/spacestation/2021/02/01/spacewalkers-wrap-up-battery-work-and-camera-installations/}{February 1, 2021} & 8:39:50 & 0:48:35 \\
\href{https://go.nasa.gov/3q0ZOSF}{June 16, 2021} & 9:23:13 & 6:35:01 \\
\href{https://go.nasa.gov/3DpeOif}{December 2, 2021} & 7:46:53 & 6:13:30 \\
\href{https://blogs.nasa.gov/spacestation/2022/03/23/astronauts-complete-spacewalk-to-install-station-upgrades/}{March 23, 2022} & 8:29:54 & 5:14:27 \\
\href{https://blogs.nasa.gov/spacestation/2022/12/03/spacewalkers-complete-new-solar-array-installation-on-station/}{December 3, 2022} & 7:44:35 & 2:12:00\\
\midrule
\multicolumn{3}{c}{\textcolor{brown}{Validation set}} \\
\href{https://go.nasa.gov/3MT4SDB}{March 15, 2022} & 9:14:30 & 5:42:27 \\
\href{https://blogs.nasa.gov/spacestation/2022/11/15/nasa-astronauts-complete-spacewalk-to-prep-for-upcoming-solar-array-upgrades/}{November 15, 2022} & 10:38:40 & 7:02:55 \\
\midrule
\multicolumn{3}{c}{\textcolor{brown}{Test set}} \\
\href{https://blogs.nasa.gov/spacestation/2019/11/15/spacewalkers-complete-first-excursion-to-repair-cosmic-particle-detector/}{November 15, 2019} & 8:25:23 & 3:50:47 \\
\href{https://blogs.nasa.gov/spacestation/2019/12/02/astronauts-wrap-up-third-spacewalk-for-cosmic-particle-detector-repairs/}{December 2, 2019} & 7:49:37 & 5:33:52 \\
\href{https://blogs.nasa.gov/spacestation/2020/06/26/cassidy-and-behnken-conclude-spacewalk-to-replace-batteries/}{June 26, 2020} & 8:14:13 & 4:08:36 \\
\href{https://blogs.nasa.gov/spacestation/2021/02/28/spacewalkers-conclude-todays-spacewalk/}{February 28, 2021} & 9:28:07 & 7:03:46 \\
\href{https://go.nasa.gov/2YGhFVq}{September 12, 2021} & 8:44:29 & 6:50:19 \\
\href{https://blogs.nasa.gov/spacestation/2023/06/09/nasa-spacewalkers-complete-solar-array-installation/}{June 9, 2023} & 8:12:25 & 5:46:47 \\
\bottomrule
\end{tabular}
}
\vspace{-0.05in}
\caption{The Spacewalk recordings used in the Spacewalk-18 dataset, and their video durations. The recordings cover the spacewalk missions from 2019 to 2023. Because each spacewalk recording includes scenes from both before the spacewalk begins and after it ends, only the video segments where the spacewalk activity happens are annotated. Moreover, some spacewalk missions (\eg, February 1, 2021 and November 15, 2019) may deviate from the tasks originally planned in the animation and carry out unplanned tasks during execution. We only annotated the parts that are aligned with the planned tasks.
}
\label{tab:videolinks}
\end{table}

\noindent\textbf{Building Transcripts.} We use Deepgram’s Automatic Speech Recognition (ASR) service to transcribe the spacewalk videos. In order to choose an ASR algorithm, we had a human verifier analyze transcriptions from Deepgram, OpenAI’s Whisper, NVIDIA’s NeMo, YouTube’s Auto Generated Captions, and Google’s Speech-to-Text on the same spacewalk audio clip. Deepgram proved to be the most accurate. Through a similar process, we found that Deepgram’s ASR algorithm performs the best with approximately 10 minute audio clips. 
Therefore, we chunk the multi-hour audio files into 10 minute clips and feed them to Deepgram for transcription. This results in a list of sentences with start and end timestamps for each spacewalk video.

\subsection{Labeling Process}
\label{app:labeling_process}
Building our dataset requires temporally segmenting and labeling very long videos (many hours). To do this, we introduce a new annotation protocol and tool. Existing methods of collecting temporal segment annotations require multiple passes and/or for clips to be pre-labeled \citep{ActivityNet, vondrick2010annotation}. While annotating our dataset would traditionally require three passes (action identification, temporal segmentation, and task labeling), we pre-segment the videos into clips containing a maximum of one step each, allowing us to collect annotations in a single pass. This drastically reduces the number of human worker hours required. The source code for the tool will be publicly released. The protocol is as follows:

\noindent\textbf{Define the Label Space.} In our case, the label space comes from the animation videos. We manually segment the animations into clips containing a single step and label each clip with a short description. We show the step lists of three spacewalk missions in Table~\ref{tab:steps1}, \ref{tab:steps3}, and \ref{tab:steps2}, one from each of the training, validation, and test sets.

\begin{table}[t]
\centering
\scalebox{0.9}{
\begin{tabular}{c|p{6.5cm}}
\toprule
Step ID & Caption \\
\midrule
1 & EV1 and EV2 exit airlock \\
2 & EV1 heads outward and places safety tether anchors \\
3 & EV2 retrieves foot restraint \\
4 & EV1 goes to carrier with solar array \\
5 & EV1 and EV2 drop of PGT and bags \\
6 & EV2 stows foot restraint and bag \\
7 & EV1 preps iROSA for removal \\
8 & EV2 installs bag and tools on mod kit \\
9 & EV2 sets up cables for future installation \\
10 & EV2 retrieves PGT and goes to EV1 \\
11 & EV1 does more prep for iROSA removal \\
12 & EV1 retrieves foot restraint from CETA cart \\
13 & EV1 installs, sets up, and enters foot restraint \\
14 & EV1 removes bolts on iROSA \\
15 & Robotic arm moves EV1 into position \\
16 & EV2 gets into position for iROSA release \\
17 & EV2 prepares iROSA for release \\
18 & EV1 removes iROSA \\
19 & EV2 stows tools and enters foot restraint \\
20 & EV1 carries iROSA on robotic arm to EV2 \\
21 & EV1 hands iROSA to EV2 \\
22 & EV1 exits foot restraint and goes to EV2 \\
23 & EV2 rotates into position \\
24 & EV1 enters foot restraint and receives iROSA \\
25 & EV2 exits foot restraint and gets into position \\
26 & EV1 and EV2 install iROSA \\
27 & EV1 and EV2 swing iROSA into single tube \\
28 & EV1 and EV2 drive mounting bolts \\
29 & EV1 and EV2 clean up and prep for next EVA \\
30 & EV1 and EV2 return to airlock \\
31 & EV1 and EV2 enter airlock \\
\bottomrule
\end{tabular}
}
\vspace{-0.04in}
\caption{Step list of the spacewalk mission on June 16, 2021, which is in the training set.}
\label{tab:steps1}
\vspace{2pt}
\end{table}

\begin{table}[h!]
\centering
\scalebox{0.9}{
\begin{tabular}{c|p{6.5cm}}
\toprule
Step ID & Caption \\
\midrule
1 & Luca and Drew exit airlock with pump system \\
2 & Luca and Drew take pump system to external support platform 2 \\
3 & Luca enter foot restraint on robotic arm \\
4 & Drew hands Luca the pump system \\
5 & Robotic arm takes Luca to AMS \\
6 & Drew move to ELC 2 \\
7 & Luca and Drew install pump system \\
8 & Luca and Drew connect power and data cables \\
9 & Robotic arm takes Luca to aft side \\
10 & Luca connect six fluid connections \\
11 & Robotic arm takes Luca to underside of AMS \\
12 & Luca and Drew complete final 2 suages \\
13 & Robotic arm takes Luca to ESP 2 \\
14 & Drew bring bags back to airlock \\
\bottomrule
\end{tabular}
}
\vspace{-0.05in}
\caption{Step list of the spacewalk mission on December 2, 2019, which is in the test set.}
\label{tab:steps3}
\end{table}

\noindent\textbf{Split the Videos.} To reduce the burden of temporal segmentation, we split the long live streams into sub-clips that each contain at most one step. We find that due to the long nature of the steps and tendency of the camera to switch angles often in spacewalk live streams, a given shot-segmented clip (between camera angle changes) will contain at most one action/step. Thus, to split the multi-hour long spacewalk video into clips for the human workers to annotate, we employ a shot detection algorithm. PySceneDetect’s ContentDetector uses changes in color and intensity between individual frames to draw boundaries between shot changes in videos. We find that it errs on the side of over-segmenting for our spacewalk live stream videos, further ensuring that clips will contain at most one step.

\noindent\textbf{Chunk into Segments.} It is unreasonable to expect human workers to annotate a multi-hour video in one sitting. Therefore, we chunk the videos into segments of approximately one hour. We design these segments to contain about one hour of recording content and only contain steps from a continuous subset of the animation. Each video has 5.11 segments on average.

\begin{figure*}[t]
    \centering
    \includegraphics[width=0.8\linewidth]{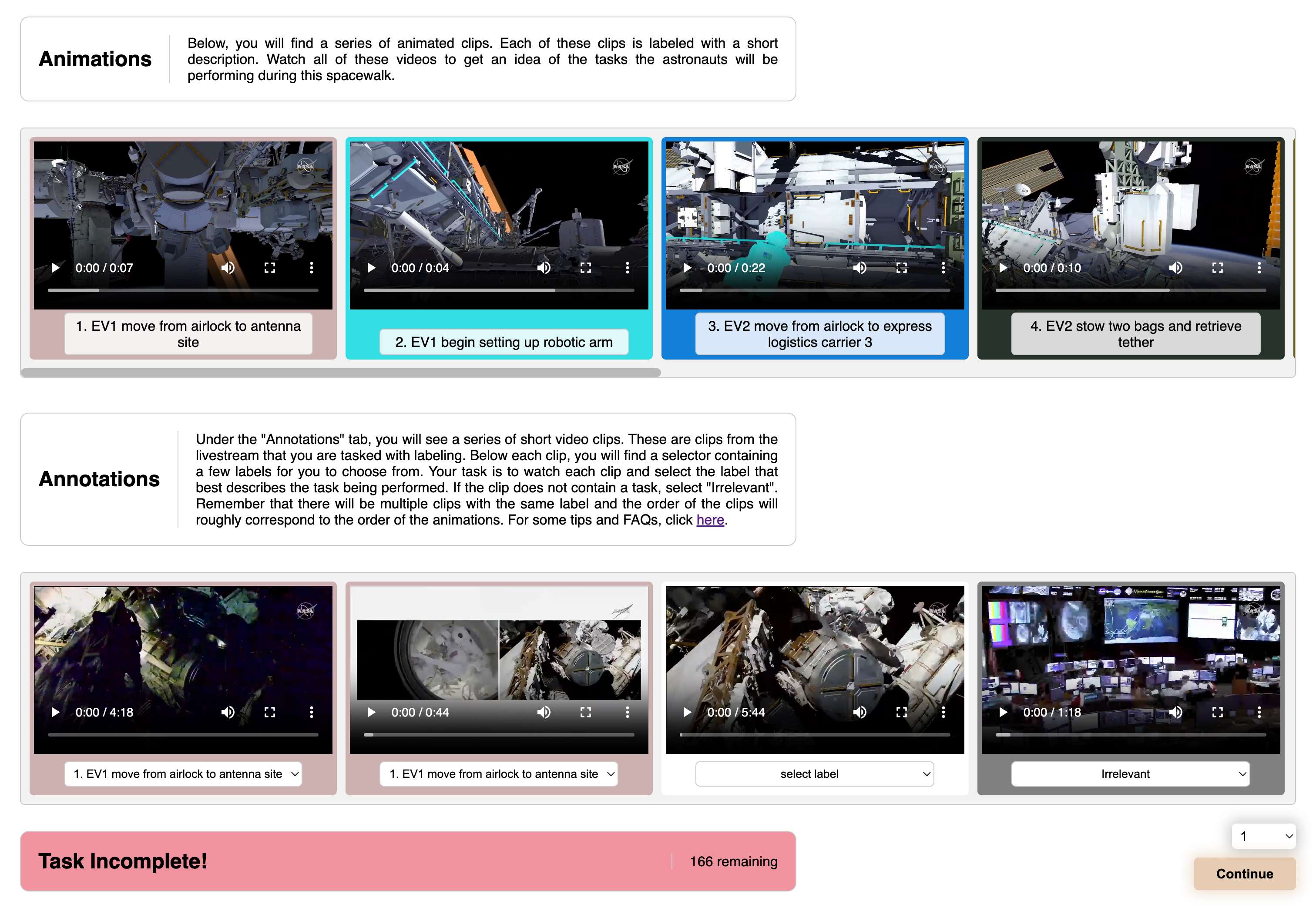}
    \caption{Spacewalk Video Annotation Tool Interface. 
    Row (1) contains a pre-labeled set of steps from the animation video. 
    Row (2) contains a set of live stream clips for the annotator to categorize into steps.}
    \label{fig:interface}
\end{figure*}

\noindent\textbf{Annotation Interface.} To collect annotations, we build the Spacewalk Video Annotation Tool, pictured in Figure~\ref{fig:interface}. The human worker inputs their unique user ID and selects the video date and segment number that they are assigned. The interface then loads the corresponding pre-labeled animation clips and the recording clips for them to label. The platform saves their progress as they annotate clips so they are able to start the task and return to it at any time. Once they complete the task, a completion URL appears.

\noindent\textbf{Source Human Annotators.} We use an online platform to source human workers to annotate the dataset. First, participants are sent the training and screening phases where they learn about the task and demonstrate proficiency in being able to accurately label a small sample of spacewalk live stream clips. In the training phase, as workers select labels for live stream clips, they are given feedback about the correctness of their annotations along with some reasoning. They spend as much time as necessary in the training phase before moving on to the screening phase. In this phase, workers are tasked with labeling a set of 10 clips without feedback, and those who achieve an accuracy of 80\% or greater are selected for the annotation phase.

\noindent\textbf{Annotate Clips.} In the annotation phase, human workers are presented with animation and annotation clips from a single segment of a spacewalk recording (Figure \ref{fig:interface}). They watch the annotation clips and select a label for each clip from the set of animation clip labels, ``Irrelevant'', or ``Unsure''. The ``Irrelevant'' label is used to categorize any clip that does not contain footage of one of the tasks for the given spacewalk. This includes shots of the mission control center, noisy shots (\emph{e.g.} blue screen), and shots of get-ahead tasks that were not originally planned for the spacewalk. We have three human workers annotate each clip and we choose the most commonly selected label as the true label.

\begin{figure}[ht]
    \centering
    \includegraphics[width=\linewidth]{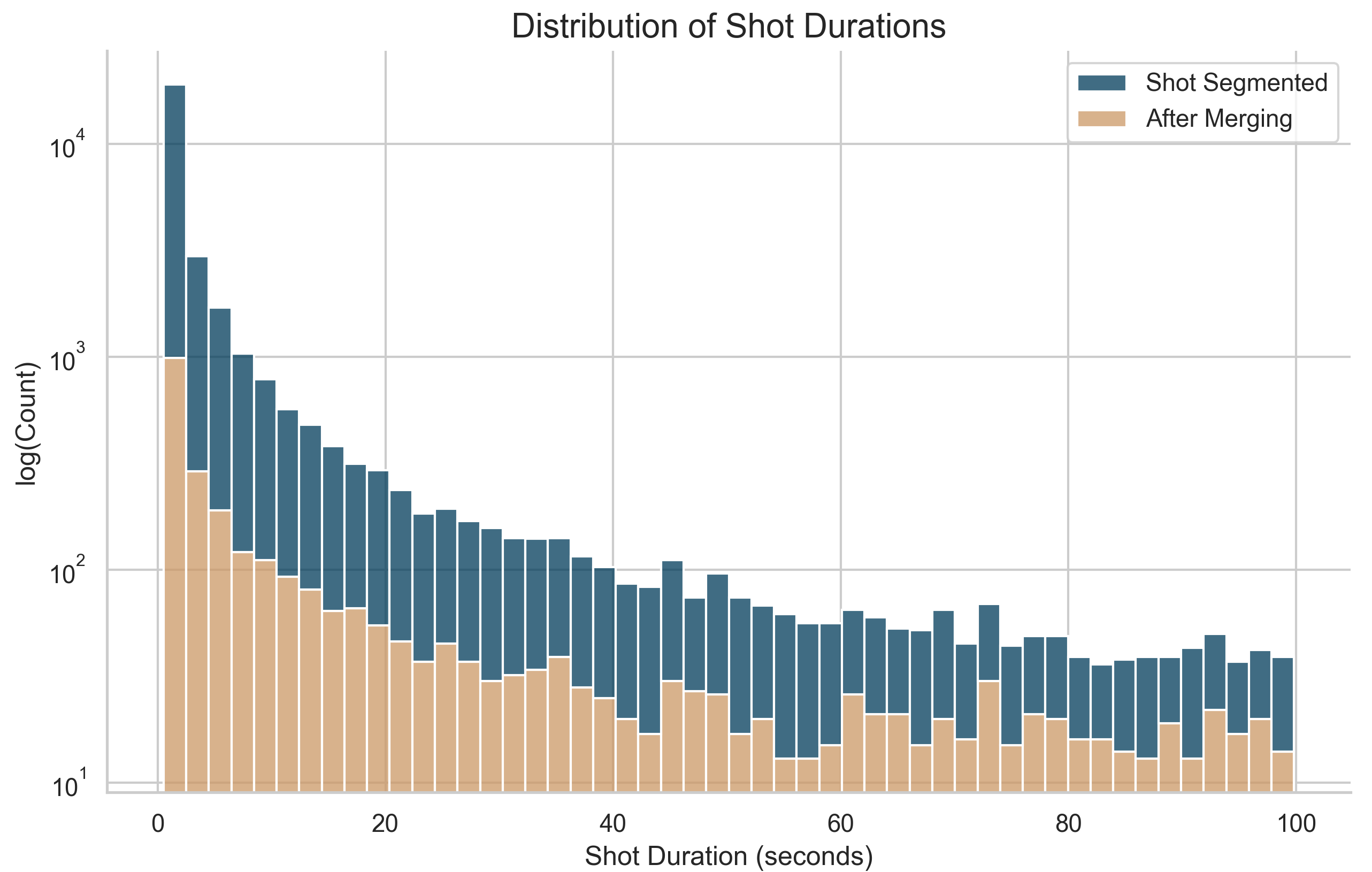}
    \caption{Distributions of video segment durations before and after merging adjacent clips.}
    \label{fig:shothist}
\end{figure}

\noindent\textbf{Merge Adjacent Clips. } We intentionally over-segment the long spacewalk recordings before collecting annotations, to avoid the same clip spanning across multiple steps. We thus include a final step to merge adjacent clips with the same label. Rather than the traditional method of having human workers provide temporal boundaries, the over-segmentation allows us to break down the challenging tasks of temporal segmentation and action recognition into a series of easier, smaller ones. We then use the collected annotations to obtain true temporal boundaries by concatenating all adjacent clips with identical labels. Figure \ref{fig:shothist} illustrates how this balances the distribution of clip durations in the dataset.

\subsection{Illustrations of Annotated Data}
\label{app:data_illustration}
\begin{figure*}[t]
    \centering
    \vspace{0pt}
    \includegraphics[width=0.8\linewidth,page=1]{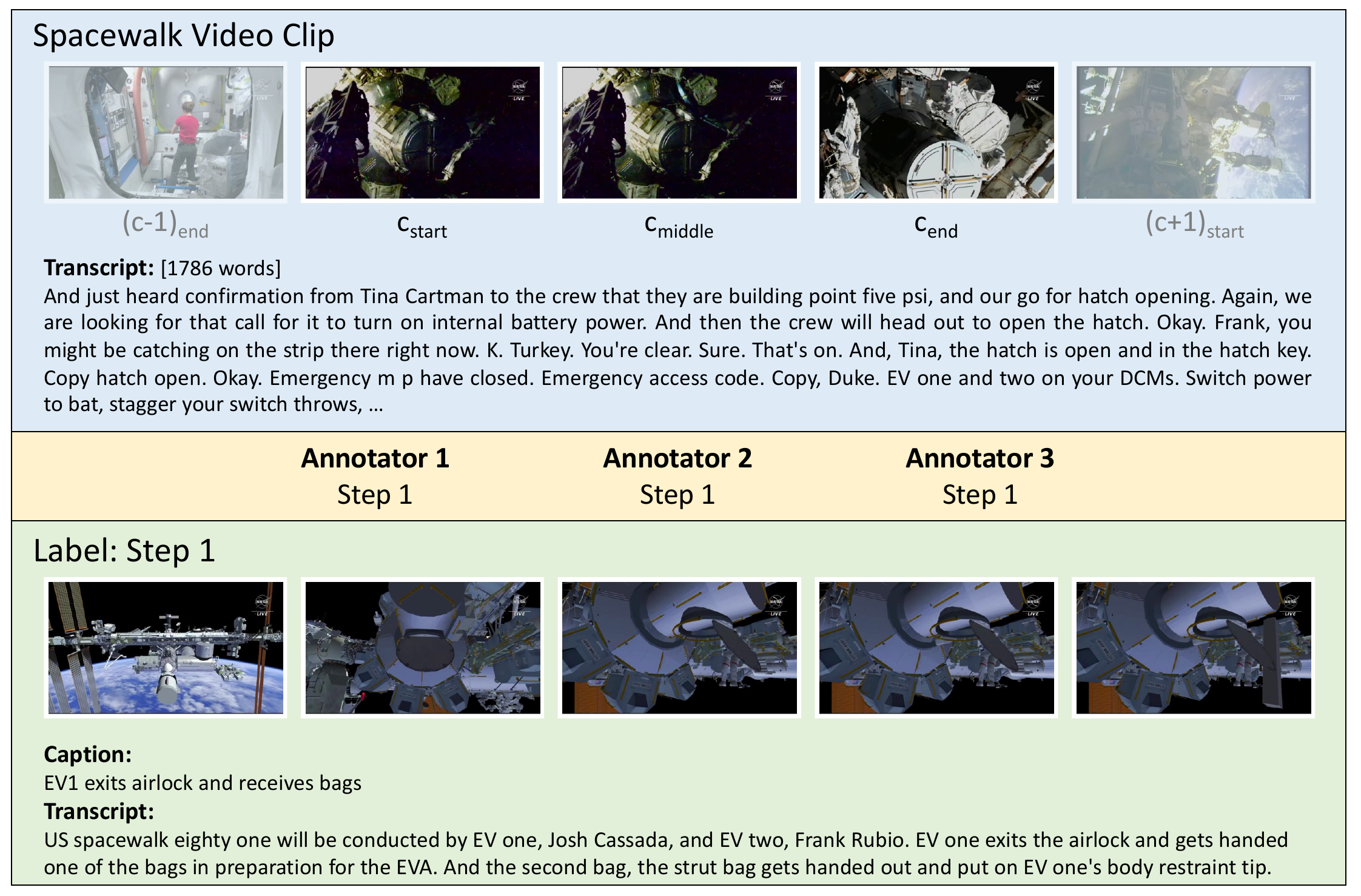}
    \caption{Annotation example 1 from the spacewalk recording from Nov. 15, 2022 (training set). All three annotators agree on the label for the video clip.}
    \label{fig:11152022_example_1}
    \vspace{-10pt}
\end{figure*}

\begin{figure*}
    \centering
    \vspace{10pt}
    \includegraphics[width=0.8\linewidth,page=2]{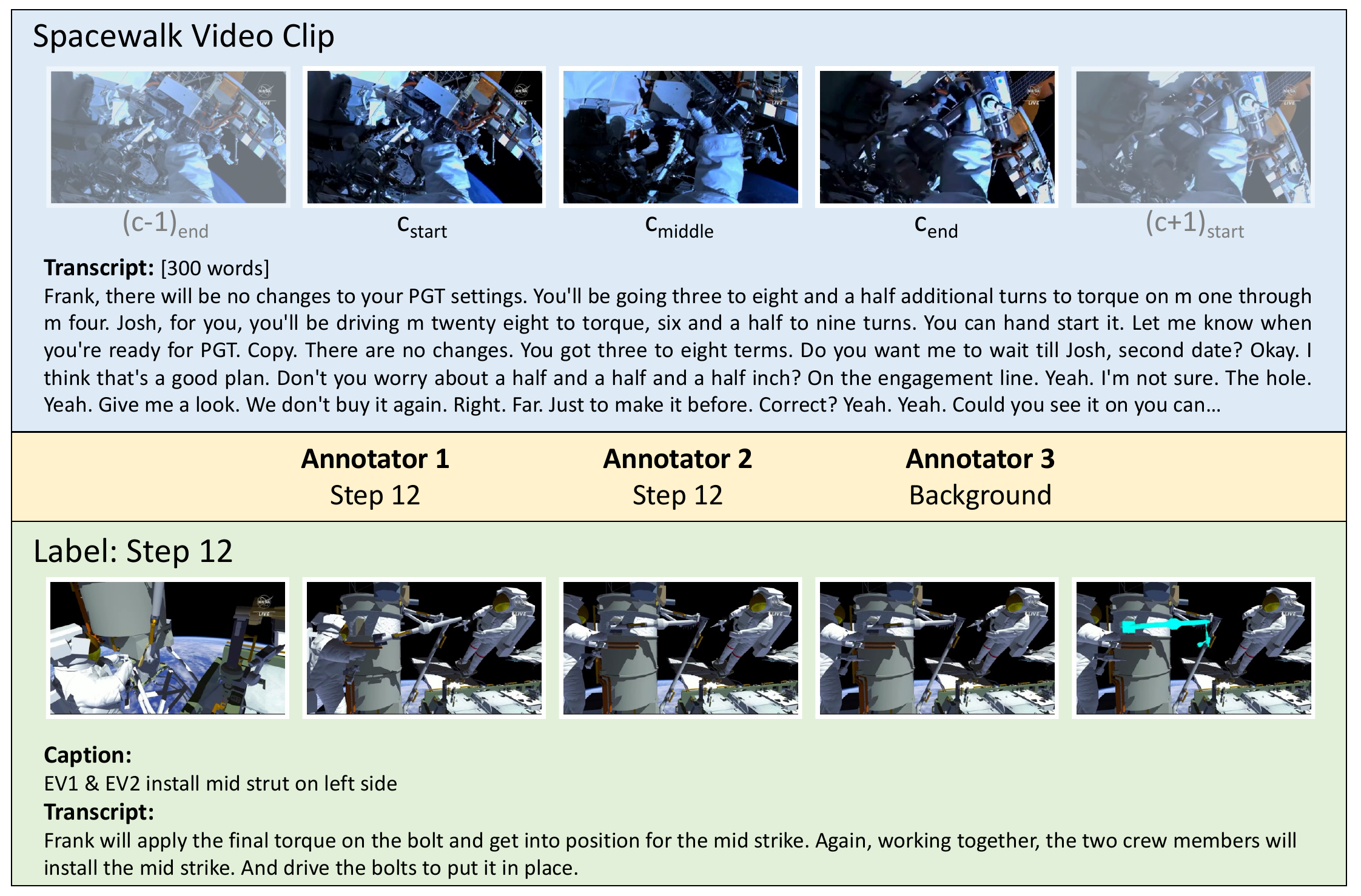}
    \caption{Annotation example 2 from the spacewalk recording from Nov. 15, 2022 (training set). One annotator disagrees with the other two but since the majority of annotators selected step 12, the clip is labeled as step 12. The clips on either side appear to be of the same step, which illustrates the effect of over-segmenting the spacewalk recordings and the necessity of merging adjacent clips after collecting annotations.}
    \label{fig:11152022_example_2}
    \vspace{-10pt}
\end{figure*}

\begin{figure*}
    \centering
    \includegraphics[width=0.8\linewidth,page=1]{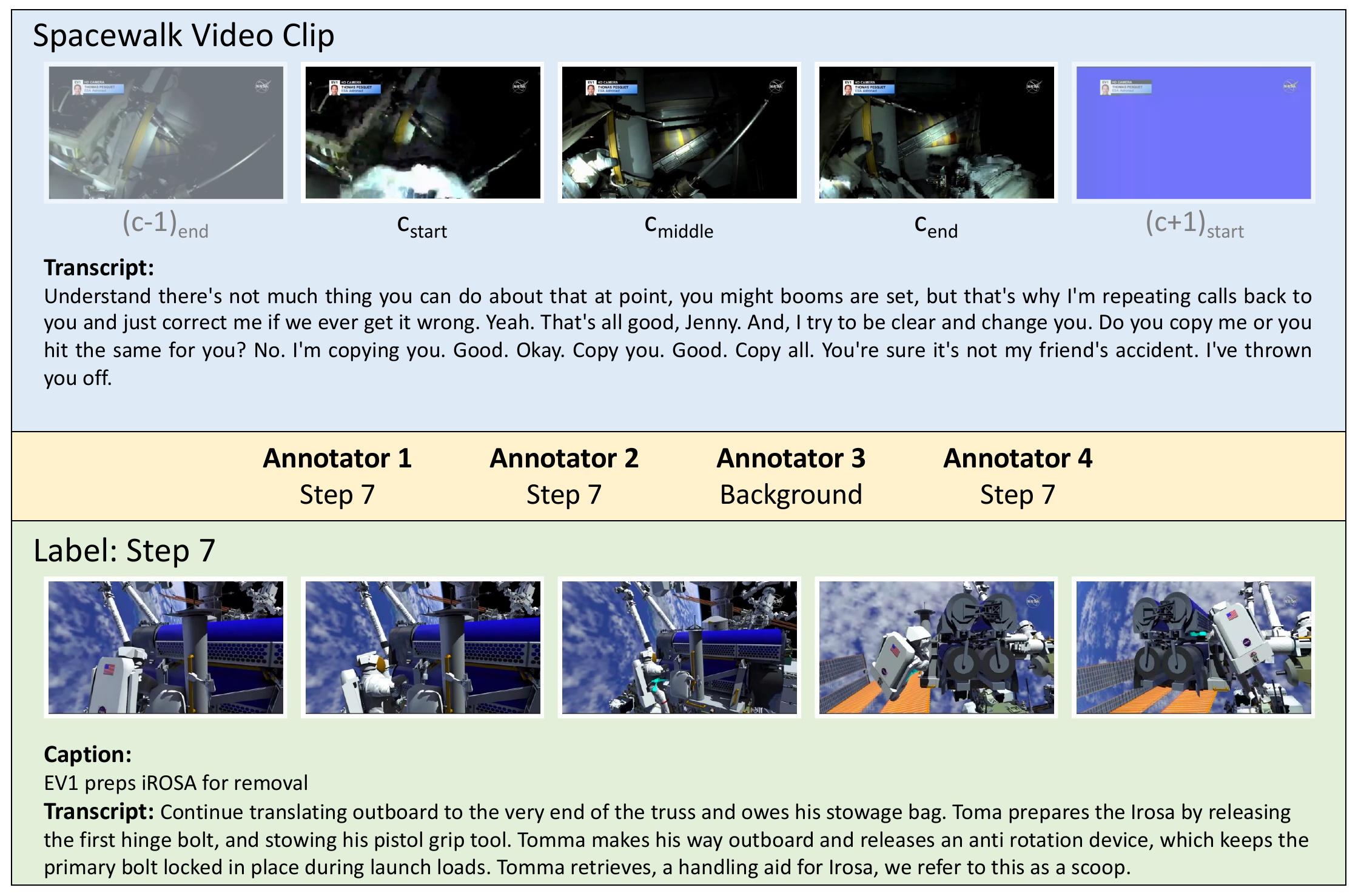}
    \caption{Annotation example 1 from the spacewalk recording from Jun. 16, 2021 (validation set). One annotator disagrees with the other three but since the majority of annotators selected step 7, the clip is labeled as step 7. The clip immediately following this one demonstrates an example of a blue screen that would be classified as ``Background''.}
    \label{fig:06162021_example_1}
    \vspace{-10pt}
\end{figure*}

Figure~\ref{fig:11152022_example_1}, \ref{fig:11152022_example_2}, and \ref{fig:06162021_example_1} illustrate three examples of annotations collected from human annotators.

\begin{table}[ht]
\centering
\scalebox{0.9}{
\begin{tabular}{c|p{6.5cm}}
\toprule
Step ID & Caption \\
\midrule
1 & EV1 exit airlock and receive large bag \\
2 & EV2 exit airlock \\
3 & EV1 move to integrated equipment assembly \\
4 & EV1 stow bag and begin prep work \\
5 & EV2 move to phase 1 \\
6 & EV2 stow crew bag and retrieve tools \\
7 & EV1 and EV2 assemble upper triangle \\
8 & EV2 move to and enter foot restraint \\
9 & EV1 hand upper triangle to EV2 \\
10 & EV2 dock upper triangle on gimbal assembly \\
11 & EV1 stow pistol grip tool (PGT) \\
12 & EV2 exit foot restraint and tilt it to the left side \\
13 & EV1 pass left mid strut to EV2 \\
14 & EV1 hands lower strut to EV2 \\
15 & EV1 and EV2 install lower strut \\
16 & EV1 and EV2 install mid strut \\
17 & EV2 exit foot restraint and tilt it to the right side \\
18 & EV1 hand right mid strut to EV2 \\
19 & EV1 and EV2 install lower strut \\
20 & EV1 and EV2 install mid strut \\
21 & EV1 finish mid strut install \\
22 & EV2 move to bag and stow tools \\
23 & EV2 return to worksite and stow tools on body restraint tether \\
24 & EV1 take pictures of completed mod kit \\
25 & EV1 move to battery charge/discharge unit and begin prep work \\
26 & EV2 translate to CETA cart and stow tools \\
27 & EV2 retreive crew lock bag and move to EV1 \\
28 & EV1 and EV2 fold and restrain insulation \\
29 & EV1 and EV2 break torque and reinstall \\
30 & EV1 and EV2 clean up and retrieve crew bag \\
31 & EV2 return to airlock \\
32 & EV1 return to airlock \\
\bottomrule
\end{tabular}
}
\vspace{-0.05in}
\caption{Step list of the spacewalk mission on March 15, 2022, which is in the validation set.}
\label{tab:steps2}
\end{table}

    \begin{figure*}[t]
        \centering
    \subfloat[Merged Video Clip Duration.]{
        \label{fig:dist_clip_duration}
        \scalebox{0.33}{
        \includegraphics[width=0.94\linewidth]{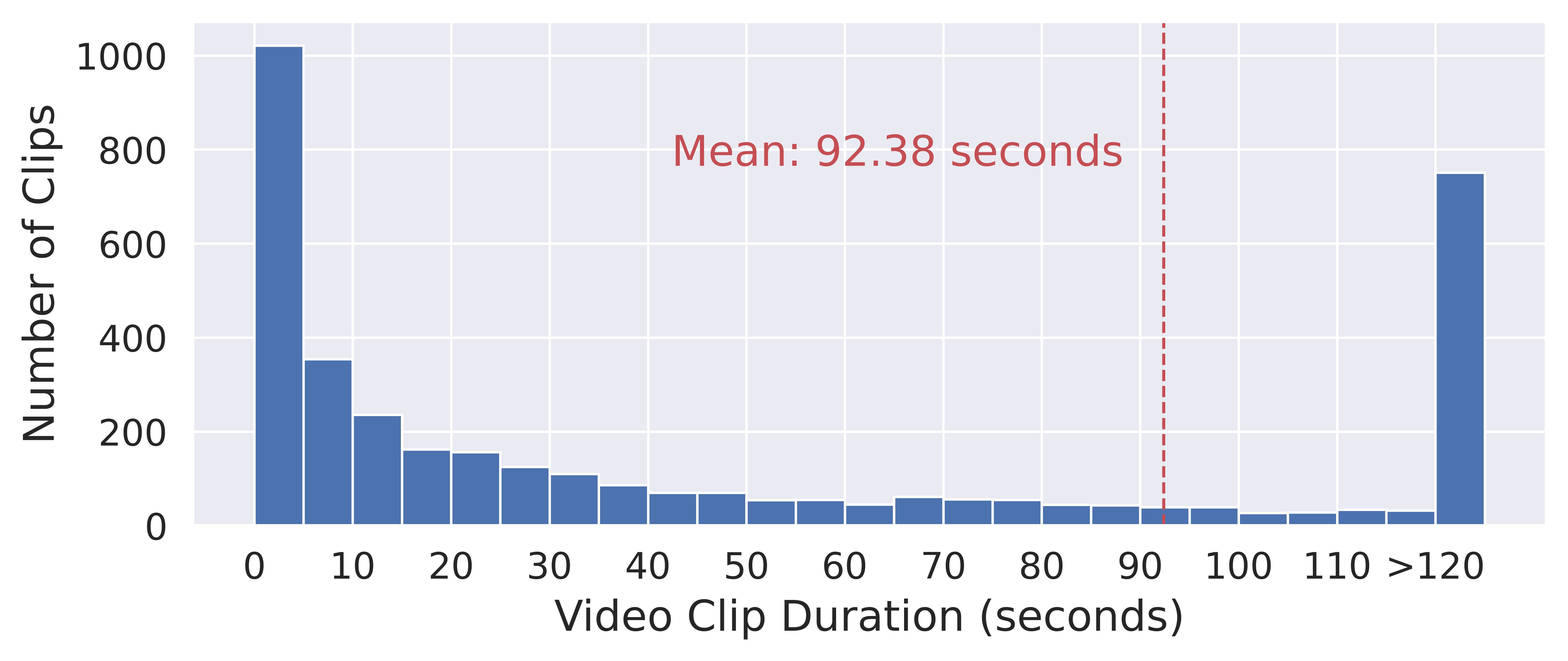}
        }
    }
    \subfloat[Total Step Duration.]{
        \label{fig:dist_step_duration}
        \scalebox{0.33}{
        \includegraphics[width=0.94\linewidth]{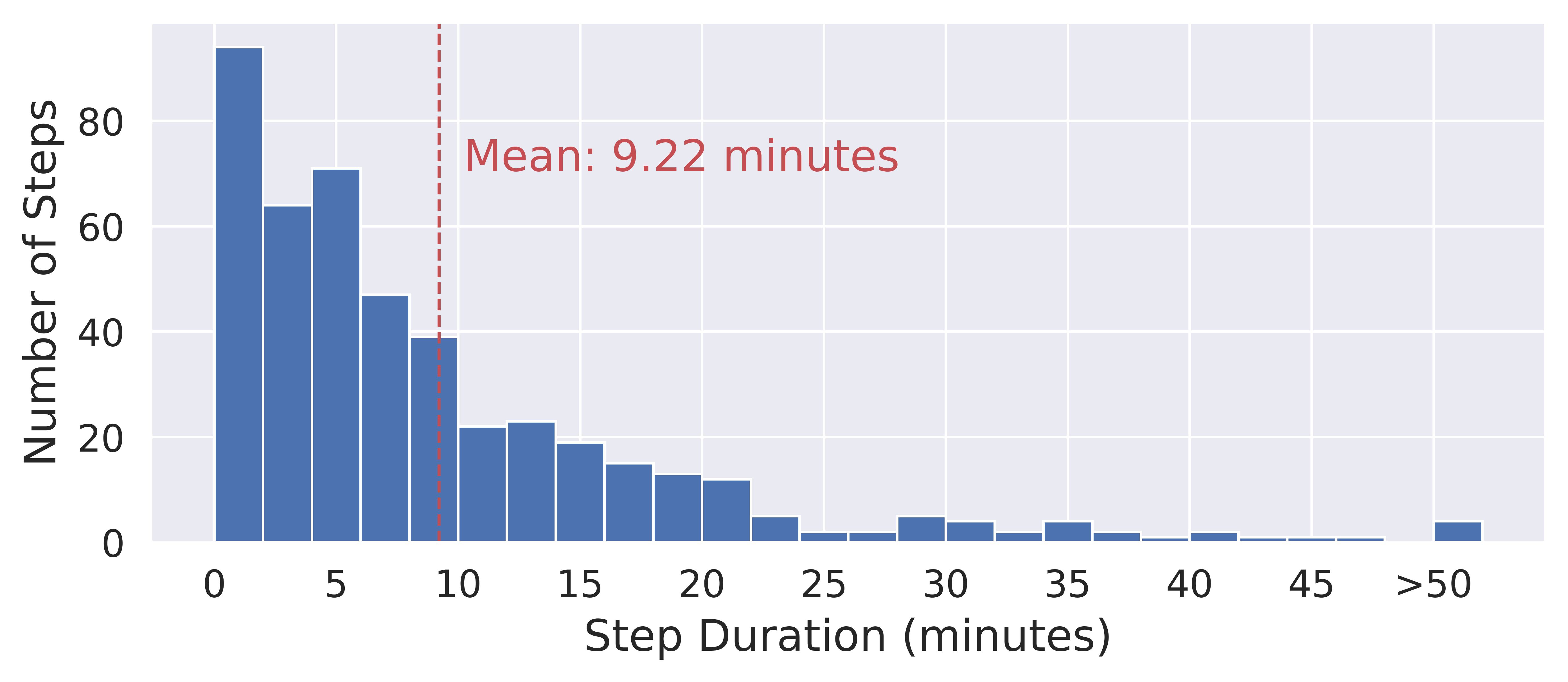}
        }
    }
    \subfloat[Video Clip Number per Step.]{
        \label{fig:dist_step_clip}
        \scalebox{0.33}{
        \includegraphics[width=0.94\linewidth]{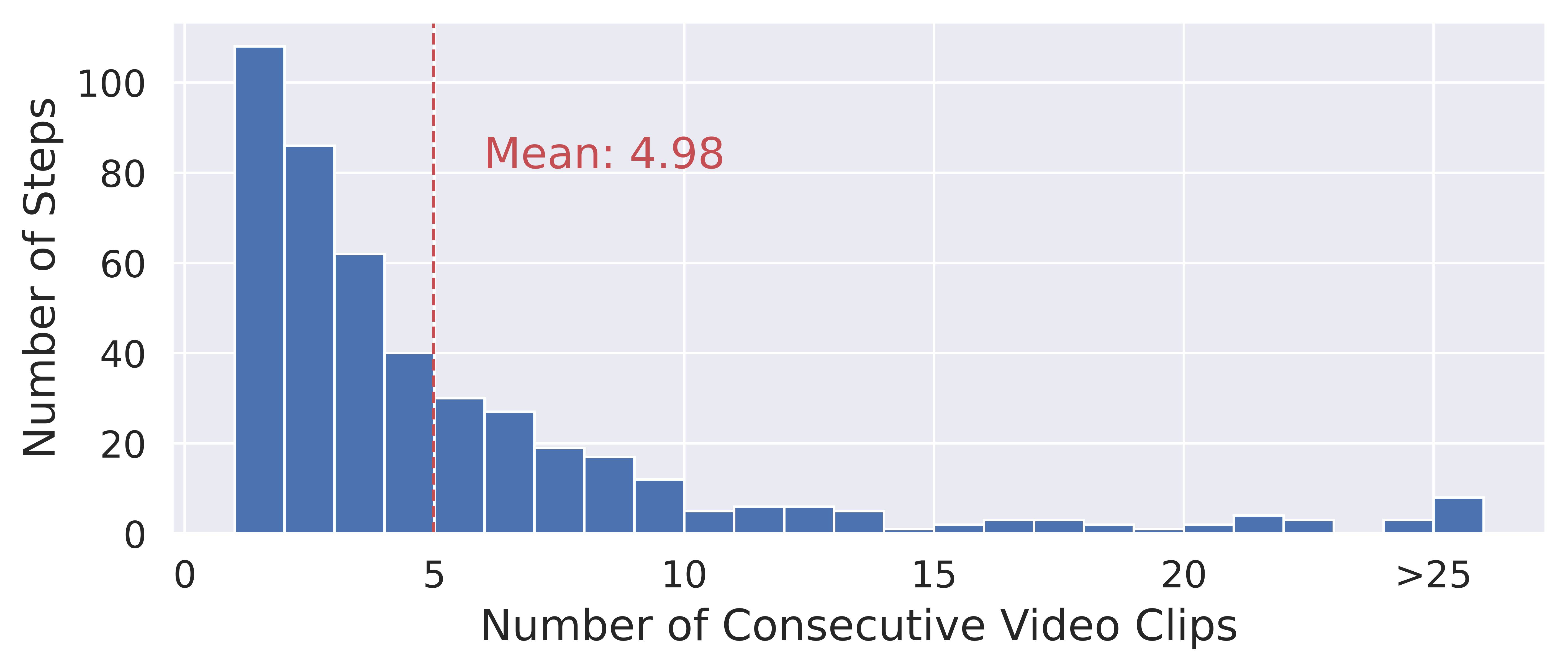}
        }
    }
    \vspace{-7pt}
        \caption{Distributions of (a) Merged video clip duration, (b) Total step duration, and (c) Video clip number per step.
        }
        \vspace{10pt}
        \label{fig:data_statistics}
    \end{figure*}

\begin{figure}[t]
    \centering
    \includegraphics[width=\linewidth]{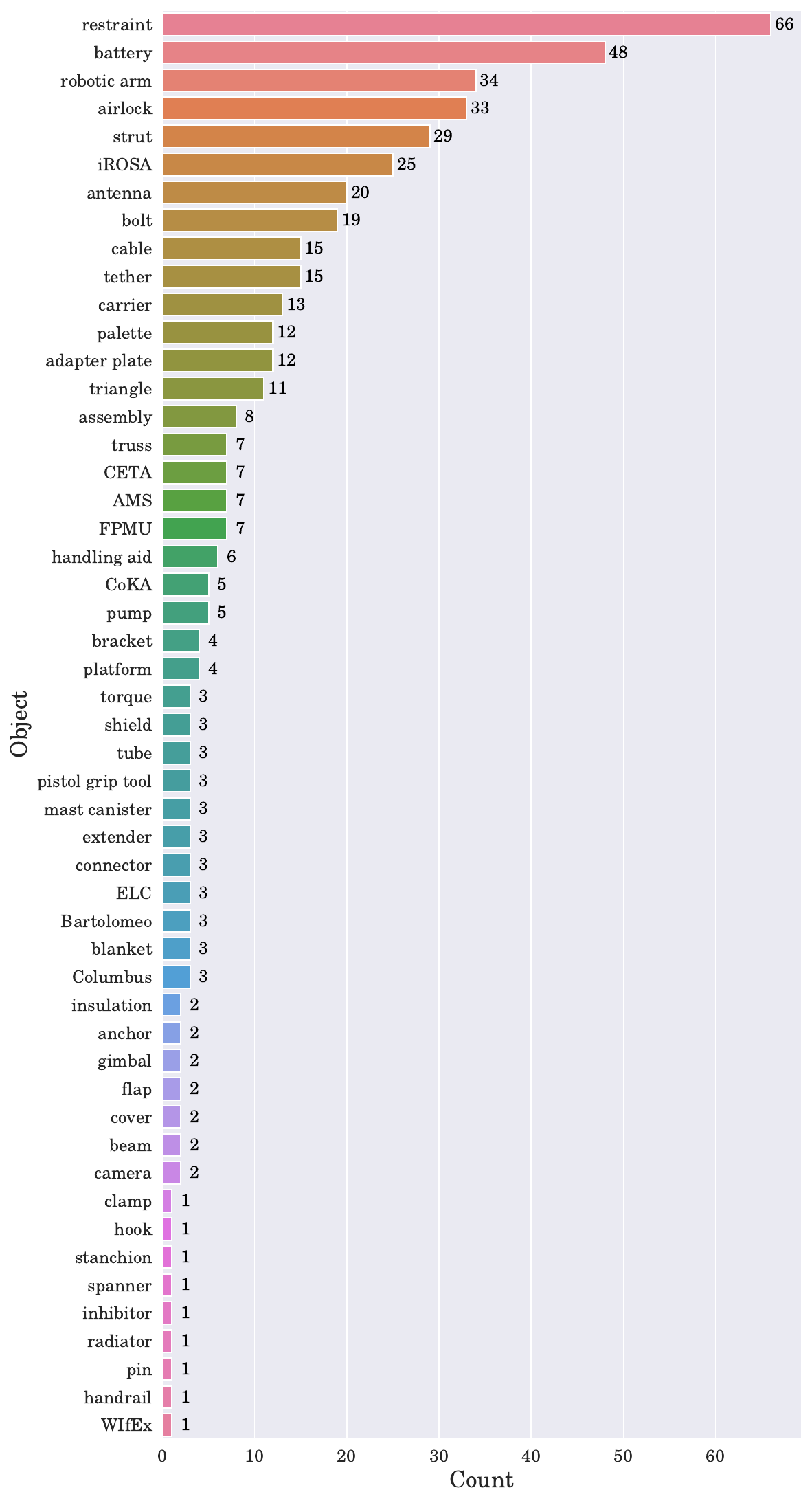}
    \caption{Distribution of objects. It counts how many steps contains each object in their captions.}
    \vspace{10pt}
    \label{fig:object_distribution}
\end{figure}

\begin{figure}[t]
    \centering
    \includegraphics[width=\linewidth]{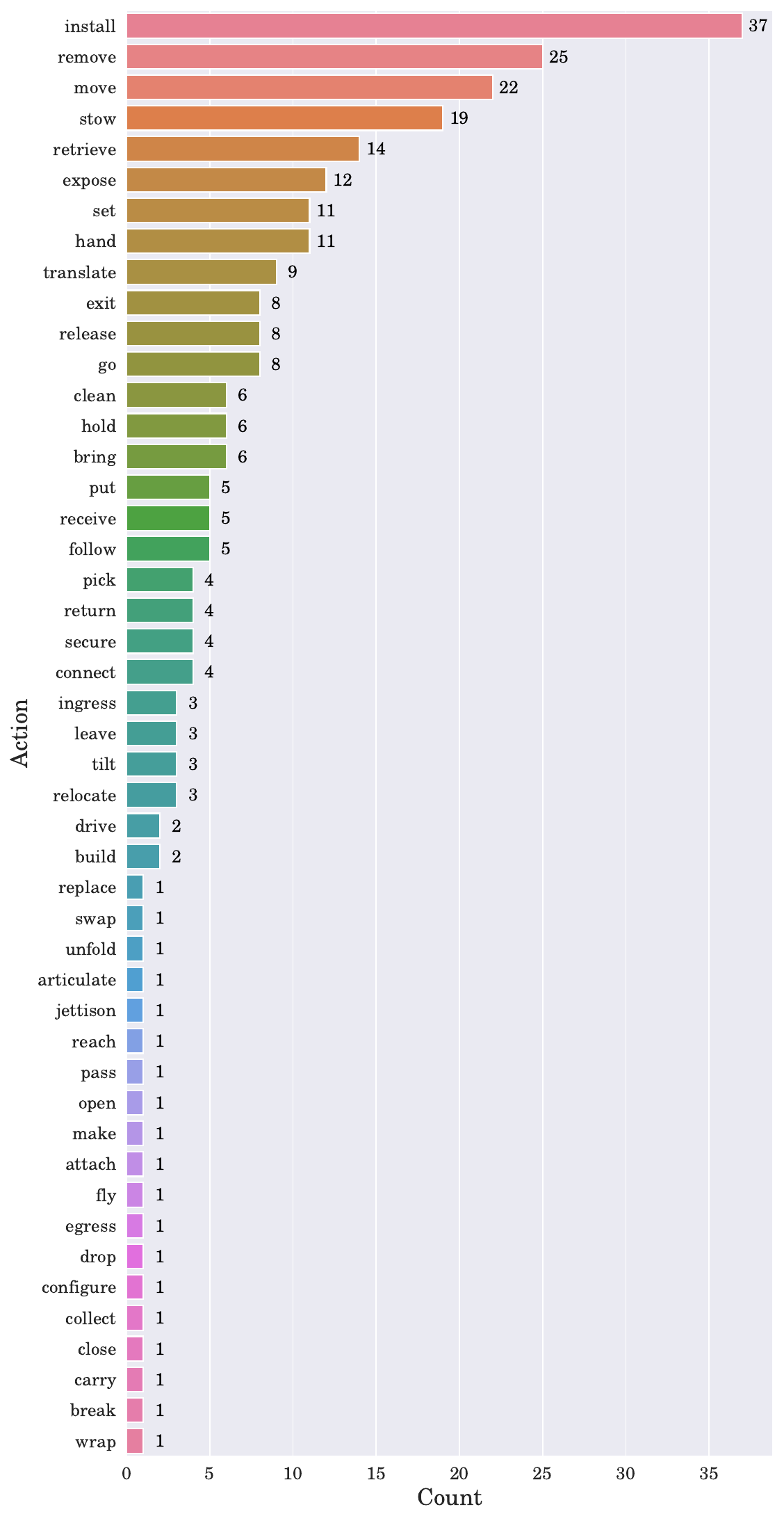}
    \caption{Distribution of actions. It counts how many steps contains each action in their captions.}
    \label{fig:action_distribution}
\end{figure}

\subsection{Statistics of Annotations}
\label{app:statistics}

After merging adjacent clips with the same labels, we obtain in total 3,753 clips with annotated spacewalk steps.
We show several distributions about the clips in \Cref{fig:data_statistics}, including the durations of the video clips, total durations of the steps, and numbers of clips per step.
On average, each merged clip has a length of 92 seconds, each step spans 9 minutes, and each step is composed of 5 clips.

While each spacewalk video has a list of 25 steps on average, a small portion of the steps do not necessarily occur following their order in the list.
After manual examination, we find that around $84\%$ of the adjacent steps are logically non-interchangable in order. For example, ``installing the battery'' must happen after ``taking the battery to the worksite''. The remaining interchangable steps are mostly due to the parallel work of the astronauts. For example, if ``EV 2 retrieves foot restraint'' happens while ``EV 1 goes to the carrier'', there is no guarantee about which step will occur first.

To analyze the diversity of the dataset, we filter the nouns and verbs from the step captions to obtain a list of objects and actions.
\Cref{fig:object_distribution} and \Cref{fig:action_distribution} illustrate the occurrence of each object and action in the step captions. In total, we observe 51 objects and 47 actions across the dataset.
For the 455 steps, which is the sum of the number of steps in each video, we emphasize that no steps are exactly the same because they have their own animation videos and narrations as contexts.
But we manually cluster semantically similar steps and observe 167 distinct groups of steps.

\subsection{Question Answering Task}
\label{sec:question_type}
    \subsubsection{Overview}
\begin{figure}[t]
    \centering
    \includegraphics[width=0.8\linewidth]{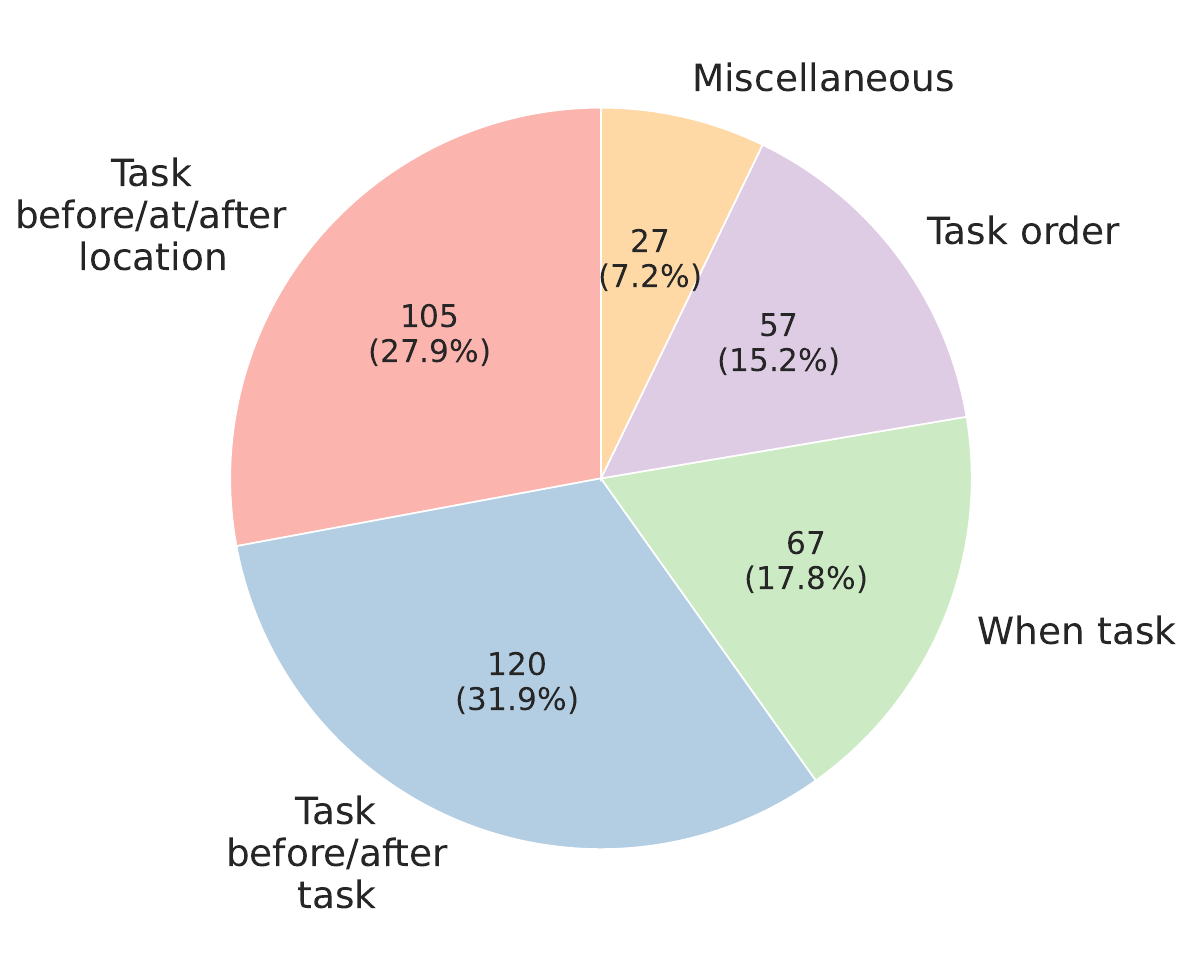}
    \caption{Distribution of the question types of the question answering task.}
    \label{fig:question_types}
\end{figure}

    As described in \Cref{sec:task_definition}, our question answering task includes 376 questions.
    349 of them are generated by templates using our step annotations, while the other 27 miscellaneous questions are manually created.
    Although the manually created questions are fewer, they are more video-specific than those generated by templates.
    For the template-generated questions, we further categorize them into four types: task before/at/after location, task before/after task, when task, and task order.
    Examples of the different question types are listed in \Cref{tab:question_types}.
    The distribution of the question types can be found in \Cref{fig:question_types}.
    In the following, we describe the construction of each type of question in detail.

\begin{table*}[ht]
\centering
\scalebox{0.85}{
\begin{tabular}{c|l}
\toprule
Question Type & \multicolumn{1}{c}{Example} \\
\midrule
 & Q: Which of the following tasks happens before the astronaut arrives at the external pallet?\\
& (A) Chris sets up tools and prepares worksite. \\
Task before/at/after location & (B) Bob moves to IEA. \\
& (C) Chris \& Bob retrieve battery from slot 1. \\
& (D) Chris sets up tether. \\
\midrule
& Q: Which of the following tasks happens after EV1 move back inboard? \\
& (A) EV1 retreive portable foot restraint with extension.\\
Task before/after task & (B) EV1 \& EV2 move to P6. \\
& (C) EV1 \& EV2 install respective bags on worksites.\\
& (D) EV2 move to P1 and install anchor hooks for safety tether.\\
\midrule
& Q: In which part of the video does the task that EV1 \& EV2 install mid strut happen?\\
& (A) The first third of the video. \\
When task& (B) The middle third of the video. \\
& (C) The task does not happen in the video. \\
& (D) The last third of the video.\\
\midrule
& Q: In which order do the tasks happen in the video?\\
& (A) (1) Robotic arm takes Luca to aft side. (2) Drew move to ELC 2. (3) Robotic arm takes Luca to AMS. \\
Task order & (B) (1) Drew move to ELC 2. (2) Robotic arm takes Luca to aft side. (3) Robotic arm takes Luca to AMS. \\
& (C) (1) Robotic arm takes Luca to AMS. (2) Drew move to ELC 2. (3) Robotic arm takes Luca to aft side. \\
& (D) (1) Drew move to ELC 2. (2) Robotic arm takes Luca to AMS. (3) Robotic arm takes Luca to aft side. \\
\midrule
& Q: What type of equipment does the astronaut retrieve first, and how is it utilized during the mission? \\
& (A) Bags containing structure to assemble modification kit.\\
Miscellaneous & (B) Articulating portable foot restraint to later attach to the Canada arm.\\
& (C) Pump to be installed on the Alpha Magnetic Spectrometer.\\
& (D) Bags containing structure to support new solar arrays.\\
\bottomrule
\end{tabular}
}
\vspace{-0.05in}
\caption{Question types and examples in the question answering task.}
\label{tab:question_types}
\end{table*}

    \subsubsection{Template Generated Questions}

    For each of the four types of question templates, we first sample some one-hour-long video segments as follows: for all 3,753 clips, we take the start time of each clip as the beginning and the time one hour later as the end to form a long video segment. We then randomly sample $M$ segments from these. To balance the number of the sub-type questions, we set $M=300$ for task before/at/after location, $M=200$ for task before/after task, $M=100$ for when task, and $M=100$ for task order. We keep $M$ at a moderate level in order to avoid generating repeated questions as much as possible.

    For each sampled long video segment, we follow the descriptions below to construct a question.
    Note that some segments are invalid to construct certain types of questions. For example, videos in which all events occur at the same location cannot be used to generate task before/at/after location questions.
    Therefore, the final size of the generated questions (\Cref{fig:question_types}) is smaller than the $M$ segments we sampled.
    
    \noindent\textbf{Task before/at/after location.}
    The template of this question type is ``Which of the following tasks happens \{before the astronaut arrives at\}/\{while the astronaut is at\}/\{after the astronaut leaves from\} \{location\}?''
    We first manually annotate the location where each step is performed.
    After that, for each sampled one-hour video segment, we uniformly sample one relationship from before/at/after and find all step-location pairs that satisfy this relationship.
    Specifically, a step is considered before/after a location only when all sub-segments related to the step happen before/after all the sub-segments related to the location.
    From the set of valid pairs, we sample one to fill in the template.
    For the negative choices, we preferentially select the other steps from this one-hour video segment that are invalid answers to the generated question.
    If there are less than three of these negative choices, we randomly sample steps within the same spacewalk mission but outside the video segment.

    \noindent\textbf{Task before/after task.}
    The template of this question type is ``Which of the following tasks happens \{before\}/\{after\} \{step\}?''
    Similar to ``task before/at/after location'', we uniformly sample one relationship from before/after and then find all step1-step2 pairs that satisfy this relationship.
    Step 1 is considered before/after step 2 only when all sub-segments of step 1 happen before/after all the sub-segments of step 2.
    From the set of valid pairs, we sample one to fill in the template.
    For the negative choices, we preferentially select the other steps from this one-hour video segment that are invalid answers to the generated question.
    If there are less than three of these negative choices, we randomly sample steps within the same spacewalk mission but outside the video segment.

    \noindent\textbf{When task.}
    The template of this question type is ``In which part of the video does \{step\} happen?''
    The answer can be the first third, the middle third, the last third, or the task does not happen.
    For a given sampled one-hour video segment, we filter the steps whose related sub-segments all fall within the same third of the video.
    We then uniformly sample one of them to construct the question.
    We did not construct questions whose answer is ``the task does not happen''.

    \noindent\textbf{Task order.}
    The question of this type is ``In which order do the tasks happen in the video?''
    The four choices are permutations of three given steps.
    Here, the order of the two steps refers to the order of their first occurrences.
    If the video contains step 1, step 2, and then step 1 again, we consider step 1 to occur before step 2.
    To construct the question, we uniformly sample three steps from the given one-hour video.
    Their ground truth order serves as the correct answer while three permutations of them are used as negative choices.
    
    \subsubsection{Manually Annotated (Miscellaneous) Questions}
    To manually annotate questions for spacewalk videos, we first determine a list of questions for each spacewalk mission -- ``What type of equipment does the astronaut retrieve first, and how is it utilized during the mission?'', ``What is the goal of this mission?'', and ``What did EV1 do while EV2 is doing \{step\}?''.
    For the first two types of questions, we watch the early part of the spacewalk video to determine the answer for each video.
    Negative choices are randomly selected from the correct answers of other spacewalks with similar tasks.
    For the third type of questions, we analyzed the step annotations to find video segments where the annotation order resembles step 1, step 2, step 1.
    Each of these was manually analyzed to determine if EV1 and EV2 were working on separate tasks at the same time.
    If so, the question was formed and the correct answer annotated.
    Negative choices are randomly selected from steps of the same spacewalk and spacewalks with similar tasks.

\subsection{Temporal Certificate}
\label{app:tempcert}
\Cref{tab:tempcert} provides the numerical values we use to plot the temporal certificate figure in \Cref{fig:tempcert}.
It is extended from \cite{EgoSchema} to include Spacewalk-18.

\begin{table*}[t]
\centering
\begin{tabular}{c|c|c|c}
\toprule
Dataset & Temporal Certificate & Average Clip Length & Test Duration \\
\midrule
Kinetics & 1.931 & 10 & 240000 \\
AVA & 0.25 & 1 & 117900 \\
HVU Concept & 0.77 & 10 & 650000 \\
HVU Action & 1.65375 & 10 & 650000 \\
UCF 101 & 1.81 & 6.66 & 9457 \\
Something Something & 1.28 & 3 & 81471 \\
LVU Relationship & 17.12 & 210 & 5364 \\
NextQA & 2.7 & 44 & 47872 \\
EgoSchema & 100 & 180 & 90000 \\
Youtube8M Segment & 0.1 & 5 & 2520000000 \\
MSRVTT & 0.7 & 13 & 38870 \\
IVQA & 0.3 & 18 & 36000 \\
AGQA & 3.7 & 30 & 57600 \\
How2QA & 1.5 & 18 & 16263 \\
ActivityNetQA & 2.4 & 124.58 & 144000 \\
\textbf{Spacewalk-18} & 140 & 89 & 119664 \\
\bottomrule
\end{tabular}
\vspace{-0.05in}
\caption{The numerical values of the temporal certificates. We extend this plot from \cite{EgoSchema} to include Spacewalk-18. The units for all three columns are seconds.  ``Test Duration'' is the total duration of each test set.
}
\label{tab:tempcert}
\end{table*}

\section{Evaluation Metrics of Step Recognition}
\label{app:metrics}
        We use accuracy, mAP, and IoU to evaluate a model's performance on step recognition.
        To eliminate the impact of uneven numbers of task steps in each spacewalk video, we first compute the following metrics per video and then average them across videos.

        \noindent\textbf{Accuracy.} It is the percentage of timestamps whose corresponding steps are correctly recognized.

        \noindent\textbf{Mean Average Precision (mAP).} For methods that can give a confidence score for each task step, we employ mAP as an evaluation metric. We first compute the average precision of each task step except “Irrelevant” and then take the mean.

        \noindent\textbf{Intersection over Union (IoU).} After the models predict the step label for each sampled timestamp, we merge adjacent timestamps with the same predictions into consecutive temporal intervals. They form a segmentation of each step across the entire spacewalk video. For each step, we calculate the IoU with the ground truth segmentation of the step. Finally, we take the average over all the steps as the IoU measurement.

\section{Implementation Details}
\label{app:implementation}

\subsection{Evaluated Models}
\label{app:evaluated_models}

    In the following, we briefly introduce the evaluated models, selected checkpoints, and their frame sampling strategies.
    We by default use the recommended number of sampled frames for each model, rather than unifying them across models.

    \noindent\textbf{EgoVLP} \citep{kevin2022egovlp} is an egocentric video-language model trained on EgoClip dataset. In our step recognition experiments, we sample $4$ frames from a video for feature extraction.

    \noindent\textbf{VideoCLIP} \citep{xu-etal-2021-videoclip} is a video-language model trained with video-text contrastive learning. It uses a Transformer to integrate S3D \cite{Xie_2018_ECCV} video features and align it with text feature. $150$ frames are sampled from a video to extract feature in the step recognition experiments.

    \noindent\textbf{InternVideo} \citep{wang2022internvideo} is a video-language model pre-trained on a large corpus of video-language datasets including HowTo100M \citep{miech2019howto100m} and WebVid10M \citep{Bain21}. 
    It achieves state-of-the-art performances across 39 video datasets from extensive tasks.
    In our step recognition experiments, we use the checkpoint further fine-tuned for video-text retrieval on MSRVTT \citep{xu2016msr-vtt}. $12$ frames are sampled from each video clip in the experiments.

    \noindent\textbf{InternVideo2} \citep{wang2024internvideo2} unifies masked video modeling, crossmodal contrastive learning, and next token prediction to develop a family of video foundation models.
    While their entire training process develops a VLLM, the training stage 2 learns video, audio, and text encoders that share an aligned feature space.
    In our experiments, we evaluated the stage 2 checkpoint \texttt{InternVideo2\_s2-1B} and sample 4 frames for each video clip.

    \noindent\textbf{LLaVA-Next-Video} \citep{llavanextvideo} is a video large language model trained on a large video/image-language corpus. We use the checkpoint \texttt{LLaVA-NeXT-Video-34B-DPO} and uniformly sample 32 frames from each video as visual inputs on both tasks.

    \noindent\textbf{VideoLLaMA2}~\citep{videollama2} is a multimodal large language model capable of understanding video, audio, and language. It employs Spatial-Temporal Convolution connector to capture the intricate spatial and temporal dynamics in the video input. We use the \texttt{VideoLLaMA2-7B} checkpoint and sample 8 frames from each video on the step recognition task.
    We use the \texttt{VideoLLaMA2-7B-16F} checkpoint and sample 16 frames on the question answering task.

    \noindent\textbf{LongVU}~\citep{shen2024longvu} proposes a spatiotemporal adaptive compression mechanism to reduce the number of video tokens for long-form video understanding. We use the \texttt{LongVU\_Qwen2\_7B} checkpoint. This model samples video frames at 1 FPS regardless of the video’s duration.
    \noindent\textbf{Qwen2.5VL}~\citep{bai2025qwen25vltechnicalreport} employs dynamic resolution processing and absolute time encoding to deal with images and videos with various resolutions and durations.
    We use the checkpoint \texttt{Qwen2.5-VL-7B-Instruct}.
    This model by default samples video frames at 2 FPS.
    However, it has a maximum limit of 768 frames.
    Consequently, the number of sampled frames is 768 on our question answering task.
    \noindent\textbf{InternVL3}~\citep{zhu2025internvl3exploringadvancedtraining} develops a VLLM through native multimodal pre-training, which consolidates language pre-training and multi-modal alignment training into a single pre-training stage.
    We evaluate the checkpoint \texttt{InternVL3-8B} and sample 32 video frames on the step recognition task and 256 frames on the question answering task.

    \noindent\textbf{GPT-4o} and \textbf{GPT-5} are a proprietary multimodal large language model API. As they receive multiple images as input but not a video file, we uniformly sample a few frames from a video and feed them to the API following the temporal order. The used checkpoint is \texttt{gpt-4o-2024-05-13} for GPT-4o and \texttt{gpt-5-2025-08-07} for GPT-5. The number of sampled frames 8 on the step recognition task and 32 on the question answering task.

    \noindent\textbf{Caption-augmented LLM}. Following \cite{zhang2023simple}, on both of our two tasks, we use \texttt{LLaVA-1.5-13B}~\citep{liu2023improvedllava} to caption 60 uniformly sampled video frames with prompt ``Describe the image in 30 words''. The generated captions are delivered into GPT-4o (\texttt{gpt-4o-2024-05-13}) to answer the given question.

    \noindent\textbf{VideoMAE}~\citep{tong2022videomae} is a video encoder trained with masked video modeling. We use the \texttt{videomae-base} checkpoint and sample 16 frames from each video on the step recognition task.
    
\subsection{VLLM Prompts}
\label{app:vllm_prompts}

To evaluate VLLMs on the step recognition task, we format the task into multi-choice video question answering. Given the video content and transcript,we ask the model to choose a step index from a given list. Specifically, the VLLM prompt for the step recognition task is as following:

\mybox{
\textcolor{blue}{<video>}\\
You are given a spacewalk video, where the spacewalk mission can be divided into \textcolor{blue}{<number of steps>} steps. \\
The transcript of the video speech is: \textcolor{blue}{<transcript>}.\\
Please provide a single-number answer (from 0 to \textcolor{blue}{<number of steps>}) to the following multiple-choice question, and your answer must be one of the numbers from 0 to \textcolor{blue}{<number of steps>}. You must not provide any other response or explanation. If you are not sure, answer with the most likely answer.\\
Here is the question: \textcolor{blue}{Which step does the frame in the middle of this video belong to?}\\
Here are the choices:\\
(0) Irrelevant: The mission control center, noisy shots (e.g. blue screen), or tasks not planned for the spacewalk.\\
(1) \textcolor{blue}{<step 1 caption>}: \textcolor{blue}{<step 1 transcript>}.\\
(2) \textcolor{blue}{<step 2 caption>}: \textcolor{blue}{<step 2 transcript>}.\\
... ...\\
(N) \textcolor{blue}{<step N caption>}: \textcolor{blue}{<step N transcript>}.
}

For our task 2, the question answering task, we use the following prompt:
\mybox{
\textcolor{blue}{<video>}\\
Please provide a single-letter answer (from A to D) to the following multiple-choice question, and your answer must be one of the letters from A to D. You must not provide any other response or explanation. If you are not sure, answer with the most likely answer.\\
Here is the question: \textcolor{blue}{<question>}\\
Here are the choices: \\
\textcolor{blue}{(A) <option\_1>}\\
\textcolor{blue}{(B) <option\_2>}\\
\textcolor{blue}{(C) <option\_3>}\\
\textcolor{blue}{(D) <option\_4>}
}

On both tasks, we set the temperature to 0 to disable sampling during answer generation for all VLLMs except GPT-5. The GPT-5 API only supports a temperature of 1.

\subsection{Last-layer Fine-tuning Contrastive VLMs}
\label{app:last_layer_hyper}

When last-layer fine-tuning the contrastive VLMs, we use Adam \citep{Kingma2014} optimizer with a learning rate of $1e-4$ and a batch size of 2048. All models are trained for 20 epochs and the checkpoint with the best validation loss is picked.

\subsection{All-layer Fine-tuning Contrastive VLMs}
\label{app:all_layer_hyper}

We follow the code base of InternVideo \citep{wang2022internvideo} to fine-tune its entire model backbone on our step recognition task. We use Adam \citep{Kingma2014} optimizer with a learning rate of $4e-6$. The learning rate warms up linearly in the first $10\%$ training steps, after which cosine annealing is adapted. Besides, a weight decay of $0.2$ is used.
We fine-tune the model for 1 epoch with a batch size of 16.
We find that training for more than 1 epoch always leads to over-fitting.

\subsection{Long-term Feature Bank}
\label{app:lfb}

To employ Long-term Feature Bank (LFB) \citep{wu2019long} to solve the step recognition task, we first divide the spacewalk recording clip $V_{t,w}$ into $k$-second-long segments and extracts their InternVideo features with a frame sampling rate of 1 FPS, respectively. Since InternVideo is pre-trained with 12 frames per video, we set $k=12$. There are in total $M=60\times w/12$ segments for a $w$-minute-long video and the query timestamp falls in segment $m=M/2$. Denote the feature of the $i$-th segment as $f_i'$. LFB aims to incorporate the history feature $f_H'=f_{1:m-1}'$ and the future feature $f_F'=f_{m+1:M}'$ into the query timestamp feature $f_m'$. Based off \cite{wu2019long}, we design four context incorporation mechanisms -- \textbf{LFB Avg}, \textbf{LFB Cat}, \textbf{LFB NL}, and \textbf{LFB TF}. Each of these methods integrate $f_m'$, $f_H'$, and $f_F'$ into a feature $f$ for the entire video clip $V_{t,w}$. We substitute $f$ for $f_s$ in Equation (1) to train the linear layer $G_\theta(\cdot)$ and the parameters of NL/TF blocks jointly. The pre-trained video-language encoders are always frozen in this process. Unless otherwise specified, the training hyperparameters are the same as those of last-layer fine-tuning in Appendix~\ref{app:last_layer_hyper}.
            
    \noindent\textbf{LFB Avg.} We pool the query, history, and future features together to form a single feature, \emph{i.e.},
            \begin{equation}
                f=\text{Average}(\left[f_H',f_m',f_F'\right])=\text{Average}(f_{1:M}').
            \end{equation}
    
    \noindent\textbf{LFB Cat.} We first pool the history and future features, respectively, and then concatenate them with the query feature, \emph{i.e.},
            \begin{equation}
                f=\left[\text{Average}(f_H'),f_m',\text{Average}(f_F')\right].
            \end{equation}
                
\begin{figure}[t]
    \centering
    \includegraphics[width=0.8\linewidth]{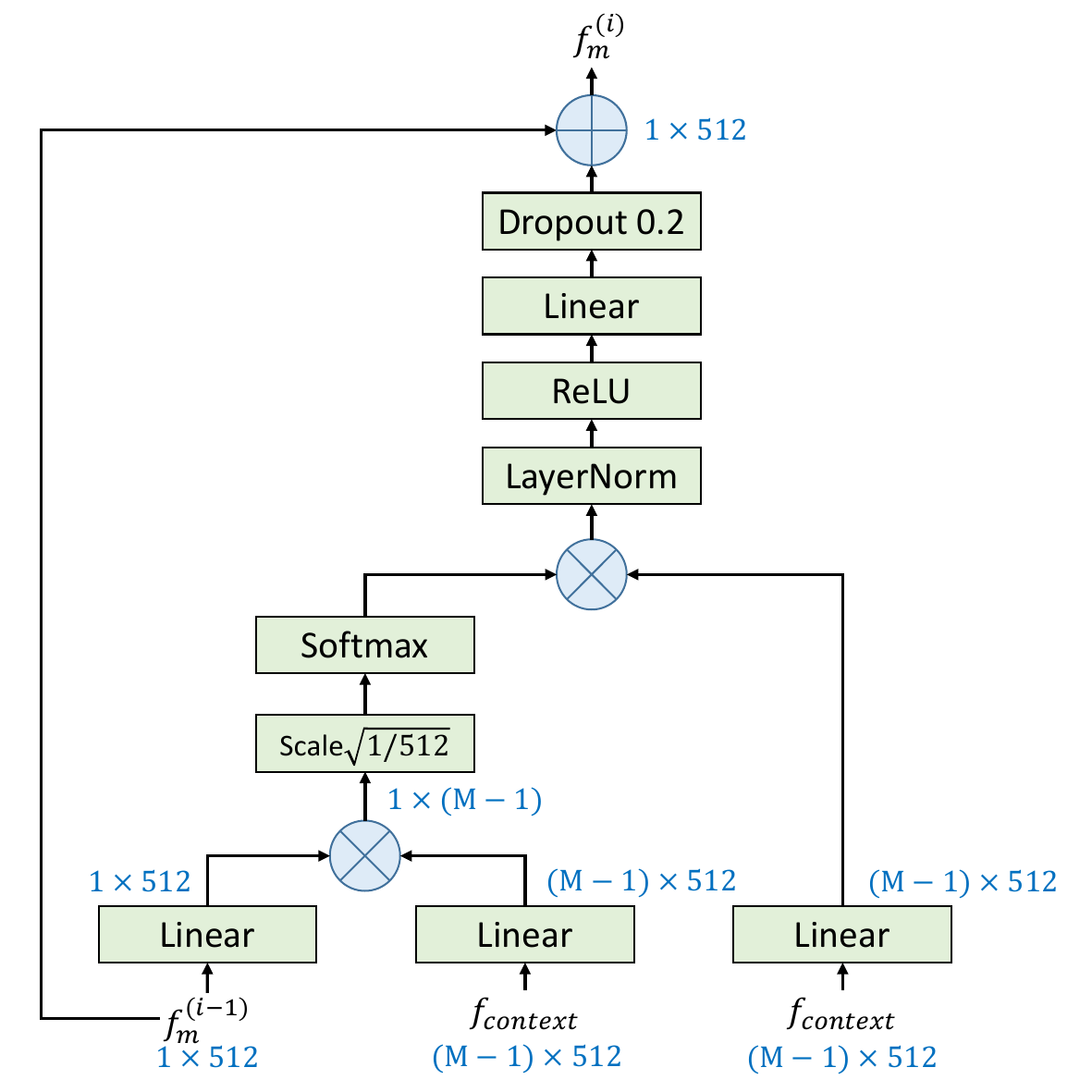}
    \caption{Architecture of a non-local block $\text{Non-local}(f_m^{(i-1)}, f_\text{context})$ borrowed from \cite{wu2019long}. It employs attention mechanism between the query feature $f_m^{(i-1)}$ and context feature $f_\text{context}$, and produces an updated query feature $f_m^{(i)}$. When stacking multiple non-local blocks, the query feature is iteratively updated while the context feature remains unchanged.}
    \label{fig:nlblock}
\end{figure}

    \noindent\textbf{LFB NL.} We stack two non-local blocks \citep{Wang_2018_CVPR} to learn the interaction between the query feature and the context features via attention mechanism. The architecture of a non-local block is shown in \Cref{fig:nlblock}.
    Before feeding the features into non-local blocks, we first add positional encoding to them and use a linear layer to project them into a $512$-dimensional space.
    The output of the non-local blocks is concatenated with the query feature to form the final clip feature. Formally,
            \begin{equation}
                f=\left[f_m',\text{Non-local}\left(L(f_m'+P_m),L(\left[f_H'+P_H,f_F'+P_F\right])\right)\right],
            \end{equation}
        where $L(\cdot)$ is a linear projection and $P_m$, $P_H$, and $P_F$ are positional encoding for the query, history, and future timestamps. We use a learning rate of $1e-3$ and a batch size of $512$ during training.
    
    \noindent\textbf{LFB TF.} We employ a two-layer Transformer encoder upon the concatenation of the history, query, and future features, \emph{i.e.},
            \begin{equation}
                f=\text{Transformer}(\left[f_H',f_m',f_F'\right]+P)=\text{Transformer}(f_{1:M}'+P),
            \end{equation}
        where $P$ is positional encoding. The hidden size is the same as the feature dimensions, which is $2304$. We use a learning rate of $1e-4$ and a batch size of $512$ to train the model.

\section{Additional Experiments}
\label{app:exp}

    \begin{table*}[t]
    \centering
    \scalebox{0.9}{
        \begin{tabular}{c|c|ccccc|c}
            \toprule
            Method & \#Frames & TL & TT & WT & TO & MI & Acc \\
            \midrule
            Random & - & 25.00 & 25.00 & 25.00 & 25.00 & 25.00 & 25.00\\
            LLaVA-Next-Video-34B & 32 & 26.67 & 33.33 & 25.37 & 29.82 & 33.33 & 29.52 \\
            VideoLLaMA2-7B & 16 & 25.71 & 40.83 & 20.90 & 36.84 & 29.63 & 31.65 \\
            InternVL3-8B & 256 & 35.24 & 34.17 & 28.36 & 19.30 & 40.74 & 31.65 \\
            Qwen2.5VL-7B & 768 & 40.00 & 33.33 & 37.31 & 21.05 & 29.63 & 33.78 \\
            LongVU-7B & 1 FPS & 32.38 & 38.33 & 50.75 & 24.56 & 33.33 & 36.44 \\
            Caption-enhanced LLM & 60 & 33.33 & 35.00 & 26.87 & 15.79 & 25.93 & 30.37 \\
            GPT-4o & 32 & 33.33 & 33.33 & 32.84 & 31.58 & 25.93 & 32.45 \\
            GPT-5 & 32 & 39.05 & 50.83 & 53.73 & 47.37 & 37.04 & 46.54 \\
            \bottomrule
        \end{tabular}
    }
    \caption{Accuracy on different question types of the question answering task. TL: task before/at/after location; TT: task before/after task; WT: when task; TO: task order; MI: miscellaneous.}
    \vspace{-5pt}
     \label{tab:qa_acc_per_type}
    \end{table*}

    \begin{figure}[t]
    \centering
    \subfloat[Number of training samples per video]{
        \label{fig:data_size_n_sample}
        \includegraphics[width=0.96\linewidth]{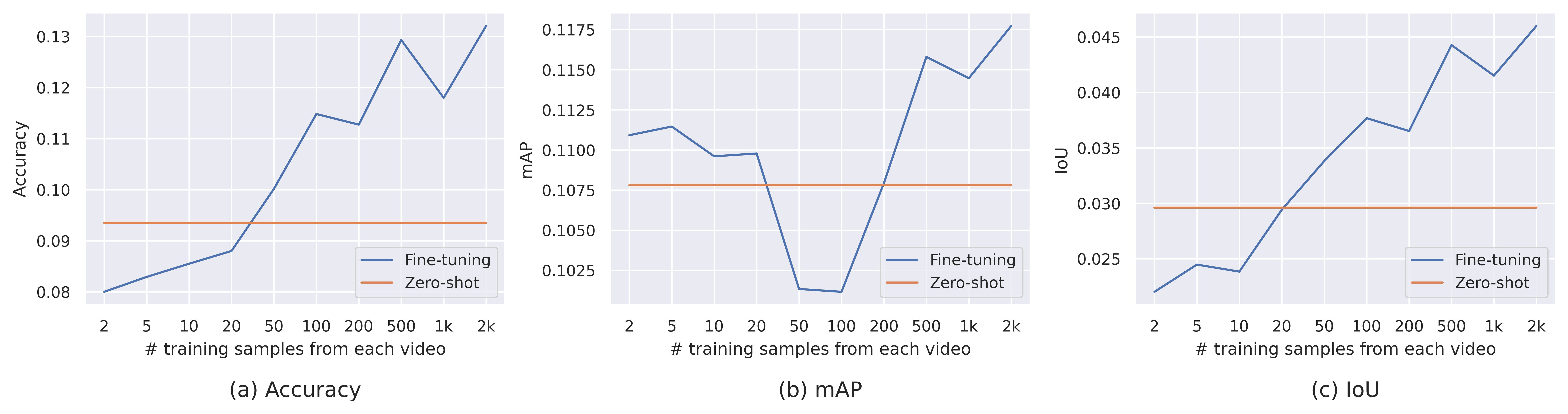}
    }
    
    \subfloat[Number of training videos]{
        \label{fig:data_size_n_video}
        \includegraphics[width=0.96\linewidth]{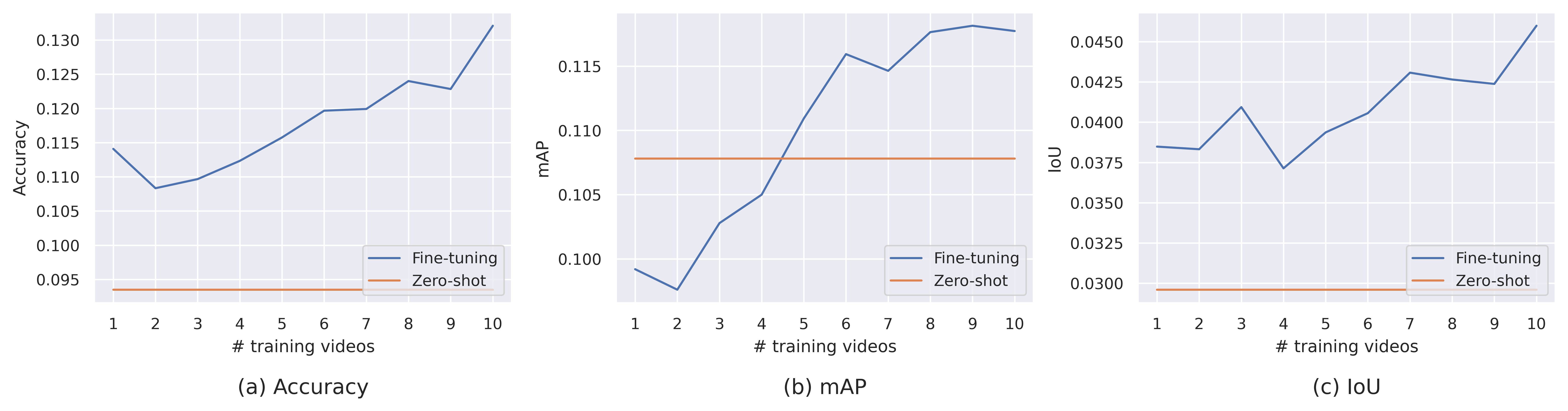}
    }
        \caption{Ablation studies on training data size. (a) Varying number of training samples per video (10 training videos in total). (b) Varying number of training videos (2000 training samples from each video). We full-layer fine-tune InternVideo with a context length of 1 minute. The blue curves are the fine-tuning performances while the orange straight lines are the zero-shot performances on the corresponding metrics.}
    \label{fig:data_size_ablation}
    \end{figure}

\subsection{Accuracy of Each Question Answering Type}

    As discussed in \Cref{sec:question_type}, our question answering task includes five different types of questions.
    In \Cref{tab:qa_acc_per_type}, we decompose the model performance in \Cref{tab:main_results_task1} into the five question types.
    We find that most of models perform the best on the task before/after task (TT) questions, while the miscellaneous (MI) questions are generally difficult to the VLLMs.
    However, given the transcript of the animation video, the oracle model makes great improvement on the miscellaneous questions, further highlighting the incapability of the VLLMs of abstracting the tasks in spacewalk videos.
    We notice that the oracle model performs poorly on the when task (WT) questions.
    This is expected because the time period options of these questions are based on the segmented one-hour video clip, while the oracle information given to the model covers the entire spacewalk mission.

\subsection{Ablation Study on Fine-tuning Data Size}
\label{app:datasize}

        We conduct experiments to explore the impact of data size on all-layer fine-tuning InternVideo on the step recognition task.
        In Figure~\ref{fig:data_size_n_sample}, we ablate the number of samples drawn from each video in the training set. The model accuracy and IoU first rise as the data size grows. However, when more than 500 samples are drawn from each video (5k samples in total), the performance becomes saturated (less than $0.3\%$ gain from 500 samples to 2k samples).
        This indicates that the number of training samples per video is not the bottleneck for our fine-tuning approach.

        In Figure~\ref{fig:data_size_n_video}, we vary the number of training videos, while 2k training samples are drawn from each video. The accuracy and IoU curves show significant upward trends when the videos get more, while mAP is saturated when more than 6 videos are used. This shows that fine-tuning on more spacewalk videos might better adapt the model the novel domain. However, the number of spacewalk videos is naturally limited by the number of real spacewalk missions, which disables large-scale data collection. So we urge the need for domain adaptation method with higher data efficiency.
        
        All these experiments are conducted under a context window length of 1 minute. For different training data sizes, we keep the number of fine-tuning epochs the same. Therefore, the less training data, the smaller number of fine-tuning steps. However, we find that naively enlarging the number of training steps always leads to over-fitting.

    \begin{table}[t]
    \centering
        \scalebox{0.72}{
        \begin{tabular}{c|cccccc}
            \toprule
            Speedup & 1$\times$ & 2$\times$ & 4$\times$ & 8$\times$ & 16$\times$ & 32$\times$ \\
            \midrule
            InternVideo & 10.03 & 10.00 & 9.88 & 9.80 & 9.60 & 9.42 \\
            VideoLLaMA2 & 5.17 & 4.00 & 10.50 & 15.00 & 17.33 & 16.00 \\
            \bottomrule
        \end{tabular}
        }
        \vspace{-8pt}        
        \caption{Step recognition accuracy under different speedups.}
    \label{tab:speedup}
    \end{table}

    \begin{table}[t]
    \centering
        \scalebox{0.8}{
        \begin{tabular}{c|cccc}
            \toprule
            Number of frames & 32 & 64 & 128 & 256 \\
            \midrule
            Accuracy & 29.52 & 29.26 & 30.85 & 31.65 \\
            \bottomrule
        \end{tabular}
        }
        \vspace{-8pt}        
        \caption{Ablation on frame sampling on QA task with InternVL3.}
    \label{tab:frame_sampling_internvl}
    \end{table}

    \begin{table}[t]
    \centering
        \scalebox{0.8}{
        \begin{tabular}{c|cccc}
            \toprule
            Number of frames & 8 & 16 & 32 & 60 \\
            \midrule
            Accuracy & 31.65 & 30.32 & 32.45 & 31.91 \\
            \bottomrule
        \end{tabular}
        }
        \vspace{-8pt}        
        \caption{Ablation on frame sampling on QA task with GPT-4o.}
    \label{tab:frame_sampling_gpt4o}
    \end{table}

    \begin{table*}[t]
    \centering
        \scalebox{0.76}{
        \begin{tabular}{c|ccc|ccc|ccc|ccc}
            \toprule
            \multirow{2}{*}{Method} &\multicolumn{3}{c}{$w$ = 1 min} & \multicolumn{3}{c}{$w$ = 2 min} & \multicolumn{3}{c}{$w$ = 3 min} & \multicolumn{3}{c}{$w$ = 5 min} \\
            \cmidrule(r){2-4}
            \cmidrule(r){5-7}
            \cmidrule(r){8-10}
            \cmidrule(r){11-13}
            & Acc. & mAP & IoU & Acc. & mAP & IoU & Acc. & mAP & IoU & Acc. & mAP & IoU \\
            \midrule
            EgoVLP & 6.34 & 8.92 & 2.17 & 7.73 & 9.83 & 2.59 & 9.68 & 10.66 & 3.03 & 10.21 & 10.86 & 3.18 \\
            VideoCLIP & 8.40 & 9.80 & 3.21 & 9.01 & 10.40 & 3.28 & 9.98 & 11.14 & 3.71 & 8.87 & 10.39 & 3.46 \\
            InternVideo & 10.12 & \textbf{12.17} & \textbf{4.02} & 11.22 & 12.30 & 4.22 & 11.13 & \textbf{12.68} & 4.04 & 10.08 & \textbf{12.53} & 4.04 \\
            LLaVA-Next-Video & \textbf{10.35} & - & 3.77 & 12.03 & - & \textbf{4.53} & 13.71 & - & 5.07 & 13.82 & - & 4.92 \\
            VideoLLaMA2 & 9.34 & - & 2.88 & \textbf{12.43} & - & 4.32 & \textbf{14.37} & - & \textbf{5.45} & \textbf{17.32} & - & \textbf{6.28} \\
            \bottomrule
          \end{tabular}
        }
        \scalebox{0.76}{
        \begin{tabular}{c|ccc|ccc|ccc}
            \toprule
            \multirow{2}{*}{Method} &\multicolumn{3}{c}{$w$ = 10 min} & \multicolumn{3}{c}{$w$ = 15 min} & \multicolumn{3}{c}{$w$ = 20 min} \\
            \cmidrule(r){2-4}
            \cmidrule(r){5-7}
            \cmidrule(r){8-10}
            & Acc. & mAP & IoU & Acc. & mAP & IoU & Acc. & mAP & IoU \\
            \midrule
            EgoVLP & 8.63 & 10.13 & 2.65 & 7.42 & 9.37 & 2.12 & 8.33 & 9.61 & 2.48 \\
            VideoCLIP & 8.08 & 10.44 & 3.36 & 8.47 & 9.79 & \textbf{3.79} & 7.23 & 8.82 & 3.20 \\
            InternVideo & 10.26 & \textbf{12.36} & 4.18 & 9.66 & \textbf{11.28} & 3.54 & 9.94 & \textbf{10.53} & 3.87 \\
            LLaVA-Next-Video & 11.85 & - & 4.58 & 11.80 & - & - & 12.05 & - & 4.00 \\
            VideoLLaMA2 & \textbf{20.98} & - & \textbf{8.03} & \textbf{19.72} & - & - & \textbf{17.49} & - & \textbf{6.69} \\
            \bottomrule
          \end{tabular}
        }
    \caption{Last-layer fine-tuning performances of contrastive VLMs and zero-shot performances of VLLMs on the step recognition task under varying context window lengths. They are used to plot Figure~\ref{fig:context_length}.}
    \label{tab:context_ft_steprecog}
    \end{table*}
    
    \begin{table*}[t]
    \centering
        \scalebox{0.76}{
        \begin{tabular}{c|ccc|ccc|ccc|ccc}
            \toprule
            \multirow{2}{*}{Method} &\multicolumn{3}{c}{$w$ = 1 min} & \multicolumn{3}{c}{$w$ = 2 min} & \multicolumn{3}{c}{$w$ = 3 min} & \multicolumn{3}{c}{$w$ = 5 min} \\
            \cmidrule(r){2-4}
            \cmidrule(r){5-7}
            \cmidrule(r){8-10}
            \cmidrule(r){11-13}
            & Acc. & mAP & IoU & Acc. & mAP & IoU & Acc. & mAP & IoU & Acc. & mAP & IoU \\
            \midrule
            Sparse Frame Sampling & 10.12 & 12.17 & 4.02 & 11.22 & 12.30 & 4.22 & 11.13 & 12.68 & 4.04 & 10.08 & 12.53 & 4.04 \\
            Dense Frame Sampling & 10.28 & 12.54 & 4.20 & 11.98 & 12.98 & 4.60 & 11.75 & 13.51 & 4.13 & 11.14 & 13.19 & 4.43 \\
            LFB Avg & 11.47 & \textbf{13.92} & 4.39 & 11.98 & \textbf{15.15} & 4.49 & 11.71 & \textbf{16.14} & 4.29 & 12.37 & 17.25 & 3.73 \\
            LFB Cat & \textbf{12.22} & 13.43 & \textbf{4.49} & \textbf{12.76} & 14.25 & \textbf{4.82} & \textbf{12.98} & 14.83 & \textbf{4.95} & \textbf{13.12} & 15.46 & \textbf{4.76} \\
            LFB NL & 9.31 & 13.23 & 3.61 & 10.22 & 13.96 & 3.48 & 9.50 & 14.98 & 3.26 & 10.70 & \textbf{17.28} & 3.01 \\
            LFB TF & 11.31 & 13.08 & 3.77 & 11.67 & 14.83 & 3.71 & 12.45 & 15.68 & 4.11 & 11.38 & 16.10 & 3.04 \\
            \bottomrule
          \end{tabular}
        }
        \scalebox{0.76}{
        \begin{tabular}{c|ccc|ccc|ccc}
            \toprule
            \multirow{2}{*}{Method} &\multicolumn{3}{c}{$w$ = 10 min} & \multicolumn{3}{c}{$w$ = 15 min} & \multicolumn{3}{c}{$w$ = 20 min} \\
            \cmidrule(r){2-4}
            \cmidrule(r){5-7}
            \cmidrule(r){8-10}
            & Acc. & mAP & IoU & Acc. & mAP & IoU & Acc. & mAP & IoU \\
            \midrule
            Sparse Frame Sampling & 10.26 & 12.36 & 4.18 & 9.66 & 11.28 & 3.54 & 9.94 & 10.53 & 3.87 \\
            Dense Frame Sampling & 10.72 & 13.23 & 4.20 & - & - & - & - & - & - \\
            LFB Avg & 10.86 & \textbf{18.11} & 3.35 & 11.39 & \textbf{18.65} & 3.03 & 10.77 & 18.29 & 2.68 \\
            LFB Cat & \textbf{13.20} & 16.69 & 4.89 & 11.98 & 16.43 & \textbf{4.52} & \textbf{12.78} & 16.41 & \textbf{4.84} \\
            LFB NL & 12.78 & 16.81 & 4.20 & 10.57 & 17.55 & 3.64 & 10.08 & 17.41 & 3.39 \\
            LFB TF & 13.18 & 16.96 & \textbf{5.00} & \textbf{12.98} & 17.06 & 4.42 & 11.17 & \textbf{18.79} & 3.35 \\
            \bottomrule
          \end{tabular}
        }
    \caption{Performances of different temporal context incorporation methods on the step recognition task. These methods are built upon frozen InternVideo. They are used to plot Figure~\ref{fig:context_length_LFB}.}
    \label{tab:context_lfb}
    \vspace{-10pt}
    \end{table*}
    
\subsection{Video Speedup}

    Astronauts on spacewalks sometimes have slower motion than earthly activities, which may cause temporal out-of-domain issues to video-language models.
    To investigate this factor, we speed up the videos at different rates and evaluate InternVideo and VideoLLaMA2 at 1 FPS with 5-minute contexts.
    In Table~\ref{tab:speedup}, the original speed suits InternVideo better, while speedup boosts VideoLLaMA2.

\subsection{Ablation on Frame Sampling on QA Task}
\label{app:qa_frame_samping}
Videos in our question answering task are hour-long.
The number of sampled frames might be essential to represent the video content.
Therefore, although each model has a suggested number of frames per query, we may increase the number of frames to achieve better performance on long-form videos.

To investigate if this is true, we conduct an ablation study with InternVL3 and GPT4o.
The results are in Table~\ref{tab:frame_sampling_internvl} and Table~\ref{tab:frame_sampling_gpt4o}.
For InternVL3, increasing the number of frames from 32 to 256 improves the accuracy from 29.52\% to 31.65\%, indicating the effectiveness of dense frame sampling.
While InternVL3 with 256 frames already saturates the GPU memory of a single RTX A6000, we did not conduct experiments with more frames.
On the contrary, increasing the number of sampled frames did not improve GPT-4o’s performance.
This shows that the effect of dense sampling is model-dependent and cannot generally boost performance.

\subsection{Temporal Context Incorporation}
\label{app:context_incorp}

    In Section~\ref{sec:context}, we test models under varying context window lengths to investigate their capability to incorporate temporal context. The numerical results used in Figure~\ref{fig:context_length} are listed in Table~\ref{tab:context_ft_steprecog}. Those used in Figure~\ref{fig:context_length_LFB} are in Table~\ref{tab:context_lfb}.

\subsection{Qualitative Results}
\label{app:qual}

In this section, we demonstrate qualitative results on the step recognition task and the question answering task.

\begin{figure*}
    \centering
    \vspace{50pt}
    \includegraphics[width=0.94\linewidth]{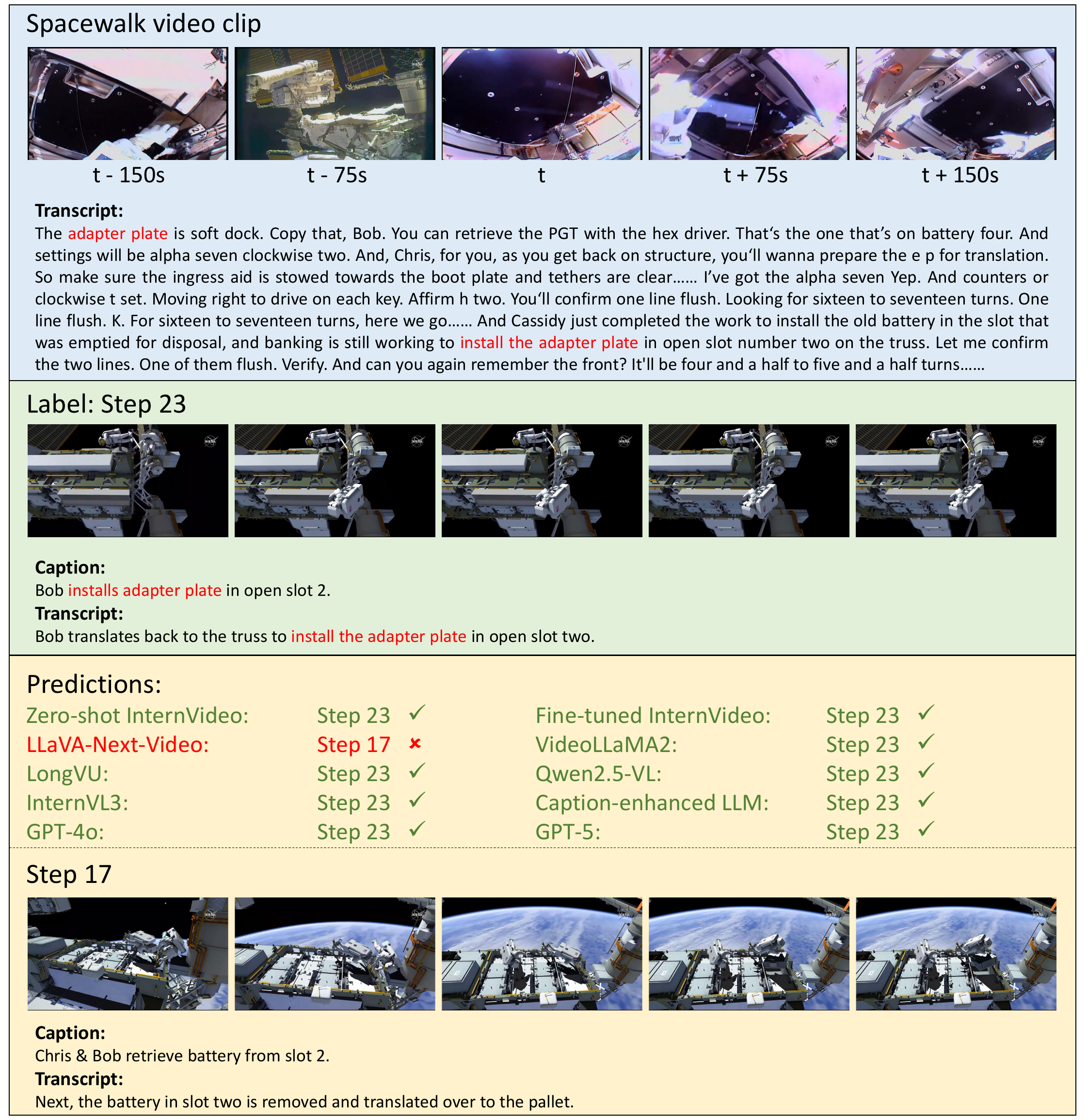}
    \caption{In this example of step recognition, the spacewalk video transcript mentions that the astronaut is installing an adapter plate, and the video also shows them tightening the screws on the adapter plate. All models except LLaVA-Next-Video match this video with the step of installing adapter plate.}
    \label{fig:example_task1_1}
    \vspace{50pt}
\end{figure*}

\begin{figure*}
    \centering
    \vspace{-5pt}
    \includegraphics[width=0.94\linewidth]{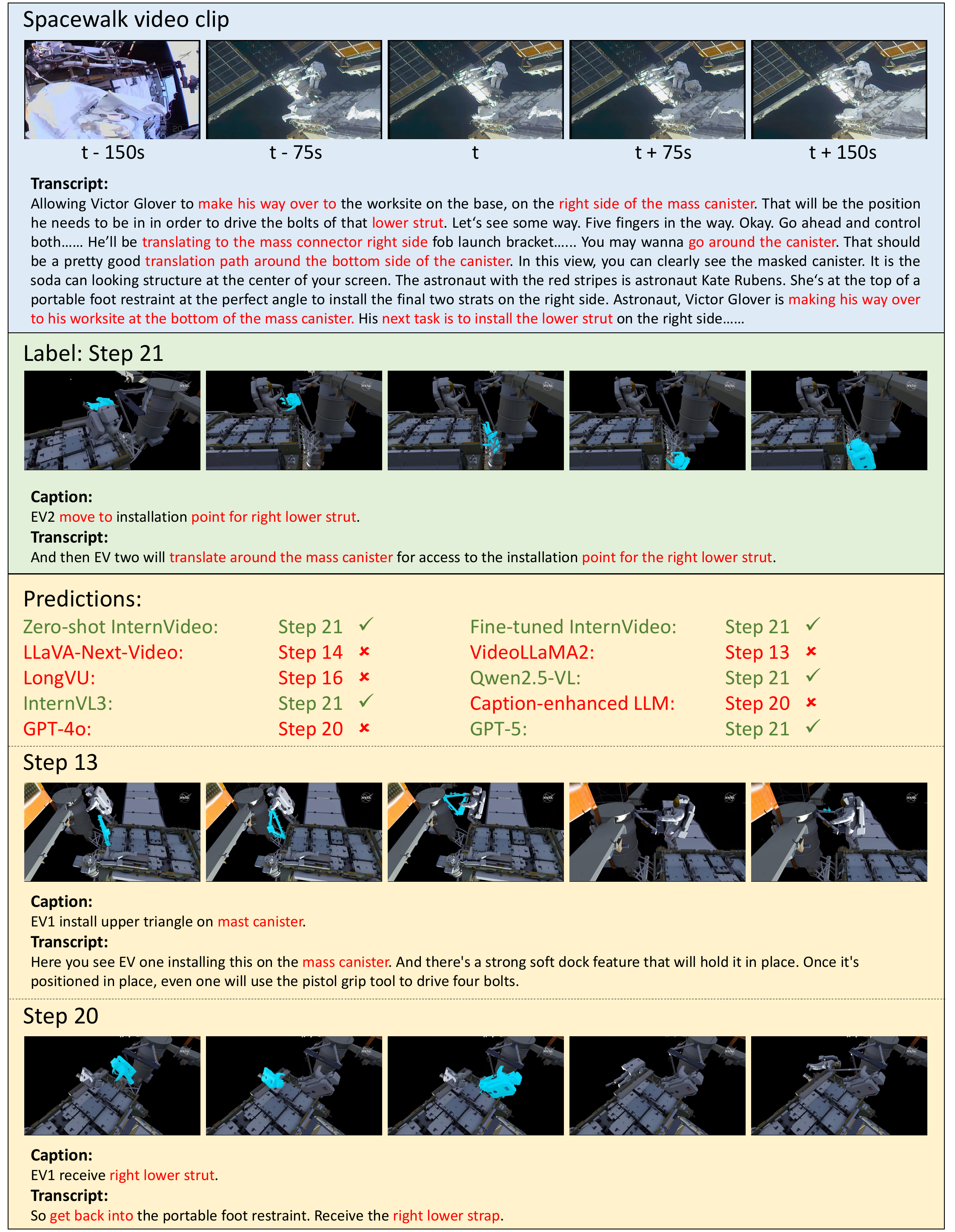}
    \caption{In this example of step recognition, both the spacewalk video and transcript indicate that the astronaut is translating around the mass canister, which aligns with the transcript of step 21. However, five of the VLLMs mismatch it, probably based on keywords ``mast canister'' (step 13) and ``lower strut'' (step 20).}
    \label{fig:example_task1_2}
    \vspace{-10pt}
\end{figure*}

\begin{figure*}
    \centering
    \vspace{-15pt}
    \includegraphics[width=0.94\linewidth]{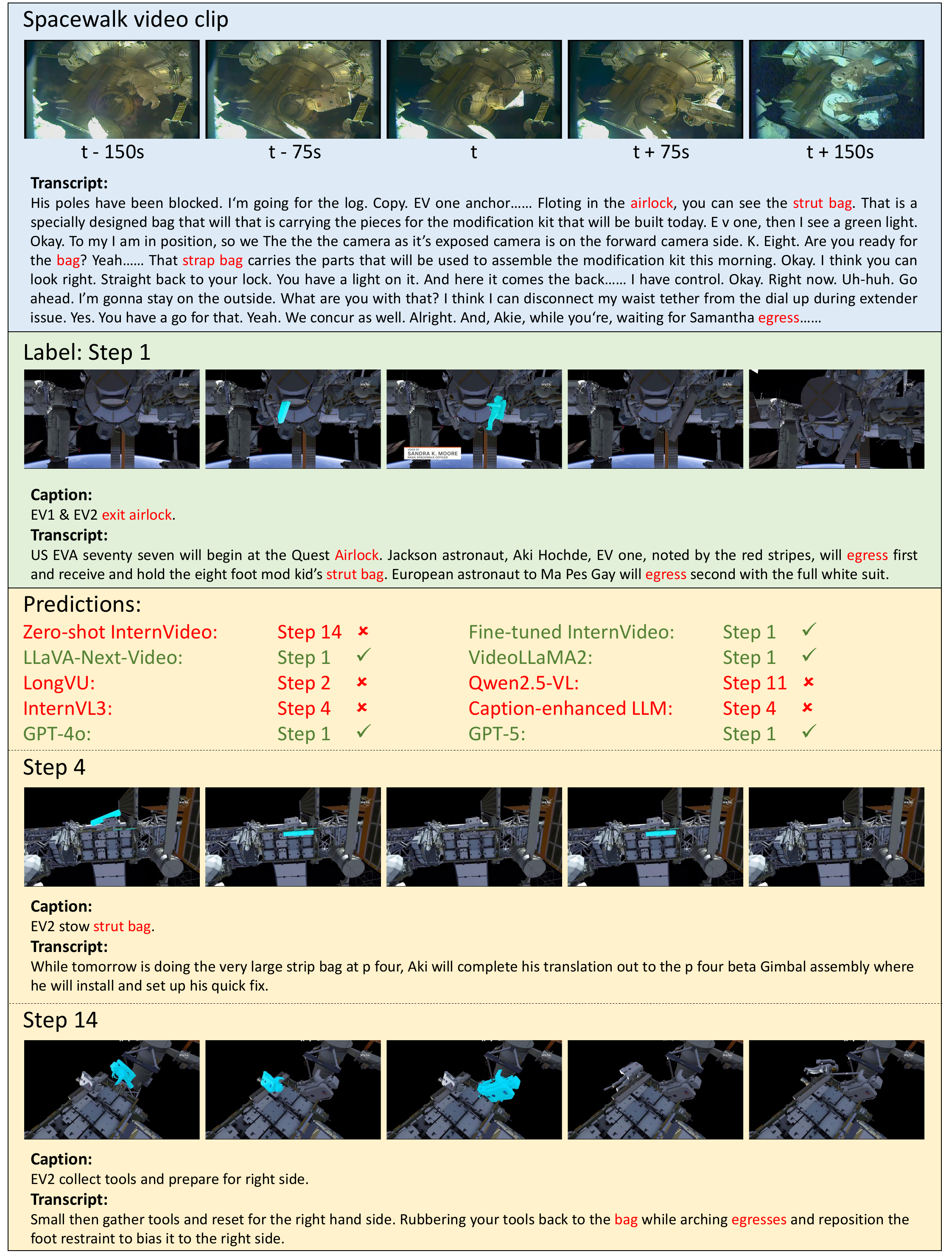}
    \caption{In this example of step recognition, the spacewalk video shows the astronauts exiting the airlock. While zero-shot InternVideo makes an incorrect prediction, fine-tuned InternVideo learns the concept of airlock and gives the true answer. The mistakes made by zero-shot InternVideo, InternVL3, and caption-enhanced LLM are probably due to keywords ``strut bag'' (step 4) and ``egress'' (step 14).}
    \label{fig:example_task1_3}
    \vspace{-10pt}
\end{figure*}

\begin{figure*}
    \centering
    \vspace{-10pt}
    \includegraphics[width=0.94\linewidth]{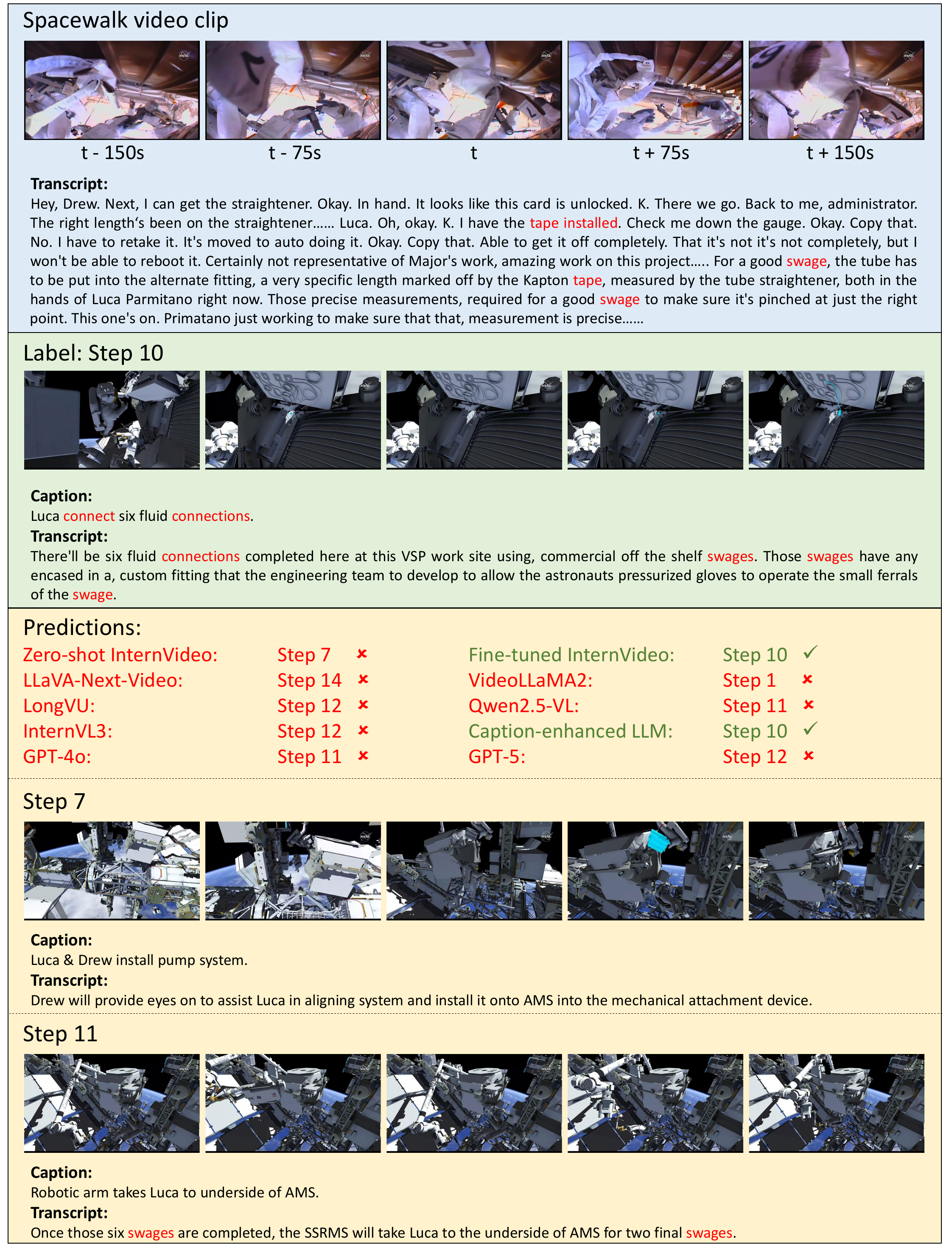}
    \caption{In this example of step recognition, the spacewalk videos shows the astronaut connecting swages, while its transcript also mentions them. However, only fine-tuned InternVideo and caption-enhanced LLM make the correct predictions. The mismatch to step 11 by GPT-4o is possibly because of the keyword ``swage''.}
    \label{fig:example_task1_4}
    \vspace{-10pt}
\end{figure*}

\begin{figure*}
    \centering
    \vspace{-20pt}
    \includegraphics[width=0.94\linewidth]{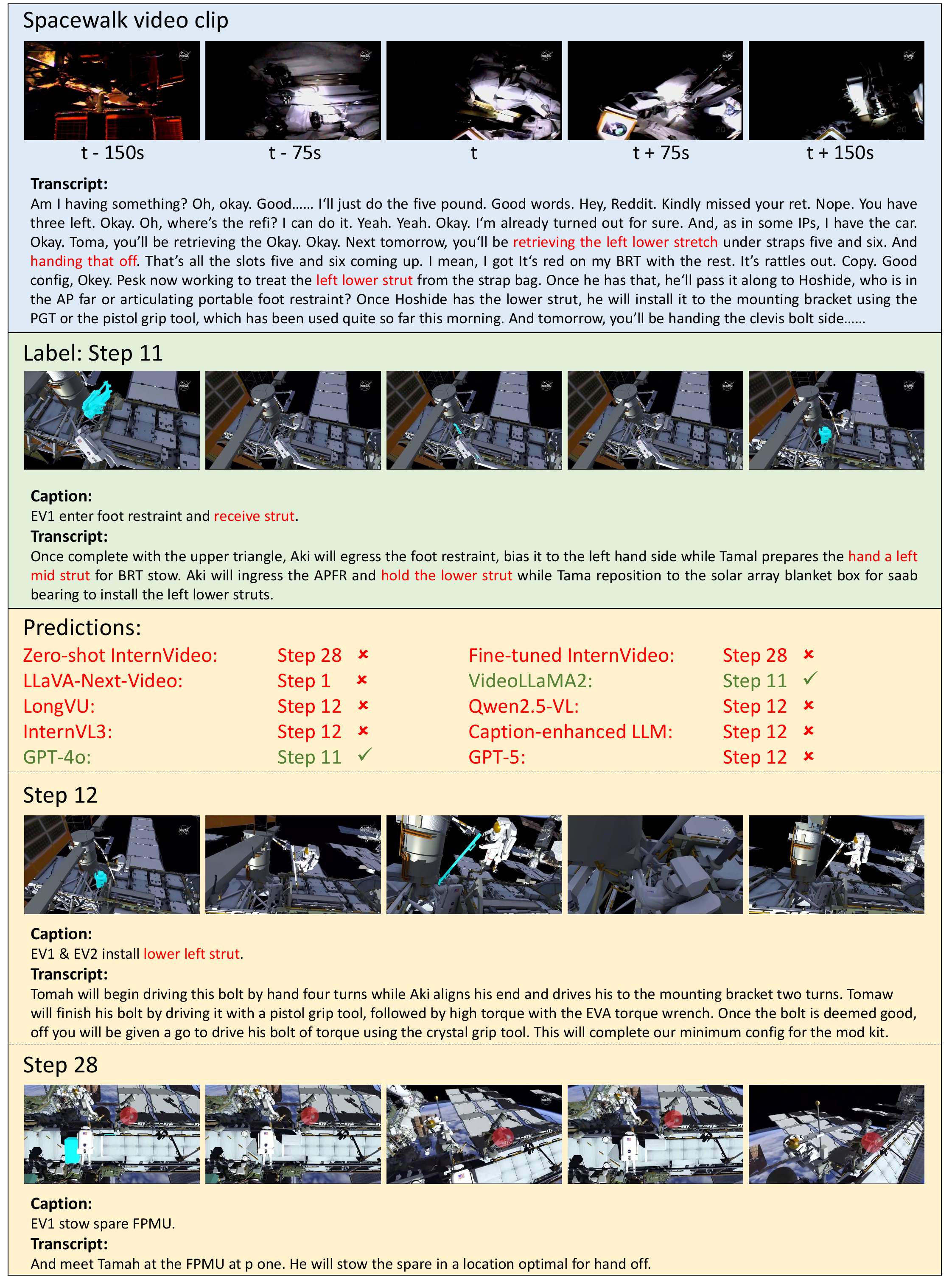}
    \caption{In this example of step recognition, the spacewalk video is dark but the transcript includes a command that one astronaut would hand off a strut to the other. As this example is complicated to understand, only VideoLLaMA2 and GPT-4o predict the step label correctly.}
    \label{fig:example_task1_5}
    \vspace{-10pt}
\end{figure*}

\begin{figure*}
    \centering
    \includegraphics[width=0.94\linewidth]{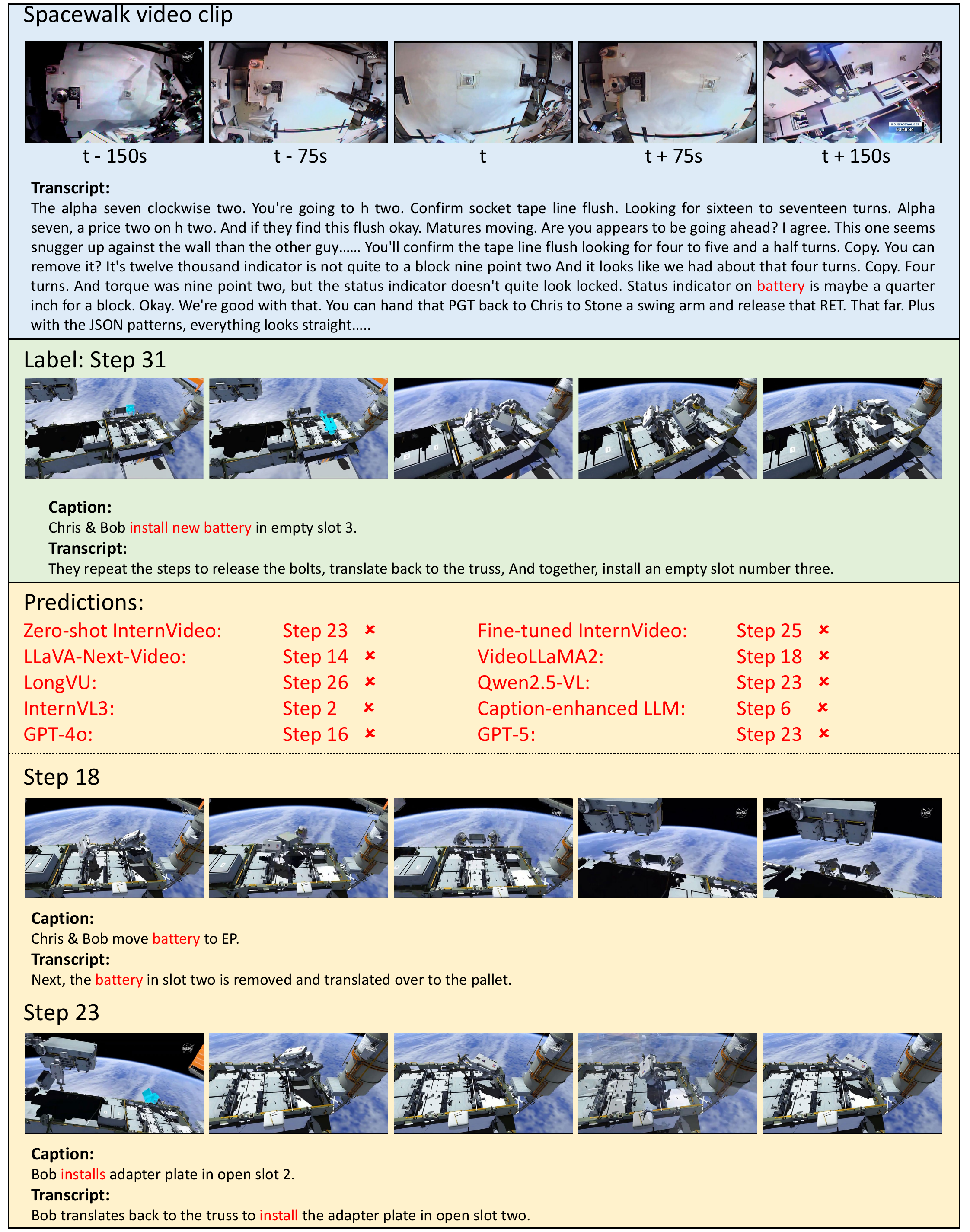}
    \caption{In this example of step recognition, the video clearly shows an astronaut tightening the screws on a white battery. However, all the models fail to recognize it.}
    \label{fig:example_task1_6}
    \vspace{-10pt}
\end{figure*}

We illustrate examples on the step recognition task with a context window length $w=5\text{min}$ from Figure~\ref{fig:example_task1_1} to Figure~\ref{fig:example_task1_6}. The model makes predictions mainly based off keywords shared by the transcript of the spacewalk video and the caption/transcript of the step animation. Besides, it is also aware of the objects (\emph{e.g.}, adapater plate in Figure~\ref{fig:example_task1_1}) and actions (\emph{e.g.}, moving around in \ref{fig:example_task1_2}) appearing in both the spacewalk and animation videos. While the models correctly recognize some steps, these cues can also lead to failure cases. Sharing keywords in the transcripts does not necessarily mean that the spacewalk clip belongs to the step. The models need to better utilize the semantic details in the spacewalk video and transcript to determine the corresponding step. Moreover, all-layer fine-tuning corrects InternVideo's predictions in some examples. The model learns the visual concept of airlock in Figure~\ref{fig:example_task1_3} and swages in Figure~\ref{fig:example_task1_4}. Finally, there are still difficult examples for the evaluated models (Figure~\ref{fig:example_task1_5} and \ref{fig:example_task1_6}), demonstrating the room for model improvement.

\begin{figure*}
    \centering
    \vspace{30pt}
    \includegraphics[width=\linewidth]{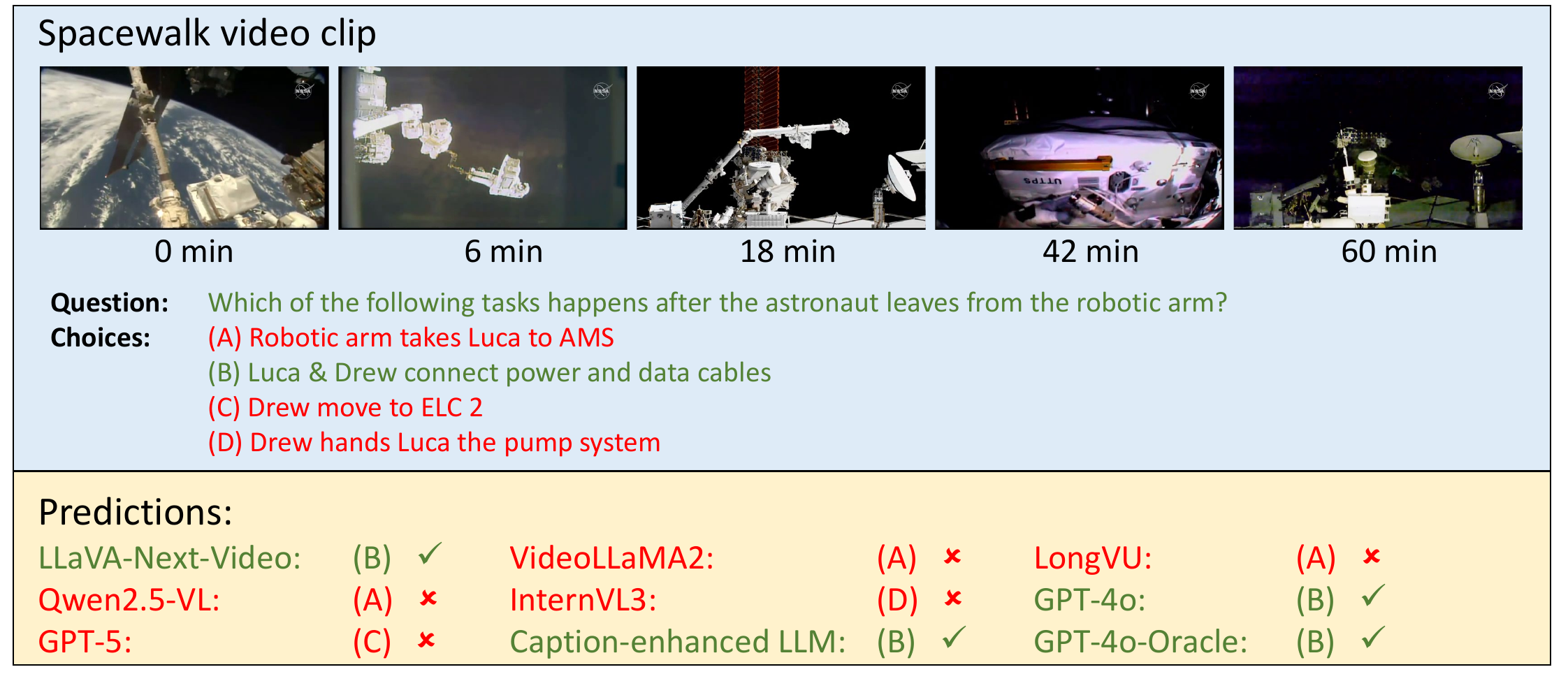}
    \caption{In this example of spacewalk question answering, the astronaut is on the robotic arm in the first 18 minutes. After that, as shown in the 42nd minute, they are connecting cables to an equipment. So the answer to the question is ``(B) Luca \& Drew connect power and data cables''.}
    \label{fig:example_qa_1}
    \vspace{30pt}
\end{figure*}

\begin{figure*}[t]
    \centering
    \includegraphics[width=\linewidth]{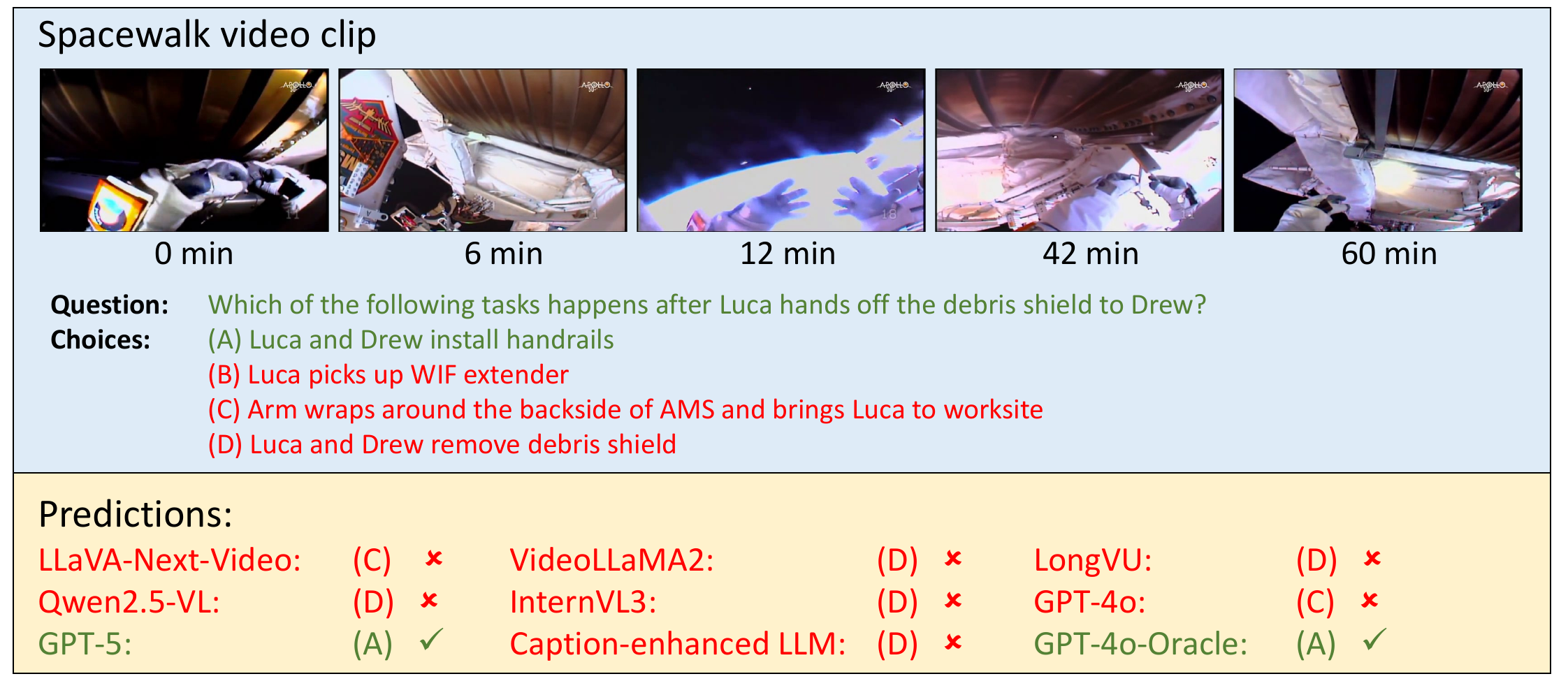}
    \caption{In this example of spacewalk question answering, the astronaut first remove and jettison the debris shield and then install handrails. They remove the debris shield in the 0th minute, hand it off in the 6th minute, and jettison it in the 12th minute. So the only option happened after they hand off the debris shield is ``(A) Luca and Drew install handrails''. Only GPT-5 and the GPT-4o oracle model with spacewalk mission animation correctly answer the question.}
    \label{fig:example_qa_2}
    \vspace{20pt}
\end{figure*}

\begin{figure*}[t]
    \centering
    \includegraphics[width=\linewidth]{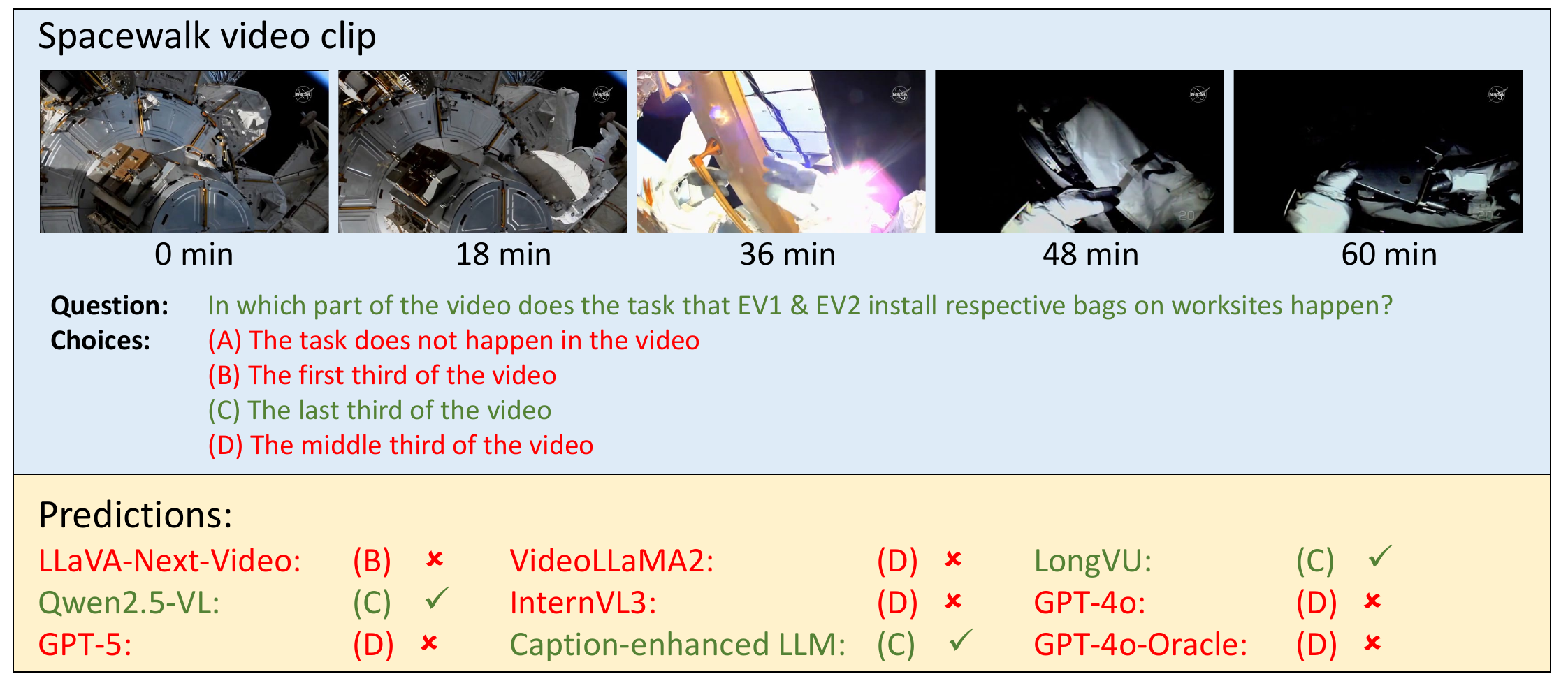}
    \caption{In this example of spacewalk question answering, the astronauts exit the airlock in the first third of the video, move along the handrails in the middle third, and finally install the respective bags in the last third. Therefore, the correct answer is ``(C) The last third of the video''. Only LongVU, Qwen2.5-VL, and the caption-enhanced LLM give the correct answer to the question.}
    \label{fig:example_qa_3}
\end{figure*}

In \Cref{fig:example_qa_1,fig:example_qa_2,fig:example_qa_3}, we show qualitative results on the question answering task. In \Cref{fig:example_qa_1}, LLaVA-Next-Video, GPT-4o, and Caption-enhanced LLM recognize that astronauts are connecting the cables after leaving from the robotic arm. However, In \Cref{fig:example_qa_2}, only GPT-5 and GPT-4o with animation oracle correctly identify the temporal relations between tasks. Some of the models choose ``(D) Luca and Drew remove debris shield'', which indeed happens before they hand off the debris shield. In \Cref{fig:example_qa_3}, many of the models have difficulty localizing the spacewalk task of interest.

\section{Trends on Ours and Other Benchmarks}
\label{app:trend}

    \begin{table}[t]
    \centering
        \scalebox{0.57}{
        \begin{tabular}{c|cccc}
            \toprule
            Models & Spacewalk-SR & Spacewalk-QA & EgoSchema~\citep{EgoSchema} & VideoMME~\citep{fu2024video}\\
            \midrule
            InternVideo & 9.5 & - & 32.1 & - \\
            LLaVA-Next-Video-7B & 11.0$^{*}$ & 25.53$^{*}$ & 44.6 & 35.6 \\
            VideoLLaMA2-7B & 17.3 & 31.7 & 51.7 & 47.9 \\
            Qwen2.5VL-7B & 23.0 & 33.8 & 65.0 & 65.1 \\
            GPT4o & 26.4 & 32.5 & 72.2 & 71.9 \\
            \bottomrule
        \end{tabular}
        }
        \vspace{-8pt}        
        \caption{Model performance reported on ours and other benchmarks. *: Different from~\Cref{tab:main_results_task1} because that uses a 34B checkpoint.}
        \vspace{-18pt}
    \label{tab:ranking}
    \end{table}

In \Cref{tab:ranking}, we list a few models' performance on Spacewalk-18 and other video benchmarks. Although our benchmark evaluates novel domain generalization of VLMs, the models exhibit a similar trend on other benchmarks, probably indicating better generalization capability of stronger models.
However, all the models perform much worse on Spacewalk-18 than on other benchmarks, highlighting the necessity of models with better domain generalization.

\section{Additional Task -- Intra-video Retrieval}
\label{app:retrieval}

    In our step recognition task, the list of steps in the corresponding spacewalk mission is provided.
    However, the pre-defined steps are not always available for all the spacewalk mission.
    In this case, the segmentation of spacewalk steps must be done in a unsupervised manner.
    Current approaches to unsupervised video segmentation rely on clustering video features~\citep{kumar2022unsupervised,dwibedi2020counting}.
    This requires models to form distinguishable embeddings across steps. With this in mind, we explore an additional task -- intra-video retrieval.
    This task evaluates the models capability of retrieving video clips that are from the same step as a query video, without any description about the step.

    \subsection{Task Definition}
        In a spacewalk video, given a query timestamp $t_q$ and two candidate timestamps $t_{c1}$ and $t_{c2}$, the task is to determine which of $t_{c1}$ and $t_{c2}$ is of the same step as $t_q$.
        Note that closer timestamps are more likely to be in the same step.
        Therefore, to avoid this shortcut, we involve only two equidistant candidates satisfying $|t_q-t_{c1}|=|t_q-t_{c_2}|$, where one of them to a different step and serves as a hard negative.
        To make them distinguishable, we ensure that the candidates are at least 30 seconds away from the query.
        Moreover, both the query and candidates should be at least 15 seconds away from their corresponding step boundaries to avoid ambiguity.
        Similar to the step recognition task, we set a context window length $w$ and the model can access the video and transcript in the $w$-long windows centered at $t_q$, $t_{c1}$, and $t_{c2}$.

        We construct 2000 samples from each spacewalk video for this task.
        For each annotated spacewalk clip, we first uniformly sample several query timestamps $t_q$'s.
        Then, for each $t_q$, we randomly sample a distance $d$ such that only one of $t_q-d$ and $t_q+d$ is in the same step as $t_q$ and they follow the restrictions in the task definition.
        Finally, we randomly choose one of $t_q-d$ and $t_q+d$ as $t_{c1}$ and the other as $t_{c2}$.
        We adjust the number of samples from each spacewalk clip to ensure roughly uniform numbers of queries from different spacewalk steps. 

        For performance measurement, we calculate the retrieval accuracy as the evaluation metric. Random chance is 50\% accuracy on this task.

    \subsection{Evaluation Methods}
    \subsubsection{Intra-video Retrieval by VLLMs}
        \noindent\textbf{Zero-shot.}
        To evaluate VLLMs on the intra-video retrieval task, we format it as multi-choice video question answering by asking ``\emph{Which candidate shows the same step as the query?}'' after providing the query and candidate videos.
        Specifically, the prompt is as following:
        
        \mybox{
        You are given three spacewalk video clips from a spacewalk mission, one query clip and two candidate clips. The mission can be divided into multiple steps. Please provide a single-number answer (1 or 2) to the following question, and your answer must be either 1 or 2. You must not provide any other response or explanation. If you are not sure, answer with the most likely answer.\\
        Here is the question: \textcolor{blue}{Which candidate shows the same step as the query?}\\
        Query video clip:\\
        Eight uniformly sampled video frames: \textcolor{blue}{<query video frame 1> ... ... <query video frame 8>}\\
        Video speech transcript: \textcolor{blue}{<query video transcript>}\\
        Candidate 1 video clip:\\
        Eight uniformly sampled video frames: \textcolor{blue}{<candidate 1 video frame 1> ... ... <candidate 1 video frame 8>}\\
        Video speech transcript: \textcolor{blue}{<candidate 1 video transcript>}\\
        Candidate 2 video clip:\\
        Eight uniformly sampled video frames: \textcolor{blue}{<candidate 2 video frame 1> ... ... <candidate 2 video frame 8>}\\
        Video speech transcript: \textcolor{blue}{<candidate 2 video transcript>}
        }
        
        Because this task has three videos as input, we only evaluate VLLMs that can process interleaved videos and texts.

    \subsubsection{Intra-video Retrieval by Contrastive VLMs}
        \noindent\textbf{Zero-shot.}
        In this task, we match a query spacewalk clip centered at timestamp $t_q$ with one of two candidate clips centered at $t_{c1}$ and $t_{c2}$ respectively.
        In contrast to step recognition, we \textbf{only use video features} and ignore the transcripts.
        This proved the best in Section~\ref{app:retrieval_result}.
        After calculating the similarities between the query and the candidates, we pick the candidate with higher score as the prediction.
        
        \noindent\textbf{Last-layer Fine-tuning.}
        We freeze the models and train two linear layers to align the query and candidate features. One is $G^q_\theta(\cdot)$ for queries and the other is $G^c_\theta(\cdot)$ for candidates. We optimize cross entropy loss during training. Denoting the query feature as $f^{q}$ and candidate features as $f_1^{c}$ and $f_2^{c}$, it is
        \begin{equation}
            \mathcal{L}=-\log\frac{\exp\left(G^q_\theta(f^q)\cdot G^c_\theta(f_y^c)\right)}{\sum_{1\leq i\leq 2}\exp\left(G^q_\theta(f^q)\cdot G^c_\theta(f_i^c)\right)}.
        \end{equation}
        After training, the candidate with higher similarity to the query, $G^q_\theta(f^q)\cdot G^c_\theta(f_i^c)$, is pick as the prediction.

        We use the same hyperparameters as the last-layer fine-tuning for the step recognition task in \Cref{app:last_layer_hyper}.

        \noindent\textbf{All-layer Fine-tuning.} We fine-tune the backbone of the pre-trained models using the same training objective as that of the last-layer fine-tuning. The models learn to extract features that are critical to distinguish spacewalk steps.
        Specifically, we fine-tune EgoVLP on the intra-video retrieval task using AdamW~\citep{adamw} optimizer with a learning rate of $1e-6$.
        The model is trained for 1 epoch with a batch size of 32.

\subsubsection{Unsupervised Segmentation Method}
\label{app:twfinch}

TW-FINCH \citep{tw-finch} is a clustering-based unsupervised action segmentation method.
We employ it to solve the intra-video retrieval task as a baseline.
For a spacewalk recording in the test set, we first extract InternVideo features with a sliding window.
The window length is 12 seconds and the step is 1 second.
We sample video frames at 1 FPS to feed the video encoder, \emph{i.e.,} 12 frames per video clip.
Based on these clip features, TW-FINCH clusters the entire spacewalk recording into $N_\text{seg}$ segments, each of which is regarded as a step.
For a retrieval query timestamp $t_q$ with two candidate timestamps $t_{c1}$ and $t_{c2}$, if exactly one of the candidates falls in the same segment as $t_q$, we regard it as our prediction.
Otherwise, we randomly pick one of them as the prediction.
We find that the best configuration is $N_\text{seg}=100$ with only video features.

\subsection{Evaluation Results}
\label{app:retrieval_result}

        Table~\ref{tab:main_results_task2_app} shows the model performances on the intra-video retrieval task.
        GPT-4o achieves the best performance of $71\%$, while the $70.93\%$ accuracy of EgoVLP is competitive.
        Different from the step recognition task, last-layer fine-tuning cannot improve the model performance in most cases.
        All-layer fine-tuning only increases the model accuracy very slightly.
        Besides, while the context window length expands, the model performance first increases and then degrades after a peak.
        This observation is in line with the step recognition results, which reveals their incapability of digesting long video contexts.

        Table~\ref{tab:modality_task2_app} ablates the modalities on intra-video retrieval in zero-shot setting.
        The video-only approach achieves the highest accuracy across all models, while the text-only one performs poorly.
        This is different from step recognition where videos contribute less than texts.
        Because the query and candidates here are the same modality with the same visual distribution, whereas we need to match the real spacewalk videos with their animations in the step recognition task. 
        Moreover, combining video and text does not improve the performance, indicating that transcript is effectively noise for the current method on this task.
        This motivates the future development of stronger models that can utilize the text signal on this task to facilitate unsupervised segmentation of spacewalk videos.

\begin{table}[t]
    \centering
        \scalebox{0.7}{
        \begin{tabular}{c|cccccc}
            \toprule
            \multirow{2}{*}{Method} & \multicolumn{6}{c}{Accuracy} \\
            \cmidrule(r){2-7}
            & ($w$=)10s & 20s & 30s & 1min & 2min & 3min\\
            \midrule
            Random & 50.00 & 50.00 & 50.00 & 50.00 & 50.00 & 50.00 \\
            TW-FINCH$^*$ & - & - & - & - & - & 64.10 \\
            \midrule
            \textcolor{gray}{Zero-shot} \\
            \midrule
            EgoVLP & 69.04 & 69.95 & \textbf{70.92} & 65.76 & \textbf{60.20} & 56.43 \\
            VideoCLIP & 66.91 & 67.48 & 67.88 & 63.52 & 50.72 & 52.07 \\
            InternVideo & 67.60 & 69.12 & 69.82 & \textbf{66.76} & 56.46 & 56.45 \\
            GPT-4o & \textbf{71.00} & \textbf{71.00} & 70.83 & 58.50 & 53.08 & \textbf{56.67} \\
            \midrule
            \textcolor{gray}{Last-layer Fine-tuning}\\
            \midrule
            EgoVLP & \textbf{68.63} & \textbf{69.91} & \textbf{70.12} & \textbf{66.51} & \textbf{57.46} & \textbf{52.36} \\
            VideoCLIP & 61.15 & 60.53 & 61.29 & 59.00 & 54.08 & 52.23 \\
            InternVideo & 65.11 & 65.33 & 64.97 & 62.62 & 56.02 & 52.24 \\
            \midrule
            \textcolor{gray}{All-layer Fine-tuning}\\
            \midrule
            EgoVLP & 68.78 & 70.18 & 70.93 & 69.79 & 62.58 & 58.80 \\
            \bottomrule
          \end{tabular}
        }
    \caption{Model performances on the intra-video retrieval task. GPT-4o achieves the highest accuracy while EgoVLP is comparative. $*$: TW-FINCH have unlimited context.}
    \label{tab:main_results_task2_app}
    \end{table}

    \begin{table}[t]
    \centering
        \scalebox{0.85}{
        \begin{tabular}{c|ccc|ccc}
            \toprule
            \multirow{3}{*}{Method} & \multicolumn{6}{c}{Accuracy} \\
            \cmidrule(r){2-7}
            & \multicolumn{3}{c}{$w$ = 20 s} & \multicolumn{3}{c}{$w$ = 30 s} \\
            \cmidrule(r){2-4}
            \cmidrule(r){5-7}
            & V & T & V+T & V & T & V+T \\
            \midrule
            EgoVLP & 69.95 & 48.73 & 63.71 & \textbf{70.92} & 47.85 & 64.95 \\
            VideoCLIP & 67.48 & 51.65 & 56.56 & 67.88 & 50.04 & 57.14 \\
            InternVideo & 69.12 & 51.55 & 63.15 & 69.82 & 50.23 & 63.08 \\
            GPT-4o & \textbf{71.00} & \textbf{52.42} & \textbf{69.25} & 70.83 & \textbf{53.67} & \textbf{69.92} \\
            \bottomrule
        \end{tabular}
        }
        \caption{Ablation about input modality on intra-video retrieval. The models are evaluated in zero-shot scenario. V: video; T: text (captions and transcripts).}
        \label{tab:modality_task2_app}
    \vspace{-10pt}
    \end{table}

\subsection{Qualitative Results}

\begin{figure*}
    \centering
    \includegraphics[width=\linewidth]{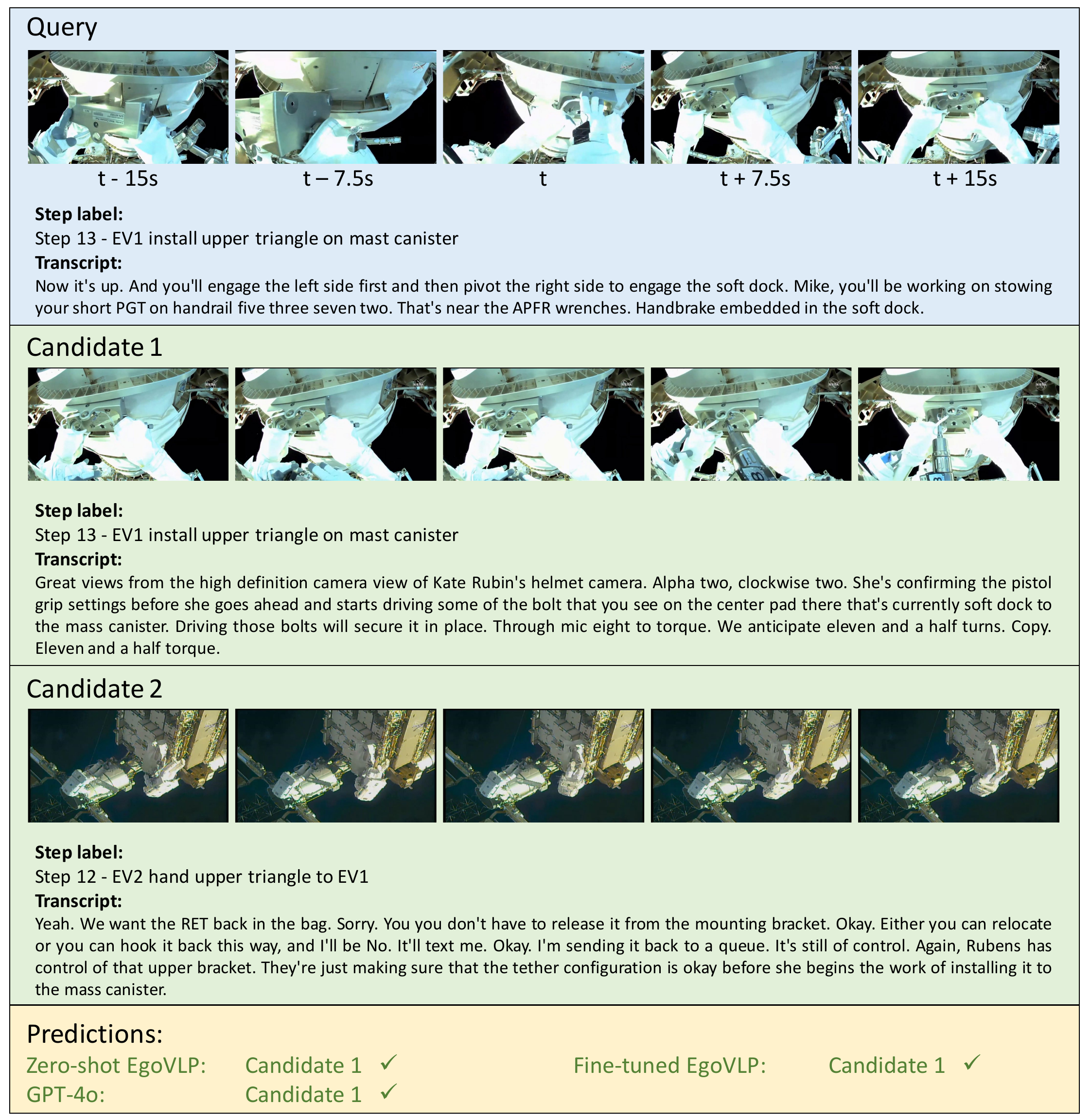}
    \caption{In this example, both the query video and candidate 1 show an astronaut installing an upper triangle. They are in the same scene while candidate 2 is different. All the models make the correct predictions.}
    \label{fig:example_task2_1}
    \vspace{-10pt}
\end{figure*}

\begin{figure*}
    \centering
    \includegraphics[width=\linewidth]{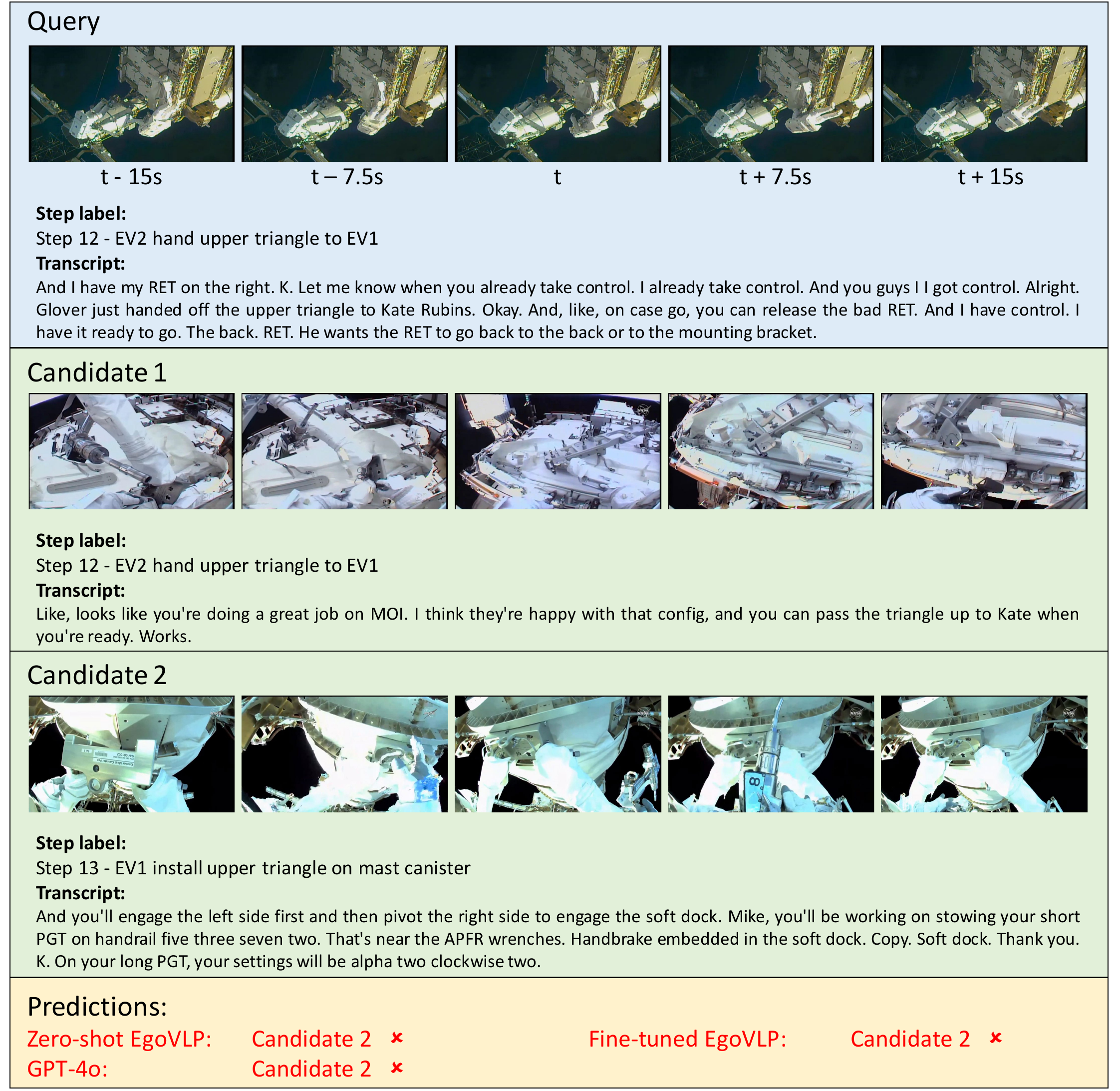}
    \caption{In this example, all three videos are visually dissimilar. The query video shows an astronaut handing an upper triangle to the other from a third-person view, while candidate 1 shows it from a first-person view. In candidate 2, they are installing the upper triangle. However, all the models fail to recognize candidate 1 as the same step as the query.}
    \label{fig:example_task2_6}
    \vspace{-10pt}
\end{figure*}

On the intra-video retrieval task, we show two qualitative examples with context window length $w=30\text{s}$ in \Cref{fig:example_task2_1,fig:example_task2_6}. We find that the models tend to retrieve candidate videos naively based off the visual scene appearance. This helps them do well in simple scenarios (\Cref{fig:example_task2_1}). However, there are more difficult test samples. In Figure~\ref{fig:example_task2_6}, the query and candidate 1 show the same step but from different views. All the models fail in this example. Better spacewalk video understanding capability is necessary for future models to solve it.

\end{document}